\documentclass{article}

\usepackage{microtype}
\usepackage{graphicx}
\usepackage{subfigure}
\usepackage{booktabs}
\usepackage{hyperref}
\usepackage{dsfont}

\usepackage{icmlpreprint}
\usepackage{amsmath}
\usepackage{amssymb}
\usepackage{mathtools}
\usepackage{amsthm}
\usepackage[capitalize,noabbrev]{cleveref}

\theoremstyle{plain}
\newtheorem{theorem}{Theorem}[section]

\newtheorem{lemma}[theorem]{Lemma}

\theoremstyle{definition}
\newtheorem{definition}[theorem]{Definition}

\theoremstyle{remark}

\usepackage[textsize=tiny]{todonotes}

\usepackage[utf8]{inputenc}
\usepackage[T1]{fontenc}
\usepackage{url}
\usepackage{booktabs}
\usepackage{amsfonts}
\usepackage{nicefrac}
\usepackage{microtype}
\usepackage{xcolor}
\usepackage{amssymb}
\usepackage{amsmath}
\usepackage{amsthm}
\usepackage{amsfonts}
\usepackage{mathtools}
\usepackage{xspace}
\usepackage{thm-restate}
\usepackage{cleveref}
\usepackage{stmaryrd}
\usepackage{enumitem}
\usepackage{algorithm}
\usepackage{dsfont}
\usepackage{subcaption}
\usepackage{multirow}
\usepackage{graphicx}
\usepackage{float}

\DeclareMathAlphabet{\mathdcal}{U}{dutchcal}{m}{n}
\SetMathAlphabet{\mathdcal}{bold}{U}{dutchcal}{b}{n}
\DeclareMathAlphabet{\mathdbcal}{U}{dutchcal}{b}{n}

\definecolor{blue}{HTML}{0077BB}
\definecolor{cyan}{HTML}{33BBEE}
\definecolor{green}{HTML}{009988}
\definecolor{orange}{HTML}{EE7733}
\definecolor{red}{HTML}{CC3311}
\definecolor{magenta}{HTML}{EE3377}
\definecolor{grey}{HTML}{BBBBBB}

\crefname{equation}{Eq.}{Eqs.}
\Crefname{equation}{Equation}{Equations}
\crefname{theorem}{Th.}{Ths.}
\Crefname{theorem}{Theorem}{Theorems}
\crefname{corollary}{Cor.}{Cors.}
\Crefname{corollary}{Corollary}{Corollaries}
\crefname{algorithm}{Alg.}{Algs.}
\Crefname{algorithm}{Algorithm}{Algorithms}
\crefname{remark}{Rm.}{Rms.}
\Crefname{remark}{Remark}{Remarks}

\newcommand{\wbf}{\mathbf{w}}
\newcommand{\xbf}{\ensuremath{\mathbf{x}}}
\newcommand{\zbf}{\ensuremath{\mathbf{z}}}
\newcommand{\zerobf}{\mathbf{0}}

\newcommand{\Acal}{\ensuremath{\mathcal{A}}}
\newcommand{\Bcal}{\ensuremath{\mathcal{B}}}
\newcommand{\Ccal}{\ensuremath{\mathcal{C}}}
\newcommand{\Dcal}{\ensuremath{\mathcal{D}}}
\newcommand{\Fcal}{\ensuremath{\mathcal{F}}}
\newcommand{\Hcal}{\ensuremath{\mathcal{H}}}
\newcommand{\Mcal}{\ensuremath{\mathcal{M}}}
\newcommand{\Ncal}{\ensuremath{\mathcal{N}}}
\newcommand{\Pcal}{\ensuremath{\mathcal{P}}}
\newcommand{\Scal}{\ensuremath{\mathcal{S}}}
\newcommand{\Tcal}{\ensuremath{\mathcal{T}}}
\newcommand{\Xcal}{\ensuremath{\mathcal{X}}}
\newcommand{\Ycal}{\ensuremath{\mathcal{Y}}}
\newcommand{\Zcal}{\ensuremath{\mathcal{Z}}}

\newcommand{\Ebb}{\ensuremath{\mathbb{E}}}
\newcommand{\Rbb}{\ensuremath{\mathbb{R}}}
\newcommand{\Nbb}{\ensuremath{\mathbb{N}}}

\newcommand{\Rfrak}{\ensuremath{\mathfrak{R}}}

\newcommand{\LB}{\left[}
\newcommand{\RB}{\right]}
\newcommand{\LC}{\left\{}
\newcommand{\RC}{\right\}}
\newcommand{\RN}{\right\vert}
\newcommand{\LN}{\left\vert}
\newcommand{\LP}{\left(}
\newcommand{\RP}{\right)}

\newcommand{\wrt}{{\it w.r.t.}\xspace}
\newcommand{\eg}{{\it e.g.}\xspace}
\newcommand{\ie}{{\it i.e.}\xspace}
\newcommand{\iid}{{\it i.i.d.}\xspace}

\newcommand{\defeq}{:=}

\DeclareMathOperator*{\EE}{\Ebb}

\DeclareMathOperator*{\vect}{\mathrm{vec}}
\DeclareMathOperator*{\leaky}{\mathrm{Leaky}}
\DeclareMathOperator*{\proj}{\mathrm{Proj}}

\newcommand{\Irm}{\mathrm{I}}
\newcommand{\KL}{\mathrm{KL}}
\newcommand{\KLr}{\overline{\KL}}
\newcommand{\Hell}{H^2}
\newcommand{\TV}{TV}
\newcommand{\kl}{\mathrm{kl}}
\newcommand{\W}{\mathrm{W}}
\newcommand{\Lip}{\mathrm{Lip}}

\newcommand{\D}{\Dcal}
\renewcommand{\H}{\Hcal}
\newcommand{\Hb}{\overline{\Hcal}}
\newcommand{\loss}{\ell}
\renewcommand{\P}{\pi}
\newcommand{\Q}{\rho}
\newcommand{\R}{\Rbb}
\newcommand{\N}{\Nbb}
\renewcommand{\S}{\Scal}
\newcommand{\Risk}{\text{R}}
\newcommand{\Riskhat}{\hat{\Risk}}

\newcommand{\x}{\xbf}
\newcommand{\Y}{\Ycal}
\newcommand{\Z}{\Zcal}
\newcommand{\z}{\zbf}
\newcommand{\varepsilonbf}{\boldsymbol{\varepsilon}}
\newcommand{\rad}{\mathdbcal{E}}

\newcommand{\yeast}{{\sc Yeast}\xspace}
\newcommand{\phishing}{{\sc Phishing}\xspace}
\newcommand{\mushrooms}{{\sc Mushrooms}\xspace}
\newcommand{\mnist}{{\sc MNIST}\xspace}
\newcommand{\fashion}{{\sc FashionMNIST}\xspace}

\newif\ifnotappendix
\notappendixtrue

\title{\vspace{-1cm}%
Tighter Generalisation Bounds via Interpolation}
\author{Paul Viallard$^1$\thanks{The work was done when the author was affiliated with Inria Paris.} \and Maxime Haddouche$^{2,3}$ \and Umut {\c S}im{\c s}ekli$^4$ \and Benjamin Guedj$^2$}

\date{
$^1$Univ Rennes, Inria, CNRS IRISA - UMR 6074, F35000 Rennes, France\\%
$^2$Inria, University College London\\
$^3$Université de Lille\\
$^4$Inria, CNRS, Ecole Normale Supérieure, PSL Research University, Paris, France\\
}

\begin{document}

\thispagestyle{fancy}

\maketitle

\begin{abstract}
This paper contains a recipe for deriving new PAC-Bayes generalisation bounds based on the $(f, \Gamma)$-divergence, and, in addition, presents PAC-Bayes generalisation bounds where we interpolate between a series of probability divergences (including but not limited to KL, Wasserstein, and total variation), making the best out of many worlds depending on the posterior distributions properties.
We explore the tightness of these bounds and connect them to earlier results from statistical learning, which are specific cases.
We also instantiate our bounds as training objectives, yielding non-trivial guarantees and practical performances.
\end{abstract}

\section{Introduction}
\label{sec:intro}

Understanding the generalisation ability of learning algorithms is a fundamental question in machine learning \citep{vapnik1974theory,vapnik2000learning} which has become even more challenging with the emergence of deep learning.
Typically, a generalisation bound is stated with respect to a \emph{learning problem} described by a tuple $(\H, \Z, \loss)$ consisting of a hypothesis (or predictor) space $\H$ equipped with a distance $d_\H$, a data space $\Z$, and a loss function $\loss : \H\times \Z \rightarrow \R$.
It aims to bound the \emph{population risk} of a given hypothesis $h$, defined as $\Risk_{\D}(h) = \EE_{\z\sim\D}[\loss(h , \z)]$ (where $\D$ denotes the unknown \emph{data distribution} over $\Z$). 

As $\D$ is typically not known in practice, a hypothesis $h$ is usually built by (approximately) minimising the \emph{empirical risk}, given by $\Riskhat_{\S}(h) = \frac{1}{m}\sum_{i=1}^{m}\loss(h , \z_i)$, where $\S = \{ \z_i \in \Z \}_{i=1}^{m}$ is a dataset of $m$ data points, independent and identically distributed (\iid) from $\D$. 
We define the generalisation gap of a hypothesis $h$ as $\Delta_\S(h)\defeq \vert\Risk_{\D}(h)-\Riskhat_{\S}(h)\vert$.
If we can obtain an upper-bound on $\Delta_\S(h)$, we immediately get an upper-bound on the population risk as $\Risk_{\D}(h) \leq  \Riskhat_{\S}(h) + \Delta_\S(h)$.
Hence, a general pattern for developing generalisation bounds has been based on deriving inequalities of the following form: with probability at least $1-\delta$, 
\begin{align*}
\Delta_{\S}(h)  \le \sqrt{\frac{\textsc{Complexity} + \ln{\frac{1}{\delta}}}{m}}.
\end{align*}
The \textsc{Complexity} term exhibits a facet of the intrinsic complexity of the learning problem such as a certain notion of richness of the predictor class $\Hcal$.
This encompasses VC dimension \citep{vapnik2000learning} and Rademacher complexity \citep{bartlett2001rademacher,bartlett2002rademacher}). 
However, the VC dimension, for instance, happens being too large when huge classes of predictors are considered, such as deep neural networks.
Then, exploiting the Bayesian paradigm of learning \emph{posterior} knowledge from data and prior modelling of the environment allows exploiting tools from information theory \citep{cover2001elements} such as mutual information as complexity measure \citep{neal2012bayesian}.  
\\
Beyond learning with the Bayesian paradigm described above, PAC-Bayes emerged as a relatively recent branch of learning theory (see \citealp{guedj2019primer,alquier2021user,hellstrom2023generalization}) and has known a significant increase of interest in the past two decades, providing non-vacuous generalisation guarantees for neural networks, alongside novel learning algorithms (see \citealp{dziugaite2017computing,perez2021progress,perez2021tighter,perezortiz2021learning}, among others).
Apart from neural networks, this framework allows tackling various learning settings such as reinforcement learning \citep{fard2010pac}, online learning \citep{haddouche2022online}, multi-armed bandits \citep{seldin2011pac,seldin2012pac,sakhi2023pac}, meta-learning \citep{amit2018meta,farid2021generalization,rothfuss2021pacoh,rothfuss2022pac,ding2021bridging} or adversarially robust learning \citep{viallard2021pac} to name but a few. 

The major part of PAC-Bayes bounds use the \emph{Kullback-Leibler (KL) divergence} between a data-dependent ``posterior'' distribution and a ``prior'' distribution (to be precised in \Cref{sec:background}) as a complexity measure \citep{mcallester1999some,mcallester2003pac,catoni2007pac,germain2009pac,tolstikhin2013pac,kuzborskij2019efron,rivasplata2020pac,haddouche2021pac,haddouche2023wasserstein}, and it has been shown recently that it is possible to obtain PAC-Bayesian bounds for any \emph{$f$-divergence} (including KL) \citep{alquier2018simpler,ohnishi2021novel,picard2022change}.
Such developments allow using various complexities to understand generalisation as there is no apparent reason to prioritise KL. 
\\
However, a major drawback of $f$-divergences is that they are irrelevant to understand the generalisation ability of deterministic predictors.  Aiming at alleviating this problem, another line of research (\citealp{amit2022integral,haddouche2023wasserstein,viallard2023learning}) developed PAC-Bayesian generalisation bounds based on \emph{integral probability metrics (IPMs)}, including the well known $1$-Wasserstein distance.
\\
Knowing which complexity must be chosen in practice is challenging. For instance, KL-based PAC-Bayes bounds are expected to go to zero for large sample size (then to explain generalisation), but they cannot deal with deterministic predictors (KL is infinite in this case), which is inconsistent, \eg, with practical optimisation of deep nets.
On the other hand Wasserstein-based bounds deal with deterministic predictors, but do not possess an explicit convergence rate with respect to the number of data points $m$ in most cases.
We tackle this question in this work and propose elements of answer in the form of a unifying framework intricating PAC-Bayes with the recent notion of $(f,\Gamma)$-divergences \citep{birrell2022divergences}.

\textbf{Contributions.} We derive two generic recipes exploiting $(f,\Gamma)$-divergences, acting as a prelude to generalisation bounds. 
Using these recipes, we further elaborate novel generalisation bounds interpolating various $f$-divergences and IPMs.
In many cases, the derived bounds are tractable, contrasting with many results in \citealp{ohnishi2021novel,picard2022change}.
In particular, we focus on a result interpolating the KL divergence and the Wasserstein distance, connecting with a large part of PAC-Bayes literature.
Our results are also general enough to unveil rather surprising links between PAC-Bayes and Rademacher complexity and to obtain effortless generalisation bounds for heavy-tailed stochastic differential equations.
All of our results show that combining PAC-Bayes learning with $(f,\Gamma)$-divergences provide a unifying framework for generalisation, which result in tighter bounds. We finally conduct experiments exhibiting the benefits of interpolating $(f,\Gamma)$-divergences and IPMs.
More precisely, we propose a novel learning algorithm involving a combination of $f$-divergences and IPMs as regulariser, yielding, in many cases, better results than considering those complexities separately.

\textbf{Outline} \Cref{sec:background} gather background for both PAC-Bayes learning and $(f,\Gamma)$-divergences.
Those two notions intricates in \Cref{sec:generic-bounds} where generic templates are provided.
New PAC-Bayes bounds interpolating $f$-divergences and IPMs are then derived in \Cref{sec:various-complexities}. 
\Cref{sec:connection} exploits previous results to connect PAC-Bayes with Rademacher complexities and heavy-tailed SGD.
Finally, \Cref{sec:expe} gathers experiments. Supplementary background and results are gathered in \Cref{sec:supp-background,sec:supp-results}.
\Cref{sec:proofs} gathers the postponed proofs and \Cref{sec:supp-expe} concludes this paper with additional insights on experiments.

\section{Notation and Background}
\label{sec:background}
\textbf{PAC-Bayes framework.} 
PAC-Bayesian bounds focus on a \emph{randomised} setting where the hypothesis is drawn from a \emph{posterior} distribution $\Q \in\Pcal(\H)$, where $\Pcal(\H)$ denotes the set of probability distributions defined on $\H$.
This posterior is designed from the training set $\S$ and a \emph{prior distribution} $\P\in\Pcal(\H)$.
A classical PAC-Bayesian result is \citet[Theorem 5]{maurer2004note} (the so-called McAllester bound), which states that, for $\ell\in[0,1]$, with probability at least $1-\delta$, for any posterior distribution $\Q\in\Mcal(\H)$,
\begin{align}
\label{eq:mcallester}
 \EE_{h\sim\Q}\Big[ \Delta_{\S}(h) \Big] \le \sqrt{\frac{\KL(\Q\|\P) + \ln{\frac{2\sqrt{m}}{\delta}}}{2m}},
\end{align}
where $\P \in \Mcal(\H)$ is any data-free distribution and $\KL$ denotes the Kullback-Leibler divergence.

\textbf{Background for $(f;\Gamma)$ divergences.}
We denote by $\Mcal(\Hcal)$ (resp. $\Mcal_b(\Hcal)$) the set of measurable (resp. bounded measurable) functions from $\mathcal{H}$ to $\Rbb$.
$(f,\Gamma)$-divergences (\Cref{def:f-gamma-divergence}) recently emerged in \citet{birrell2022divergences} as complexity measures interpolating $f$-divergences and $\Gamma$-IPMs (\Cref{def:div-ipm}).
They require two elementary building blocks: a convex function $f\in \Irm\rightarrow \Rbb$ where $\Irm\subseteq \Rbb$ is an interval determining the $f$-divergence and a set $\Gamma\subseteq\Mcal(\Hcal)$.

\begin{definition}
\label{def:div-ipm}
For any distribution $\Q,\P\in\Pcal(\H)^2$, the $\Gamma$-IPM between $\P$ and $\Q$ is defined as $W^{\Gamma}(\Q, \P) \defeq \sup_{\varphi\in\Gamma}\LC \EE_{h\sim\Q}\varphi(h) - \EE_{g\sim\P}\varphi(h) \RC$.
Also, if $\Q \ll \P$, then the $f$-divergence between $\P$ and $\Q$ is $D_{f}(\Q\|\P) \defeq \EE_{h\sim\P}f\LP\frac{d\Q}{d\P}(g)\RP,$
otherwise $D_{f}(\Q\|\P) \defeq +\infty$.
\end{definition}

Both $f$-divergences and $\Gamma$-IPMs have been broadly used in PAC-Bayes as those complexity measures possess useful variational formulations allowing to transfer the generalisation ability of the posterior distribution of interest to a prior one, usually verifying better properties (\eg, not depending on $\S$).
A famous example stands for the KL divergence (take $f\colon x \mapsto x\ln x$) which satisfies the 'change of measure formula'  \citep{csizar1975divergence,donsker1976asymp}, leading to various PAC-Bayes bounds as the McAllester \citep{mcallester2003pac} and Catoni ones \citep{catoni2007pac}.
Such a variational formulation is more generally attained for any $f$-divergence (\citealp{nguyen2010estimating,broniatowski2006minimization}, see \citealp[Proposition 50]{birrell2022divergences} for a proof) and leads to generalisation bounds for hostile data \citep{alquier2018simpler}. 
The variational formulation is recalled:
\begin{align}
D_{f}(\Q\|\P) \defeq \sup_{\varphi\in\Mcal_{b}(\H)}\LC \EE_{h\sim\Q}\varphi(h) - \Lambda^{\P}_{f}(\varphi) \RC\label{eq:variational-f-div}
\end{align}
where $\Lambda^{\P}_{f}(\varphi) \defeq \inf_{c\in\R}\LC c + \EE_{h\sim\P}f^{*}(\varphi(h)-c)\RC,$ and $f^*$ is the Legendre transform of $f$. Note that, for the KL divergence, $\Lambda^{\P}_{f}(\varphi)=\ln\EE_{h\sim\P}e^{\varphi(h)}$. 
\\
Thus, a variational formulation is at the core of the interplay between $f$-divergences and generalisation, and this conclusion also holds for $\Gamma$-IPMs by definition (even if the supremum is taken on a restricted subset $\Gamma\subset \Mcal_b(\Hcal)$.
Indeed, PAC-Bayesian generalisation bounds involving IPMs have been proposed \citep{amit2022integral}, with a particular focus on Wasserstein distances \citep{haddouche2023wasserstein,viallard2023learning}. 

Despite this similarity, $f$-divergences and $\Gamma$-IPM have been mainly thought separately until the recent work of \citet{birrell2022divergences}, which proposed the unifying notion of $(f,\Gamma)$-divergences detailed in \Cref{def:f-gamma-divergence}.

\begin{definition}[$(f, \Gamma)$-divergence]\label{def:f-gamma-divergence}
For any $\Gamma\subset\Mcal_{b}(\H)$, we define for any $\Q\in\Pcal(\H)$ and $\P\in\Pcal(\H)$ the $(f, \Gamma)$-divergence (with $f\in\Fcal(a,b)$) defined by
\begin{align*}
D^{\Gamma}_{f}(\Q\|\P) \defeq \sup_{\varphi\in\Gamma}\LC \EE_{g\sim\Q}\varphi(g) - \Lambda^{\P}_{f}(\varphi) \RC.
\end{align*}
\end{definition}
$(f,\Gamma)$-divergences consider the variational formulation \eqref{eq:variational-f-div} of $f$-divergences restrained to the set $\Gamma$ of the associated IPM.
They satisfy an important variational formula (namely the \emph{infimal convolution formula}) \citep[Theorem 8-1.]{birrell2022divergences}, recalled in \eqref{eq:inf-conv-formula}.
\begin{align}
D^{\Gamma}_{f}(\Q\|\P) \le \inf_{\eta\in\Pcal(\H)}\LC W^{\Gamma}(\Q, \eta) + D_{f}(\eta\|\P) \RC.\label{eq:inf-conv-formula}
\end{align} 
\Cref{eq:inf-conv-formula} justifies describing the $(f,\Gamma)$-divergences as 'interpolations' between $f$-divergences and $\Gamma$-IPMs as the former is controlled by an optimal combination of the last two.
We show in \Cref{sec:generic-bounds}  that \Cref{eq:inf-conv-formula} paves the way towards generalisation bounds. 

\section{Elementary Steps Towards Generalisation}
\label{sec:generic-bounds}

We propose two new results (\Cref{theorem:pb-f-gamma,theorem:pb-f-gamma-mcdiarmid})  involving $(f,\Gamma)$-divergences. Those bounds act as general templates to recover generalisation bounds throughout this work when particularising the function $f$ and the set $\Gamma$. We start with \Cref{theorem:pb-f-gamma}, our most general statement. 

\begin{restatable}{theorem}{theorempbfgamma}\label{theorem:pb-f-gamma}
Let $\phi_\Scal\in\Gamma$, $\delta\in[0,1]$ and $\P\in \Pcal(\Hcal)$. With probability at least $1-\delta$ over $\Scal\sim\D^m$, we have for all $\Q\in\Pcal(\H)$
\begin{align*}
\EE_{h\sim\Q}\phi_{\Scal}(h) \le D^{\Gamma}_{f}(\Q\|\P){+}\ln\!\frac{1}{\delta}{+}\ln\!\LB\EE_{\Scal\sim\D^m}\exp\LP\Lambda^{\P}_{f}(\phi_\Scal)\RP\RB\!.
\end{align*}
\end{restatable}

As the exponential moment of \Cref{theorem:pb-f-gamma} may be hard to control, we propose \Cref{theorem:pb-f-gamma-mcdiarmid} which alleviate this assumption at the cost of the \emph{bounded difference} property (see \eg \citealp{boucheron2013conc}).

\begin{restatable}{theorem}{theorempbfgammamcdiarmid}\label{theorem:pb-f-gamma-mcdiarmid}
Let $\phi_\Scal\in\Gamma,\delta\in[0,1]$ and $\P\in \Pcal(\Hcal)$. Assume there exists a function $g_\S^\P$ such that for any $\S,\P$, $\Lambda^{\P}_{f}(\phi_\Scal) \leq B^{\P}(\phi_\Scal)$.
Assume the bounded-difference property on $B^{\P}$, \ie, 
\begin{align*}
\forall i\in\{1,\dots,m\},\ \sup_{\Scal\in \Zcal^m, \zbf_i\in\Zcal}| B^{\P}(\phi_\Scal) - B^{\P}(\phi_{\S_i'}) | \le c_i,
\end{align*}
where $\Scal'_i$ is the learning sample where $\zbf_i$ in $\Scal$ is replaced by $\zbf_i'$.
Then, we have, with probability at least $1-\delta$ over $\Scal\sim\D^m$, we have for all $\Q\in\Pcal(\H)$
\begin{align*}
\EE_{h\sim\Q}\phi_{\Scal}(h) \le D^{\Gamma}_{f}(\Q\|\P) + \EE_{\Scal\sim\D^m} B^{\P}(\phi_\Scal) + \sqrt{\frac{\ln\frac{1}{\delta}}{2}\sum_{i=1}^{m}c_i^2}.
\end{align*}
\end{restatable}

\textbf{Analysis of \Cref{theorem:pb-f-gamma,theorem:pb-f-gamma-mcdiarmid}.} 
In PAC-Bayes, $\phi_\S$ is often a function of the generalisation gap, \eg, $\phi_\S= \lambda \Delta_\S$ or  $\lambda \Delta_\S^2$ with $\lambda>0$.
Our two results holds with the assumption $\phi_\S \in \Gamma$.
In particular, $\phi_\S$ has to be bounded (satisfied \eg, for classification tasks).
Such an assumption has positive consequences on tightness: from \Cref{eq:inf-conv-formula}, we know that $D^{\Gamma}_{f}(\Q\|\P) \le \min\LP D_{f}(\Q\|\P), W^{\Gamma}(\Q,\P)\RP$, thus our bounds are tighter than existing results involving separately either $f$-divergences or $\Gamma$-IPMs: interpolating complexities leads to tightness.
\Cref{theorem:pb-f-gamma} involves an exponential moment on the Legendre transform $\Lambda^{\P}_{f}(\phi_\Scal)$, which is of interest in \Cref{sec:kl-wass}, when considering KL divergence.
To go beyond KL, we exploit \Cref{theorem:pb-f-gamma-mcdiarmid}, which uses the bounded difference assumption to relax the dependency on $\Lambda^{\P}_{f}(\phi_\Scal)$.
We show the benefits of this relaxation in \Cref{sec:various-complexities} to obtain generalisation bounds holding with various $f$-divergences such as the Hellinger distance or the Reverse KL.

\section{Generalisation Bounds with Various Complexity Measures}
\label{sec:various-complexities}
    
We particularise the results of \Cref{sec:generic-bounds} to obtain novel PAC-Bayes bounds involving various pairs of divergence and IPM.
We focus on the important particular case of \Cref{sec:kl-wass}, which shows that a generalisation bound interpolating KL and Wasserstein is reachable and tightens existing results.
\Cref{sec:bound-div} provides PAC-Bayes bounds involving a wide range of divergences alongside a generic IPM.

\subsection{A Fundamental Example: a PAC-Bayes Bound Interpolating KL Divergence and Wasserstein}
\label{sec:kl-wass}

When $f_{\KL}(x)= x\ln(x)$ and $\Gamma={\Lip^L_b}$ is the set of bounded $L$-Lipschitz functions, then the associated $(f,\Gamma)$-divergence interpolates between the KL divergence and Wasserstein distance in the sense that the optimal combination of those complexities upper bounds the $(f,\Gamma)$:
\begin{align}
    D^{\Gamma_{\Lip^L_b}}_{f_{\KL}}(\Q\|\P) \le \inf_{\eta\in\Pcal(\H)}\LC LW_1(\Q, \eta) + \KL(\eta\|\P) \RC.\label{eq:inf-conv-kl-wass}
\end{align}
\Cref{eq:inf-conv-kl-wass} comes immediately from \Cref{eq:inf-conv-formula} using that $D_{f_{\KL}} = \KL$ and $W^{Lip^L_b} \leq LW_1$ as $\Lip^L_b = L \Lip_b^1\subseteq L\Lip^1$.
We now are able to particularise \Cref{theorem:pb-f-gamma}.

\begin{restatable}{theorem}{theoremKLgamma}\label{theorem:KL-wass}
Assume that $\ell \in [0,1]$. Assume that, for any $\delta'\in (0,1)$, with probability $1-\delta'$, $h\rightarrow\Delta_\S^2(h)$ is $L(m,\delta')$-Lipschitz. Thus, for any data-free prior $\P$, with probability at least $1-\delta$ over $\Scal\sim\D^m$, we have for all $\Q\in\Pcal(\H)$, any $\eta\in\Pcal(\H)$,
\ifnotappendix
\begin{multline*}
\EE_{h\sim\Q}\Delta_{\S}(h)\le \\
\sqrt{ L(m,\nicefrac{\delta}{2})\W_1(\Q, \eta) + \frac{\KL(\eta\|\P)+ \ln\frac{4\sqrt{m}}{\delta}}{2m}}.
\end{multline*}
\else
\begin{align*}
\EE_{h\sim\Q}\Delta_{\S}(h)\le \sqrt{ L(m,\nicefrac{\delta}{2})\W_1(\Q, \eta) + \frac{\KL(\eta\|\P)+ \ln\frac{4\sqrt{m}}{\delta}}{2m}}.
\end{align*}
\fi
\end{restatable}

\textbf{Comparison with literature.}
To our knowledge, \Cref{theorem:KL-wass} is the first PAC-Bayes generalisation bounds involving \emph{simultaneously} a KL divergence alongside a Wasserstein distance. 
\Cref{theorem:KL-wass} shows that considering bounded Lipschitz losses leads to a bound tighter than the McAllester bound \citep{mcallester2003pac} involving a KL term and also \citet[Theorem 11]{amit2022integral} involving a Wasserstein distance. Our additional tightness comes from considering and $(f,\Gamma)$-divergence and the infimal convolution formula \eqref{eq:inf-conv-kl-wass}. Note also that the proof technique behind \Cref{theorem:KL-wass} allow us to similarly improve on various PAC-Bayes bounds. Indeed, we provide in \Cref{sec:improv-existing} variants of Catoni's bound (\citealp{catoni2007pac}, Theorem 4.1 of \citealp{alquier2016properties}), of the Catoni's fast rate bound \citet[Theorem 2]{mcallester2013pacbayesian} and of the supermartingale bounds of \citet{haddouche2023pac,viallard2023learning} involving simultaneously a Wasserstein distance and a KL divergence. This comes at the cost of bounded Lipschitz losses while unbounded subgaussian losses are allowed in \citet[Theorem 4.1]{alquier2016properties}, even heavy-tailed ones in \citet{haddouche2023pac,viallard2023learning}. 

\textbf{On the value of the Lipschitz constant.}
If $\ell$ is $1$-Lipschitz and bounded by $1$, then $\Delta_S^2$ is at least $4$-Lipschitz.
This deterministic constant may be insufficient as the KL divergence is always attenuated by the impact of dataset size $m$. 
\citet{amit2022integral} showed that, for a finite $\H$, $\Delta_\S^2$ is $\frac{8}{m}L\ln(\nicefrac{2|\H|}{\delta})$-Lipschitz.
\citet{haddouche2023wasserstein} extended this result to any compact $|\H|$ of $\Rbb^d$ at the cost of an explicit dependency on the dimension.
In order to obtain sharper Lipschitz constants for general $\H$ without an explicit impact of the dimension, we develop in \Cref{theorem:lipschitz-computable} a novel high probability Lipschitz constant based on an empirical surrogate of the Rademacher complexity.
This new result, while independent of the $(f,\Gamma)$-divergence, is of interest for the practical optimisation of \Cref{theorem:KL-wass} as it diminishes the impact of the Wasserstein distance \wrt the KL divergence.

\subsection{PAC-Bayes Bounds Beyond KL Divergence}
\label{sec:bound-div}

The general results of \Cref{sec:generic-bounds} go beyond KL divergence.
We show in \Cref{theorem:KL-reverse-hellinger-gamma} that tractable PAC-Bayes bounds are reachable through $(f,\Gamma)$-divergences for the \emph{reverse KL} ($\KLr$) defined as
$\KLr(\Q\|\P) \defeq -\EE_{g\sim\P}\ln(\tfrac{d\Q}{d\P}(g))$ and the \emph{squared Hellinger} $\Hell(\Q\|\P) \defeq \EE_{g\sim\P}(\sqrt{\nicefrac{d\Q}{d\P}(g)}-1)^2$. 
\\
Our results, to our knowledge, are the first PAC-Bayes generalisation bounds involving those complexities. 

\begin{restatable}{theorem}{theoremKLreverseHellgamma}\label{theorem:KL-reverse-hellinger-gamma}
For $\Delta_\Scal\in\Gamma$. With probability at least $1-\delta$ over $\Scal\sim\D^m$, we have for all $\Q\in\Pcal(\H)$ and $\eta\in\Pcal(\H)$:\\
Reverse-KL bound:
\ifnotappendix%
\begin{multline}
\label{eq:reverse-KL}
\EE_{h\sim\Q}\Delta_{\S}(h) \le 2\W^{\Gamma}(\Q, \eta) + 2\KLr(\eta\|\P)\\
+ \sqrt{\frac{1}{m}} + \ln\LP 1{+}\frac{1}{m}\RP\sqrt{2m\ln\frac{1}{\delta}}.
\end{multline}
\else
\begin{align}
\label{eq:reverse-KL}
\EE_{h\sim\Q}\Delta_{\S}(h) \le 2\W^{\Gamma}(\Q, \eta) + 2\KLr(\eta\|\P)
+ \sqrt{\frac{1}{m}} + \ln\LP 1{+}\frac{1}{m}\RP\sqrt{2m\ln\frac{1}{\delta}}.
\end{align}
\fi
Hellinger bound:
\ifnotappendix%
\begin{multline}
\label{eq:hellinger}
\EE_{h\sim\Q}\Delta_{\S}(h) \le 2\W^{\Gamma}(\Q, \eta) + 2\Hell(\eta\|\P)\\
+ \sqrt{\frac{1}{m}} + \frac{2}{m+1}\sqrt{2m\ln\frac{1}{\delta}}.
\end{multline}
\else
\begin{align}
\label{eq:hellinger}
\EE_{h\sim\Q}\Delta_{\S}(h) \le 2\W^{\Gamma}(\Q, \eta) + 2\Hell(\eta\|\P)
+ \sqrt{\frac{1}{m}} + \frac{2}{m+1}\sqrt{2m\ln\frac{1}{\delta}}.
\end{align}
\fi
TV bound:
\ifnotappendix%
\begin{multline}
    \label{eq:tv}
    \EE_{h\sim\Q}\Delta_{\S}(h) \le \W^{\Gamma}(\Q, \eta) +\TV(\eta, \P)\\
    + \sqrt{\frac{1}{4m}} + \sqrt{\frac{\ln\frac{1}{\delta}}{2m}}.
    \end{multline}
\else
\begin{align}
    \label{eq:tv}
    \EE_{h\sim\Q}\Delta_{\S}(h) \le \W^{\Gamma}(\Q, \eta) +\TV(\eta, \P) + \sqrt{\frac{1}{4m}} + \sqrt{\frac{\ln\frac{1}{\delta}}{2m}}.
\end{align}
\fi
\end{restatable}

\textbf{Analysis of the bounds.}
To our knowledge, we propose the first tractable PAC-Bayes bounds involving the reverse KL or squared Hellinger.
Our novelty lies in the use of the bounded difference property in \Cref{theorem:pb-f-gamma-mcdiarmid}, which alleviates the exponential moment of \Cref{theorem:pb-f-gamma}.
This why the work of \citet{ohnishi2021novel} only proposes PAC-Bayesian bounds for Rényi or $\chi^2$ divergences and that \citet{picard2022change} is forced to consider raw exponential moments.
On the contrary, we bypass this constraint and show that reverse KL and squared Hellinger enjoy empirical generalisation bounds involving IPMs. Note that here, both $f$-divergences and $\Gamma$-IPMs are not explicitly attenuated by $m$.
This issue has already been unveiled in \citet{viallard2023learning} (only for Wasserstein distances), and they showed that this limitation could be attenuated through data-dependent priors (see their 'Role of data-dependent priors' paragraph on page 5), the same remark applies here as our bounds encompass Wasserstein PAC-Bayes bounds.

\section{Novel Connections in Statistical Learning}
\label{sec:connection}
Combining $(f,\Gamma)$-divergences with PAC-Bayes learning allows us {\it (i)} to tighten existing PAC-Bayes bounds by interpolating and {\it (ii)} exhibit new links within generalisation theory.
More precisely, we bridge the gap between PAC-Bayes and the Rademacher complexity, showing that a Rademacher-based generalisation bound is retrievable from \Cref{theorem:pb-f-gamma-mcdiarmid}.
We also exploit the flexibility of \Cref{theorem:KL-wass} to build a bridge between generalisation and optimisation, through a novel generalisation bound for heavy-tailed Stochastic Gradient Descent (SGD).

\subsection{A Rigorous Link Between PAC-Bayesian and Rademacher-based Bounds}
\label{sec:rademacher}

We connect the PAC-Bayes framework with another subfield of statistical learning in an unexpected way.
Indeed, we show that, from a PAC-Bayes bound, it is possible to derive a generalisation bound involving the Rademacher complexity $\Rfrak(\H)\!\defeq\! \EE_{\Scal\sim\D^m}\EE_{\varepsilonbf\sim\rad^m} \LP \frac{1}{m}\sum_{i=1}^m \varepsilon_i\ell(h,z_i) \RP$, where $\varepsilon_i\sim\rad$ are $\{-1,+1\}$ random Rademacher variables \citep[Definition 2]{bartlett2001rademacher}.
To do so, we first consider \eqref{eq:tv} derived in \Cref{theorem:KL-reverse-hellinger-gamma} with the Wasserstein distance as IPM.
However, instead of considering the Euclidean distance, we use the set of Lipschitz functions \wrt the Kronecker distance $d_{\text{Kron}}(h,h')\defeq\mathds{1}_{h\neq h'}$. Then, the Wasserstein boils down to the TV distance~\citep{lindvall1992lectures,gibbs2002choosing} and we need to estimate the Lipschitz constant.   

\textbf{Lipschitz constant based on the Rademacher complexity.} 
Surprisingly, the Rademacher complexity emerges from the estimation of the Lipschitz constant \wrt the Kronecker distance, this fact is stated in \Cref{theorem:lipschitz} below.

\begin{restatable}{theorem}{theoremLipschitz}\label{theorem:lipschitz}
For any hypothesis set $\H$, for any $L$-Lipschitz loss $\loss:\Hcal\times\Zcal\to\R$ (by considering $d_{\text{Kron}}$), with probability at least $1-\delta$ over $\S\sim\D^m$
\begin{align*}
h\mapsto \Delta_{\Scal}(h) \ \ \text{is}\ \ L(m,\delta){=}\!\LP 4\Rfrak(\H){+}L\sqrt{\tfrac{2\ln\frac{2}{\delta}}{m}}\RP\!\text{-Lipschitz}.
\end{align*}
\end{restatable}
Put into words, the Lipschitz constant $L(m,\delta)$ of the gap $\Delta_\S$ is upper bounded by the Rademacher complexity. 
This also confirms the intuition that we could hope for beneficial statistical effects for the value of this constant, since we are dealing with a generalisation gap.\\

\textbf{Retrieving classical Rademacher-based generalisation bound from PAC-Bayes bounds.}
Using \Cref{theorem:lipschitz}, we propose a rigorous link between PAC-Bayes and Rademacher complexity in \Cref{corollary:rademacher}.

\begin{restatable}{corollary}{corollaryrademacher}\label{corollary:rademacher}
Assume that $\ell \in [0,1]$. With probability at least $1-\delta$ over $\Scal\sim\D^m$, we have for all $\Q\in\Pcal(\H)$,
\ifnotappendix
\begin{multline}
    \label{eq:mix-rad-pb}
    \EE_{h\sim\Q}\Delta_{\S}(h) \le \!\LP 4\Rfrak(\H){+}L\sqrt{\tfrac{2\ln\frac{4}{\delta}}{m}}\RP\!\TV(\Q,\eta)  \\+\TV(\eta\|\P)
    + \sqrt{\frac{1}{4m}} + \sqrt{\frac{\ln\frac{2}{\delta}}{2m}},
\end{multline}
and in particular, for all $h\in \H$,
\begin{multline}
    \label{eq:rad}
 \Delta_{\S}(h) \le 4\Rfrak(\H){+}\sqrt{\frac{2\ln\frac{4}{\delta}}{m}} + \sqrt{\frac{1}{4m}} + \sqrt{\frac{\ln\frac{2}{\delta}}{2m}}.
\end{multline}
\else
\begin{align*}
    \EE_{h\sim\Q}\Delta_{\S}(h) \le \!\LP 4\Rfrak(\H){+}L\sqrt{\tfrac{2\ln\frac{4}{\delta}}{m}}\RP\!\TV(\Q, \eta) +\TV(\eta\|\P)
    + \sqrt{\frac{1}{4m}} + \sqrt{\frac{\ln\frac{2}{\delta}}{2m}},
\end{align*}
and in particular, for all $h\in \H$,
\begin{align*}
 \Delta_{\S}(h) \le 4\Rfrak(\H){+}\sqrt{\frac{2\ln\frac{4}{\delta}}{m}} + \sqrt{\frac{1}{4m}} + \sqrt{\frac{\ln\frac{2}{\delta}}{2m}}.
\end{align*}
\fi
\end{restatable}

In other words, \Cref{eq:rad} shows that it is possible to obtain a Rademacher-based generalisation bound from a PAC-Bayes one. 
This connection is, to the best of our knowledge, novel. However, a similar link between PAC-Bayes and VC dimension has been shown in \citet[Corollary 8]{amit2022integral}.
Note that our result directly implies a link with VC dimension, as Rademacher complexity involves a VC-dimension based generalisation bound \citep[Chapter 3]{mohri2012foundations}.
Moreover, it has already been shown that Rademacher-based bound implies PAC-Bayes results.
Indeed, \citet{kakade2008complexity} derived a PAC-Bayes bound, starting from Rademacher complexity, and later \cite{yang2019fast} managed to obtain fast-rate PAC-Bayes bounds from shifted Rademacher processes.
In this context, \Cref{eq:rad} draws additional insights on the interplays between PAC-Bayes, Rademacher complexity, and VC dimension.

\Cref{eq:mix-rad-pb} goes even further and fully exploit the flexibility of $(f,\Gamma)$-divergences (and the two TV distances) by interpolating between two different types of generalisation bounds.
More precisely, if $\eta=\Q$, we recover a PAC-Bayes bound, meaning we aim to explain generalisation through the Bayesian paradigm of a meaningful prior, available before training.
On the other hand, taking $\eta= \P$ and crudely bounding the TV by 1 gives \Cref{eq:rad} which fully focuses on the structure of the predictor class through Rademacher complexity, ignoring prior knowledge.
Then, the take-home message is that, to understand generalisation, the information provided by both prior knowledge and the structure of the predictor class can be simultaneously exploited.

\subsection{Generalisation Bounds for Heavy-tailed SDEs}
\label{sec:ht-sgd}

It has been recently shown that SGD might exhibit a heavy-tailed behavior when a large step-size is chosen \cite{simsekli2019tail,gurbuzbalaban2020heavy,hodgkinson2021multiplicative,pavasovic2023approximate} and several works have illustrated that such heavy tails can have a direct impact on the generalisation properties of SGD \cite{simsekliS2020haussdorf,barsbey2021heavy}. 
\\
One fruitful approach for analysing the heavy-tailed behavior in SGD has been based on modelling the SGD recursion by using a heavy-tailed continuous-time dynamics that is expressed by the following stochastic differential equation (SDE) \cite{ simsekliS2020haussdorf,raj2022algorithmic, raj2023algo,dupuis2023mutual}:
\begin{align}
\mathrm{d}h_{t} =  -  \mathrm{d} \nabla_h \hat{F}_\S(h_t)  \mathrm{d}t + \sigma  \mathrm{d} \mathrm{L}_t^{\alpha}, \; h_0\in\mathcal{H},\label{eq:sgd-continuous}
\end{align}
where $\H=\Rbb^d$, 
$\hat{F}_\S(h) = \Riskhat_\S(h) + \lambda \|h\|^2$ is the regularised empirical risk with some $\lambda >0$, $\sigma\in\mathbb{R}_+$ and $\mathrm{L}_t^{\alpha}$ is a rotationally symmetric $\alpha$-stable Lévy process, defined in \Cref{sec:supp-background}, which is a random process whom parameter $\alpha\in (0,2)$ characterises the heavy-tailed behaviour: the smaller $\alpha$, the heavier the tail.
On the other hand, when $\alpha=2$, the process becomes the Brownian motion, hence having Gaussian tails. 

Under certain conditions on $\hat{F}_\S$, the process defined by \eqref{eq:sgd-continuous} is ergodic with an invariant distribution $\Q_\alpha$ corresponding to the asymptotic distribution of $h_t$ \cite{wang2016wasserstein,xie2020ergodicity}.
Hence, it has been of interest to understand the generalisation error of this process at stationarity, \ie, $\Delta_{\S}(h_t)$ when $t \to \infty$.
In other words, we set $\Q$ to be the invariant distribution of \eqref{eq:sgd-continuous}, and we would like to estimate $\Delta_{\S}(h)$ when $h \sim \Q$. 
\\
Attacking this problem with existing PAC-Bayes approaches would require directly estimating a notion of distance between $\P$ and $\Q_\alpha$, such as the KL divergence or the $1$-Wasserstein distance.
Unfortunately, directly upper-bounding such quantities is a highly non-trivial task since $\rho$ does not admit an analytical form except for $\alpha =2$.
Hence, for obtaining generalisation bounds for these processes, alternative approaches have been developed, mainly based on algorithmic stability \cite{raj2022algorithmic,raj2023algo}: and fractal geometry 
\cite{ simsekliS2020haussdorf,dupuis2023generalization,hodgkinson2022generalization,lim2022chaotic,dupuis2023mutual}.

In this section, we will show that, thanks to \Cref{theorem:KL-wass}, we can easily provide a  generalisation bound for the distribution $\Q$ by reducing the task into two sub-tasks that readily have solutions in the literature.
More precisely, to invoke \Cref{theorem:KL-wass}, we will design an intermediate distribution $\eta$, such that both $\W_1(\Q, \eta)$ and $\KL(\eta\|\P)$ can be estimated by existing tools.\\
For designing such $\eta$, we consider the classical overdamped Langevin SDE \cite{roberts1996exponential}:
\begin{align}
    \mathrm{d} \tilde{h}_{t} =  -  \mathrm{d} \nabla_h \hat{F}_\S(\tilde{h}_t)  \mathrm{d}t + \sigma  \mathrm{d} \mathrm{B}_t, \; \tilde{h}_0\in\mathcal{H},\label{eq:sgd-approx-brownian}
\end{align}
where $B_t$ denotes the standard Brownian motion. We then choose $\eta$ as the invariant measure of this SDE. 
\\
Thanks to this choice for $\eta$, we can now appeal to existing results for invoking \Cref{theorem:KL-wass}: {\it (i)} very recently, \citet{deng2023wasserstein} proved upper-bounds on the Wasserstein-1 distance between $\Q$ and $\eta$, {\it (ii)} whereas the KL-divergence between $\eta$ and appropriate priors $\P$ has also been investigated in PAC-Bayesian frameworks, see \eg, \cite{mou2018generalization}.
These observations are formalised in the next corollary.

\begin{restatable}{corollary}{HTSGD}
\label{theorem:heavy-tailed-sgd}
Let $\P$ be a data-free prior and $\delta\in (0,1)$.
Assume $\ell \in [0,1]$. Assume that, for any $\delta'\in (0,1)$, with probability $1-\delta'$, $h\rightarrow\Delta_\S^2(h)$ is $L(m,\delta')$-Lipschitz.
Assume that $\nabla \hat{F}_\S$ is quasi-smooth and regular (\textbf{(H1)} and \textbf{(H2)} in \Cref{sec:proof-ht-sgd}).
Thus, for any $\alpha_0\in (1,2)$, with probability at least $1-\delta$ over $\Scal\sim\D^m$, we have, for any $\alpha \in [\alpha_0,2]$,
\ifnotappendix
\begin{multline*}
\EE_{h\sim\Q_\alpha}\Delta_{\S}(h) \le \\
\sqrt{ L(m,\nicefrac{\delta}{2}) C_{\alpha_0} f(\alpha,d) + \frac{\KL(\eta\|\P)+ \ln\frac{4\sqrt{m}}{\delta}}{2m}},
\end{multline*}
\else
\begin{align*}
\EE_{h\sim\Q_\alpha}\Delta_{\S}(h) \le 
\sqrt{ L(m,\nicefrac{\delta}{2}) C_{\alpha_0} f(\alpha,d) + \frac{\KL(\eta \|\P)+ \ln\frac{4\sqrt{m}}{\delta}}{2m}},
\end{align*}
\fi
where $f(\alpha,d)= d\ln(d)(2-\alpha)\ln\LP \nicefrac{1}{2-\alpha}\RP$ and $C_{\alpha_0}>0$.
\end{restatable}
    
This result provides a novel high-probability generalisation bound for heavy-tailed SDEs with minimal effort.
Note that $(2-\alpha) \log(1/(2-\alpha)) \leq \alpha -1$, this factor then attenuates the value of $C_{\alpha_0}$, as it scales proportionally to $\nicefrac{1}{\alpha_0-1}$, according to \citet{deng2023wasserstein}.
Here, the $\KL(\eta,\P)$ term can be upper-bounded by using \cite{mou2018generalization} and is attenuated by $m$.
Furthermore, while implicit, the impact of the dimension is also attenuated by $m$ through $L(m,\delta)$ and \Cref{theorem:lipschitz,theorem:lipschitz-computable}. 

Furthermore, notice that \Cref{theorem:heavy-tailed-sgd} is strictly better than considering a single Wasserstein distance in \Cref{theorem:KL-wass} (\ie, $\mathrm{W}_1(\rho,\P)$) and applying the triangle inequality so that we would need to estimate $\mathrm{W}_1(\rho,\eta)$ and $\mathrm{W}_1(\eta,\P)$.
Indeed, by doing so, we would lose the convergence rate of $\nicefrac{1}{m}$, which attenuates the KL divergence, and hence gives arbitrarily more importance to the prior.

\section{Experimental Study}
\label{sec:expe}

In the following section, we first present in \Cref{sec:algorithm} a learning algorithm to minimise the bound of \Cref{theorem:KL-wass} while \Cref{sec:setting-result} introduces the setting and the results.

\subsection{A Novel Learning Algorithm}
\label{sec:algorithm}

Many PAC-Bayesian bounds, including those of \Cref{sec:various-complexities}, have the great advantage to be fully empirical.
Thus, minimising such bounds yields theory-driven learning algorithms. 
More precisely, by rearranging the terms in our results, we can upper-bound the (expected) population risk $\Risk_{\D}(h)$ by the (expected) empirical risk $\Riskhat_{\S}(h)$ alongside the term containing the complexity measures of interest.
In this section, we focus on the minimisation of the bound in \Cref{theorem:KL-wass}; the considered optimisation problem is
\begin{multline}
\min_{\Q\in\Ccal_{\Q},\ \eta\in\Ccal_{\eta}}\Bigg\{\EE_{h\sim\Q}\Riskhat_{\S}(h)\\
+ \sqrt{ L(m{,}\nicefrac{\delta}{2})\W_1(\Q,\eta)+\tfrac{\KL(\eta\|\P)+\ln\tfrac{4\sqrt{m}}{\delta}}{2m}} \Bigg\},\label{eq:alg-KL-wass}
\end{multline}
where $\Ccal_{\Q}\subseteq \Pcal(\Hcal)$ and $\Ccal_{\eta}\subseteq \Pcal(\Hcal)$ are the sets of probability distributions for $\Q$ and $\P$ (to be precised further).
Thanks to this optimisation problem, we are able to find a posterior distribution $\Q$ and intermediate distribution $\eta$ that minimise (approximately) the bound. 
Hence, this algorithm is expected to exploit simultaneously the strengths of $\KL$ and Wasserstein.
\\
However, to compute the bound, it remains {\it (i)} to approximate $L(m{,}\nicefrac{\delta}{2})$ and {\it (ii)} to determine the sets $\Ccal_{\Q}$ and $\Ccal_{\eta}$ on which we can compute the Wasserstein distance and the KL divergence.
To do so, we perform experiments on models $h_{\wbf}\in\H$ parametrised by a vector $\wbf\in\R^d$.
This allows us to consider two types of posterior distributions: a Dirac distribution $\Q=\delta_{\wbf}$ centered on the weights $\wbf$ and a Gaussian distribution $\Q=\Ncal(\wbf, \sigma^2\Irm_d)$, where $\sigma$ is the standard deviation and $\Irm_d$ is the identity matrix in $\Rbb^{d\times d}$.
The metric $d_{\H}(h_{\wbf}, h_{\wbf'})=\|\wbf-\wbf'\|$ becomes the Euclidean one.

\textbf{Computing the Lipschitz constant $L(m, \delta)$.}
Unfortunately, \Cref{theorem:lipschitz} is suited only for the Kronecker distance and is not estimable because the Rademacher complexity involves the expectation on $\Scal\sim\D^m$ and $\varepsilonbf\sim\rad^m$.
Therefore, in the next theorem, we prove an additional bound on the Lipschitz constant (for any metric $d_{\H}$) of the gap that we can estimate.

\begin{restatable}{theorem}{theoremLipschitzcomputable}\label{theorem:lipschitz-computable}
For any hypothesis set $\H$, for any $L$-Lipschitz loss $\loss:\Hcal\times\Zcal\to\R$, with probability at least $1-\delta$ over $\S\sim\D^m$ and $\varepsilonbf\sim\rad^m$
\begin{align*}
h\mapsto\Delta_{\Scal}(h) \ \text{is}\ L(m{,}\delta){=}\!\LP\! 2\mathdbcal{R}_{\Scal}^{\varepsilonbf}(\H){+ }3L\sqrt{\tfrac{2\ln\frac{4}{\delta}}{m}}\RP\!\text{-Lipschitz}, 
\end{align*}
where 
$\mathdbcal{R}_{\Scal}^{\varepsilonbf}(\H)\!\defeq\! \sup_{h\neq h' \in \H}\frac{1}{m}\sum_{i=1}^{m}\varepsilon_i\frac{\LB\loss(h',\zbf_i){-}\loss(h,\zbf_i)\RB}{d_{\H}(h,h')}$.
\end{restatable}

\Cref{theorem:lipschitz-computable} paves the way to the approximation of the Lipschitz constant by mini-batch stochastic optimisation.
Indeed, at each iteration, we have to sample a mini-batch $\Bcal\subseteq\Scal$ and update the vector $\wbf$ and $\wbf'$ by maximizing 
\begin{align*}
\textstyle\frac{1}{|\Bcal|}\sum_{\zbf_i\in\Bcal}\varepsilon_i\frac{\LB\loss(h_{\wbf'},\zbf_i){-}\loss(h_{\wbf},\zbf_i)\RB}{\|\wbf-\wbf'\|}, 
\end{align*}
where $\varepsilon_i$ has been previously drawn.
Moreover, \Cref{theorem:lipschitz-computable} allows going beyond \citet[Theorem 12]{amit2022integral}.
Compared to their theorem, \Cref{theorem:lipschitz-computable} does not require the hypothesis set $\H$ to be finite, while still being a computable constant.
Moreover, our theorem allows us to bypass more assumptions about the learning problem, as is done in Theorem 14 of \citet{amit2022integral} for linear regression.

\textbf{Computing a tractable Wasserstein and KL divergence.}
In order to compute the Wasserstein distance $\W_1(\Q,\eta)$ and the KL divergence $\KL(\eta\|\P)$, we have to define the set of probability distributions $\Ccal_{\Q}$ and $\Ccal_{\eta}$.
Note that restricting the sets $\Ccal_{\Q}\subset\Pcal(\H)$ and $\Ccal_{\eta}\subset\Pcal(\H)$ provides an upper-bound on the infimal convolution formula, recalled in \Cref{eq:inf-conv-formula}.
We restrict the prior distribution $\P$ and the intermediate distribution $\eta$ to be two Gaussian distributions, \ie, we have $\P=\Ncal(\wbf_{\P}, \Irm_d\sigma^2_{\P})$ and $\eta=\Ncal(\wbf_{\eta}, \Irm_d\sigma^2_{\eta})$, where $\wbf_{\P}$ and $\wbf_{\eta}$ are the means while $\sigma_{\P}$ and $\sigma_{\eta}$ are the standard deviations of the Gaussians.
Thanks to these two definitions, the KL divergence is given by
\begin{align*}
\KL(\eta\|\P)\! =\! \frac{1}{2}\!\left[\frac{\sigma^2_{\eta}}{\sigma^2_{\P}}d - d +\frac{1}{\sigma^2_{\P}}\|\wbf_{\eta}{-}\wbf_{\P}\|_2^2 +d\ln\!\LP\frac{\sigma^2_{\P}}{\sigma^2_{\eta}}\RP\right]\!.
\end{align*}
The value of the Wasserstein distance depends on the posterior distribution we are considering.
When the posterior $\Q$ is a Dirac, we have
\begin{align*}
\W_1(\delta_{\wbf},\eta) &\le \W_1(\delta_{\wbf},\delta_{\wbf_{\eta}}) + \W_1(\delta_{\wbf_{\eta}}, \eta)\\
&= \|\wbf{-}\wbf_{\eta}\|_2 + {\textstyle\EE_{\varepsilonbf\sim\Ncal(\zerobf,\Irm_d\sigma^2_{\eta})}\|\varepsilonbf\|}\\
&=\|\wbf{-}\wbf_{\eta}\|_2 + \sigma_{\eta}\sqrt{2}\frac{\Gamma((d+1)/2)}{\Gamma(d/2)},
\end{align*}
where $\Gamma(\cdot)$ is the gamma function. 
The inequality follows from the triangle inequality, and the last equality follows by the mean of the Chi distribution.
Furthermore, when $\Q$ is a Gaussian distribution, we first use Jensen's inequality to obtain $\W_1(\Q,\eta) \le \W_2(\Q,\eta)$ and then use the closed form solution of the Wasserstein distance (of order 2) between two Gaussian distributions.

\textbf{Learning $\eta$.}
To simplify the optimisation of $\eta$, we define its mean by $\wbf_{\eta}=\lambda\wbf+(1-\lambda)\wbf_{\P}$ and $\sigma^2_{\eta}=\lambda\sigma^2+(1-\lambda)\sigma^2_{\P}$ for the Gaussian case, where $\lambda\in[0,1]$ is a learned parameter.
Note that $\sigma^2_{\eta}$ is not restrained in the Dirac case; hence, it is learned during the optimisation.

\subsection[Experiments]{Experiments\footnote{We introduce more details and additional experiments in \Cref{sec:supp-expe}.}}
\label{sec:setting-result}

\newcommand{\scalesize}{0.81}

\begin{table*}[!ht]
\caption{
Results of the bound minimisation of \Cref{theorem:KL-wass} (see \Cref{eq:alg-KL-wass}) and the ones associated with \citet{amit2022integral} and \citet{maurer2004note}.
``Test'' is the test risk $\Riskhat_{\Tcal}(h_{\wbf})$, ``Bnd'' represents the bound value, ``Wass'' represents the upper bound of the Wasserstein distance multiplied by the Lipschitz constant, and ``KL'' is the KL divergence divided by $2m$.}
\centering
\begin{minipage}{.55\linewidth}
  \centering
  \scalebox{\scalesize}{
  \begin{tabular}{l|c@{\hspace{0.1cm}}c@{\hspace{0.1cm}}@{\hspace{0.1cm}}c@{\hspace{0.1cm}}@{\hspace{0.1cm}}c|c@{\hspace{0.1cm}}@{\hspace{0.1cm}}c@{\hspace{0.1cm}}@{\hspace{0.1cm}}c|}
\toprule
 & \multicolumn{4}{c}{\Cref{theorem:KL-wass}} & \multicolumn{3}{c}{\citet{amit2022integral}} \\
 & Test & Bnd & Wass. & KL & Test & Bnd & Wass. \\
\midrule
\fashion & 0.115 & 0.317 & 0.017 & 0.025 & 0.361 & 1.040 & 0.465 \\
\mnist & 0.077 & 0.294 & 0.018 & 0.027 & 0.304 & 1.078 & 0.583 \\
\mushrooms & 0.026 & 0.190 & 0.009 & 0.015 & 0.498 & 0.614 & 0.012 \\
\phishing & 0.085 & 0.225 & 0.005 & 0.013 & 0.497 & 0.569 & 0.004 \\
\yeast & 0.353 & 0.566 & 0.014 & 0.017 & 0.504 & 1.203 & 0.482 \\
\bottomrule
\end{tabular}
}
  \caption*{(a) Linear models $h_\wbf$ with $\Q=\delta_{\wbf}$}
\end{minipage}%
\begin{minipage}{.4\linewidth}
  \centering
  \scalebox{\scalesize}{
  \begin{tabular}{|c@{\hspace{0.15cm} }c@{\hspace{0.15cm}}c@{\hspace{0.15cm}}c|c@{\hspace{0.15cm}}c@{\hspace{0.15cm}}c|}
\toprule
\multicolumn{4}{c}{\Cref{theorem:KL-wass}} & \multicolumn{3}{c}{\citet{amit2022integral}} \\
Test & Bnd & Wass. & KL & Test & Bnd & Wass. \\
\midrule
0.162 & 0.683 & 0.135 & 0.139 & 0.884 & 3.537 & 7.041 \\
0.111 & 0.673 & 0.152 & 0.158 & 0.776 & 2.414 & 2.680 \\
0.082 & 0.645 & 0.167 & 0.148 & 0.487 & 3.832 & 11.247 \\
0.123 & 0.642 & 0.132 & 0.136 & 0.438 & 1.903 & 2.095 \\
0.335 & 0.707 & 0.061 & 0.059 & 0.383 & 2.469 & 4.318 \\
\bottomrule
\end{tabular}
}
  \caption*{(b) Neural networks $h_\wbf$ with $\Q=\delta_{\wbf}$}
\end{minipage}
\begin{minipage}{.55\linewidth}
  \centering
  \scalebox{\scalesize}{
  \begin{tabular}{l|c@{\hspace{0.1cm}}c@{\hspace{0.1cm}}@{\hspace{0.1cm}}c@{\hspace{0.1cm}}@{\hspace{0.1cm}}c|c@{\hspace{0.1cm}}@{\hspace{0.1cm}}c@{\hspace{0.1cm}}@{\hspace{0.1cm}}c|}
\toprule
 & \multicolumn{4}{c}{\Cref{theorem:KL-wass}} & \multicolumn{3}{c}{\citet{maurer2004note}} \\
 & Test & Bnd & Wass. & KL & Test & Bnd & KL \\
\midrule
\fashion & 0.140 & 0.364 & 0.000 & 0.019 & 0.140 & 0.359 & 0.019 \\
\mnist & 0.097 & 0.338 & 0.000 & 0.021 & 0.096 & 0.333 & 0.021 \\
\mushrooms & 0.017 & 0.265 & 0.000 & 0.022 & 0.035 & 0.258 & 0.016 \\
\phishing & 0.094 & 0.302 & 0.000 & 0.013 & 0.094 & 0.297 & 0.013 \\
\yeast & 0.372 & 0.644 & 0.000 & 0.016 & 0.372 & 0.638 & 0.016 \\
\bottomrule
\end{tabular}
}
  \caption*{(c) Linear models $h_{\wbf}$ with $\Q=\Ncal(\wbf, \sigma^2\Irm_{d})$}
\end{minipage}%
\begin{minipage}{.4\linewidth}
  \centering
  \scalebox{\scalesize}{
  \begin{tabular}{|c@{\hspace{0.15cm} }c@{\hspace{0.15cm}}c@{\hspace{0.15cm}}c|c@{\hspace{0.15cm}}c@{\hspace{0.15cm}}c|}
\toprule
\multicolumn{4}{c}{\Cref{theorem:KL-wass}} & \multicolumn{3}{c}{\citet{maurer2004note}} \\
Test & Bnd & Wass. & KL & Test & Bnd & KL \\
\midrule
0.663 & 1.132 & 0.001 & 0.143 & 0.647 & 1.115 & 0.147 \\
0.610 & 1.105 & 0.001 & 0.160 & 0.623 & 1.105 & 0.154 \\
0.384 & 0.935 & 0.001 & 0.209 & 0.376 & 0.934 & 0.220 \\
0.287 & 0.972 & 0.001 & 0.348 & 0.243 & 0.813 & 0.232 \\
0.648 & 2.295 & 0.003 & 2.342 & 0.652 & 2.335 & 2.472 \\
\bottomrule
\end{tabular}
}
  \caption*{(d) Neural networks $h_{\wbf}$ with $\Q=\Ncal(\wbf, \sigma^2\Irm_{d})$}
\end{minipage}
\label{table:expe}
\end{table*}

In this section, we propose to evaluate our algorithm on linear classifiers and neural networks on various datasets.
To do so, we follow the setting of \citet{viallard2023learning}.
\\
\textbf{Setting.}
We stand in the classification setting where the data space is $\Zcal=\Xcal{\times}\Ycal$, with $\Xcal=\{ \xbf \in \R^n | \|\xbf\|_2 \le 1 \}$ the input space and $\Ycal=\{1,\dots,|\Ycal|\}$ the label space.
Hence, our goal is to learn a model $h_{\wbf}: \Xcal\to \R^{|\Y|}$ that outputs scores given  $\x\in\Xcal$; we denote by $h_{\wbf}(\x)[y']\in\R$ the score for each label $y'$.
Based on this score, we define our $\alpha$-Lipschitz loss $\loss(h, (\x, y)) = \frac{1}{|\Y|}\sum_{y'\ne y} \max(0, 1{-} \alpha(h[y]{-}h[y']))$ (\wrt $h[1],\dots,h[|\Y|]$).
We set $\alpha=25$ for the linear models and $\alpha=250$ for the neural networks.
The confidence parameter is fixed to $\delta=0.05$.
\\
\textbf{Datasets.} 
We consider 3 datasets (\yeast, \phishing, \mushrooms) from the UCI repository \citep{dua2017uci}, \mnist \citep{lecun1998mnist} and \fashion \citep{xiao2017fashion}.
Moreover, for \mnist and \fashion, we keep the original training $\Scal$ and test set $\Tcal$, while for the UCI datasets, we perform a 50\%/50\% split.
We denote by $\Riskhat{\Tcal}(h)$, the test risk of $h$ (computed on the test set $\Tcal$) to estimate the population risk $\Risk_{\Dcal}(h)$.
\\
\textbf{Optimisation.} 
To solve all our optimisation problems, we use the COCOB-Backprop optimiser \citep{orabona2017training}; its parameter is fixed to $10$. 
Moreover, for each optimisation, we optimise (with a batch size of $256$) for at least $10000$ iterations (and finish the epoch when the number of iterations is attained).
\\
\textbf{Models.} 
We instantiate the algorithm described in \Cref{eq:alg-KL-wass} for linear classifiers and neural networks.
We first define a linear model by $h_{\wbf}(\xbf)\defeq W\xbf+b$, where the weight matrix and the bias are respectively defined by $W\in\R^{|\Ycal|\times n}$ and $b\in\R^{|\Ycal|}$; we denote by $\wbf=\vect(\{W, b\})$ the vectorisation of all the parameters.
We know from \citet[Lemma 8]{viallard2023learning} that the loss is $\sqrt{2}\alpha$-Lipschitz \wrt the parameters $\wbf$.
The initialisation of $W$ and $b$ is done with zeros. 
For neural networks, we define the model by $h_{\wbf}(\x) \defeq Wh^{K}(\cdots h^{1}(\x))+b$, which is composed of $K$ layers $h^1(\cdot),\dots,h^K(\cdot)$.
Similarly to linear models, $W\in\R^{|\Ycal|\times N}$ and $b\in\R^{N}$ are the weight matrix and the bias of the last layer.
Moreover, the $i$-th layer $h^{i}$, composed of $N$ nodes, is defined by $h^{i}(\xbf)\defeq\proj(\leaky(W_i \xbf + b_i))$, where $W_i\in\R^{N\times N}$ and the bias $b_i\in \R^{N}$ are its weight matrix and bias respectively; $\leaky : \R^{N} \to \R^{N} $ is the Leaky ReLU applied element-wise and $\proj$ project the vector $\leaky(W_i \xbf + b_i)$ in the unit $\ell_2$-ball.
The weights $\wbf=\vect(\{W, W_{K}, \dots, W_1, b, b_{K}, \dots, b_1\})$ represent the vectorisation of all parameters of the network.
Note that \citet[Lemma 9]{viallard2023learning} show the loss is Lipschitz \wrt the parameters $\wbf$. 
However, the Lipschitz constant is not explicit, thus, we provide it in \Cref{lemma:neural-lipschitz}. 
Following \citet{viallard2023learning}, each parameter of the matrices is {\it (i)} initialised with a Gaussian distribution centered on zero and with a standard deviation of $0.04$ and {\it (ii)} clipped between $-0.08$ and $+0.08$.
Each parameter in the biases is initialised with a zero except the parameter in $b_1$, which is set to $0.1$.
Lastly, we further set $N=600$ and $K=1$.
\\
\textbf{Empirical findings.}
We show in \Cref{table:expe} the results of the bound's minimisation; for more details, we refer the reader to \Cref{sec:supp-expe}.
As we can remark, the behaviour of our \Cref{theorem:KL-wass}'s bound differs when we consider a Dirac or a Gaussian posterior distribution $\Q$.
For instance, in the Dirac case, our bounds and test risks are lower than the one of \citet{amit2022integral}.
This is due to the fact that the minimisation of our bound allows us to learn $\eta$ that minimises better the KL divergence and the Wasserstein distance.
However, in the Gaussian case, the bound values and the test risks are similar, illustrating that when $\Q$, $\P$, and $\eta$ are Gaussians, interpolating between the Wasserstein and the KL does not bring any advantage. Indeed, our algorithm puts almost no weight on the Wasserstein here and the value of the KL divergence is then similar with the algorithm based on the bound of \citet{maurer2004note}.
Hence, in this case, \Cref{theorem:KL-wass}'s bound becomes the same as the one of \citet{maurer2004note} since $\Q=\eta$. In any case, results show that interpolating KL and Wasserstein brings the best-of-both worlds with increased performance \wrt Wasserstein method for Diracs and equivalent performance \wrt KL method for the Gaussian case.

\section{Conclusion}

We derived novel PAC-Bayes bounds based on $(f, \Gamma)$-divergences.
We show its theoretical interest for analysing the generalisation of SGD and retrieving Rademacher-based bounds.
We also provide non-trivial generalisation bounds for two models and bring tighter bounds in the Dirac case. 

\section*{Broader Impact Statement}
This paper explores a novel strategy to understand and analyse theoretical properties of machine learning algorithms, and notably generalisation.
As such, we do not anticipate any immediate or longer-term negative societal impact -- however, we believe a better theoretical understanding of machine learning ultimately contributes to a more virtuous use and deployment of AI systems.

\section*{Acknowledgements}

Paul Viallard and Umut {\c S}im{\c s}ekli are partially supported by the French program ``Investissements d’avenir'' ANR-19-P3IA-0001 (PRAIRIE 3IA Institute).
Umut {\c S}im{\c s}ekli is also supported
by the European Research Council Starting Grant DYNASTY – 101039676.
Benjamin Guedj acknowledges partial support from the French Research Agency through the programme ``France 2030'' and PEPR IA on grant SHARP ANR-23-PEIA-0008.

\bibliography{main}
\bibliographystyle{icml2024}

\newpage
\appendix
\notappendixfalse
\onecolumn

The appendix is organised as follows: 
\begin{enumerate}
    \item We provide supplementary background on $\alpha$-stable Lévy processes in \Cref{sec:supp-background} ;
    \item \Cref{sec:supp-results} introduces additional theoretical results.
    More precisely, we introduce additional generic generalisation bounds in \Cref{sec:improv-existing} and we provide another application of \Cref{theorem:KL-wass} for heavy-tailed SGD in \Cref{sec:supp-sgd} ;
    \item \Cref{sec:proofs} contains all the postponed proofs ;
    \item Additional details on the experiments are gathered in \Cref{sec:supp-expe}.
\end{enumerate}

\section{Supplementary Background on $\alpha$-stable Lévy Processes}
\label{sec:supp-background}

We first define $\alpha$-stable Lévy processes properly, re-using the definition of \citet{simsekli2019tail}.
\begin{definition}[Symmetric $\alpha$-stable Lévy process]
    A symmetric $\alpha$-stable Lévy process in dimension $d$, denoted as $\mathrm{L}_t^\alpha$), is constituted of independent components 
    (meaning each component of $\mathrm{L}_t^\alpha$ is an independent $\alpha$-stable Lévy motion in $\mathbb{R}$). 
    For the scalar case, it is defined as follows for $\alpha \in(0,2]$ :
    \begin{itemize}
        \item $\mathrm{L}_0^\alpha=0$ almost surely.
        \item For $t_0<t_1<\cdots<t_N$, the increments $(\mathrm{L}_{t_i}^\alpha-\mathrm{L}_{t_{i-1}}^\alpha)$ are independent (for $i=1, \dots, N$).
        \item The difference $\left(\mathrm{L}_t^\alpha-\mathrm{L}_s^\alpha\right)$ and $\mathrm{L}_{t-s}^\alpha$ follow the distribution $\mathcal{S} \alpha \mathcal{S}\left((t-s)^{1 / \alpha}\right)$ for $s<t$.
        Here, $\S\alpha\S$ distributions are defined through their characteristic function via $X\sim \S\alpha\S \Leftrightarrow \EE[\exp(i\omega X)] = \exp(-\sigma\vert \sigma\omega\vert^\alpha)$. 
        \item $\mathrm{L}_t^\alpha$ is continuous in probability (it has stochastically continuous sample paths), \ie, for all $\delta>0$ and $s \geq 0$, we have $p\left(\left|\mathrm{~L}_t^\alpha-\mathrm{L}_s^\alpha\right|>\delta\right) \rightarrow 0$ as $t \rightarrow s$.
    \end{itemize}

Moreover, when $\alpha=2$, we know that $\mathrm{~L}_t^\alpha$ coincides with a scaled version of Brownian motion, $\sqrt{2}\mathrm{~B}_t$. 
\end{definition}

\section{Supplementary Results}
\label{sec:supp-results}

\subsection{Improving on existing PAC-Bayes bounds using $(f,\Gamma)$-divergences}
\label{sec:improv-existing}

This section introduces the claims made below \Cref{theorem:KL-wass}: it is possible to improve on many bounds in the literature at the cost of a Lipschitz bounded assumption by using the same proof technique as \Cref{theorem:KL-wass}. 
We state and prove those results.

\begin{theorem}
Assume that $\ell \in [0,1]$.
Assume that, for any $\delta'\in (0,1)$, with probability $1-\delta'$, the loss $\ell(,\z)$ is $L(m,\delta)$-Lipschitz for all $\z\in\Z$.
Thus, for any data-free prior $\P$, any $\lambda>0$, with probability at least $1-\delta$ over $\Scal\sim\D^m$, we have for all $\Q\in\Pcal(\H)$, any $\eta\in\Pcal(\H)$,

\text{(i) -- Catoni's bound:}
\begin{align*}
\EE_{h\sim\Q} \Risk_{\D}(h)-\Risk_{\S}(h) & \le 2L(m,\nicefrac{\delta}{2})\W(\Q,\eta)+ \frac{\KL(\eta,\P)+ \log\LP \frac{2}{\delta} \RP}{\lambda} + \frac{\lambda}{2m};
\end{align*}
\text{(ii) -- Supermartingale bound:}
\begin{align*}
\EE_{h\sim\Q} \Risk_{\D}(h)-\Risk_{\S}(h) & \le \LP 2 + \lambda\RP L(m,\nicefrac{\delta}{2})\W(\Q,\eta) + \frac{\KL(\eta,\P)+ \log\LP \frac{2}{\delta} \RP}{\lambda} + \frac{\lambda}{2}\EE_{h\sim\Q}\EE_{z\sim\D}[\ell^2(h,z)];
\end{align*}
\text{(iii) -- Catoni's bound with fast rate:}
\begin{align*}
\EE_{h\sim\Q}\Risk_{\D}(h) & \le \frac{1}{1-\frac{\lambda}{2}}\LP\EE_{h\sim \Q}\Riskhat_\S(h) + \LP 2 + \lambda\RP L\LP m,\nicefrac{\delta}{2} \RP \W(\Q,\eta) + \frac{\KL(\eta,\P)+ \log\LP \frac{2}{\delta} \RP}{\lambda m}\RP.
\end{align*}
\end{theorem}
Under our assumptions, Catoni's bound is an improvement of \citet[Theorem 4.1]{alquier2016properties}, the supermartingale bounds improves \citet{haddouche2023pac,viallard2023learning}, and the Catoni's fast rate bound improves on \citet[Theorem 2]{mcallester2013pacbayesian}.
\begin{proof}
We first start from \Cref{eq:proof-theorem-KL-1} in \Cref{theorem:KL-wass}'s proof alongside with the infimal convolution formula \eqref{eq:inf-conv-kl-wass}.
Assume that $\phi_\S\in \Lip_b^{\alpha(m,\delta)}$ with probability at least $1-\delta/2$. 
Then, with probability at least $1-\delta$, for any $\Q,\eta$ we have
\begin{align}
\EE_{h \sim \Q}\LB\phi_\S(h)\RB \le  \alpha(m,\delta)\W_1(\Q, \eta) + \KL(\eta\|\P) {+}\ln\!\frac{2}{\delta}{+}\ln\!\LB\EE_{h\sim\P}\EE_{\Scal\sim\D^m}e^{\phi_\S(h)}\RB.\label{eq:improve}
\end{align}
Note that we use that $\P$ was data-free to swap the integrals in the last term.

\textbf{Catoni's bound.}
We use \eqref{eq:improve} with $\phi_\S= \lambda(\Risk_\D-\Riskhat_\S)$, and $\alpha(m,\delta)= \lambda L(m,\nicefrac{\delta}{2})$, then we have, with probability at least $1-\delta$, for any $\Q,\eta$, 
\begin{align*}
\EE_{h \sim \Q}\LB\Risk_\D- \Riskhat_\S(h)\RB \le  L(m,\nicefrac{\delta}{2})\W_1(\Q, \eta) + \frac{\KL(\eta\|\P) {+}\ln\!\frac{2}{\delta}}{\lambda}{+} \frac{1}{\lambda}\ln\!\LB\EE_{h\sim\P}\EE_{\Scal\sim\D^m}e^{\lambda \Delta_\S(h)}\RB .
\end{align*}
As $\S$ is \iid, applying Hoeffding's lemma $m$ times on the random variables $\Risk_D(h)-\ell(h,\z_i)\in [-1,1]$, gives that for any $h$, the inequality $\EE_{\Scal\sim\D^m}e^{\lambda \Risk_\D- \Riskhat_\S(h)}\leq \frac{\lambda^2}{2m}$.
This concludes the proof.

\textbf{Supermartingale bound.} We use \Cref{eq:improve} with $\phi_\S(h)= m\lambda\Risk_\D(h)- \Riskhat_\S(h) - m\frac{\lambda^2}{2}\EE_{\z\sim\D}[\ell(h,\z)^2]$, and $\alpha(m,\delta)= 2m\LP\lambda + \frac{\lambda^2}{2}\RP L(m,\nicefrac{\delta}{2})$.
Indeed, because $\ell$ is $L_1\defeq L(m,\nicefrac{\delta}{2})$ with probability at least $1-\delta/2$, $\Risk_\D- \Riskhat_\S$ is $2L_1$-Lipschitz and $h\rightarrow\EE_{\z\sim\D}[\ell(h,\z)^2]$ is $2L_1$ Lipschitz.
Then, applying \eqref{eq:improve} and dividing by $m\lambda$ gives, with probability at least $1-\delta$, for any $\Q,\eta$, 
\begin{align*}
    \EE_{h \sim \Q}\LB\Risk_\D(h)- \Riskhat\S(h)\RB \le  \LP 2 + \lambda\RP L(m,\nicefrac{\delta}{2})\W(\Q,\eta) + \frac{\KL(\eta\|\P) {+}\ln\!\frac{2}{\delta}}{m\lambda}{+} \frac{1}{m\lambda}\ln\!\LB\EE_{h\sim\P}\EE_{\Scal\sim\D^m}e^{\phi_\S(h)}\RB + \frac{\lambda}{2}\EE_{h\sim\Q}\EE_{z\sim\D}[\ell^2(h,z)].
\end{align*}
Then, \citet{chugg2023unified} proved in their corollary 4.8 that $\EE_{h\sim\P}[e^{\phi_\S(h)}]$ is a supermartingale with respect to the data $\z_i,i\geq 1$ and an adapted filtration.
In particular, because $\ell$ is non-negative,  $\EE_{\Scal\sim\D^m} \EE_{h\sim\P}e^{\phi_\S(h)} \leq 1$. 
This concludes the proof. 

\textbf{Catoni's fast rate.}
We start from the supermartingale bound and notice that, because the loss lies in $[0;1], \ell^2\leq \ell$.
Upper-bounding the last term of the supermartingale bound and re-organising the terms conclude the proof.
\end{proof}

\subsection{Additional result for heavy-tailed SGD: going beyond continuous processes.}
\label{sec:supp-sgd}

In a similar spirit to \Cref{sec:ht-sgd}, we show that it is possible to provide sound generalisation bounds for heavy-tailed SGD when discrete modelisation is involved instead of continuous SDEs.  

\textbf{Modelling heavy-tailed SGD beyond continuous processes.}
\Cref{theorem:heavy-tailed-sgd} gives quantitative results on the generalisation ability of the asymptotic heavy-tailed distribution $\Q_\alpha$ and thus fills a gap between PAC-Bayes learning and continuous approximations of SGD. 
However, $\Q_\alpha$ remains mainly theoretical, as it is the continuous approximation of a discrete optimisation process. 
Thus, in order to get closer to practical optimisation, we modelise the heavy-tailed behaviour of SGD by a multivariate $p$-Student distribution with parameters $\mu, \Sigma$.
Recall that a multivariate $p$-Student variable $y$ over $\Rbb^d$ with parameters $\mu,\Sigma$ (namely $Stud_p(\mu,\Sigma)$) and can be written as $y= \mu +  x\sqrt{\nicefrac{p}{u}}$ where $x\sim\Ncal(0_{\Rbb^d},\Sigma)$ and $u\sim \chi^2_p$ and $x,u$ are independent.
The heavy-tailed behaviour of $y$ is determined by $p$; note that $y$ has a mean (equal to $\mu$) only if $p>1$ and a covariance equal to $\nicefrac{p}{p-2}\Sigma$ only if $p>2$.
As $p$ goes to infinity, the student distribution converges to $\Ncal(\mu,\Sigma)$.
Such a process can then be seen as a discrete approximation of an $\alpha$ stable Levy process with the rescaling $p= \nicefrac{\alpha}{2-\alpha}$.
For practical instantiations, we can assume, for instance, that $\mu$ is the averaged predictor over a few runs of SGD and $\Sigma\defeq \sigma \Irm_d$ is a fixed level of noise. 

Following the idea of \Cref{theorem:heavy-tailed-sgd}'s proof, we exploit \Cref{theorem:KL-wass} in order to bound the generalisation ability of multivariate $p$-Student distribution.
The resulting upper bound lies in \Cref{theorem:student} and is tractable in practice. 

\begin{restatable}{theorem}{studentbound}
    \label{theorem:student}
    Let $\sigma>0,\mu_0\in\mathbb{R}^d$ and $\P= \Ncal(\mu_0, \sigma\Irm_d)$. Assume $\ell \in [0,1]$ and that, for any $\delta'\in (0,1)$, with probability $1-\delta'$, $h\rightarrow\Delta_\S^2(h)$ is $L(m,\delta')$-Lipschitz.
    Thus, with probability at least $1-\delta$ over $\Scal\sim\D^m$, we have, for any multivariate student $Stud_p(\mu,\sigma^2\mathrm{Id})$, with $p>1, \mu \in \Rbb^d$,
    \ifnotappendix
    \begin{multline*}
    \EE_{h\sim\Q_\alpha}\Delta_{\S}(h) \le \\
    \sqrt{ L(m,\nicefrac{\delta}{2}) \sigma f(p,d)  + \frac{\|\mu-\mu_0\|^2}{2\sigma m} +\frac{\ln\frac{4\sqrt{m}}{\delta}}{2m}},
    \end{multline*}
    \else
    \begin{align*}
    \EE_{h\sim\Q_\alpha}\Delta_{\S}(h) \le 
    \sqrt{ L(m,\nicefrac{\delta}{2}) \sigma f(p,d)  + \frac{\|\mu-\mu_0\|^2}{2\sigma m} +\frac{\ln\frac{4\sqrt{m}}{\delta}}{2m}},
    \end{align*}
    \fi
    where $f(p,d)\defeq\sqrt{d}\EE\LB \LN \sqrt{\frac{p}{u}}-1 \RN \RB $ with $u\sim \chi^2_p$.
\end{restatable}
\begin{proof}
We start from \Cref{theorem:KL-wass}.
We have with probability at least $1-\delta$, for any $\eta,\Q$: 
\begin{align*}
\EE_{h\sim\Q}\Delta_{\S}(h) \le \sqrt{ L(m,\nicefrac{\delta}{2})\W_1(\Q, \eta) + \frac{\KL(\eta\|\P)+ \ln\frac{4\sqrt{m}}{\delta}}{2m}}.
\end{align*}
We recall that $\P=\Ncal(\mu_0,\sigma^2\Irm_d)$, we then pick $\eta= \Ncal(\mu,\sigma\Irm_d)$ and $\Q= Stud_p(\mu,\sigma\Irm_d)$.
We then know that $\KL(\eta,\P)= \frac{\|\mu-\mu_0\|^2}{\sigma}$.
The only thing left to control is $W_1(\rho,\eta)$.
To do so, we exploit the following definition of the Wasserstein distance coming from optimal transport: 
\begin{align*}
\W_1(\Q,\eta) = \inf_{\gamma \in \Gamma(\Q,\eta)} \EE_{(Y_1,Y_2)\sim \gamma} \LP \|Y_1-Y_2\| \RP,
\end{align*}
where $\Gamma(\Q,\eta)$ is the set of all distributions over $\H\times \H$ such that the marginal distribution of $(Y_1,Y_2)\sim \gamma$ are respectively $\Q$ and $\eta$.
We then consider the coupling $\gamma$ such that $Y,Y_2\sim \gamma \leftrightarrow (Y_1,Y_2)= \LP \mu + \sigma Z \sqrt{\frac{p}{U}}, \mu+ \sigma Z \RP$ with $Z\sim \Ncal(0,\Irm_d), U\sim \chi^2_p$ and $U,Z$ are mutually independent.
Then, $\gamma\in \Gamma(\Q,\eta)$ and we have
\begin{align*}
    \W_1(\Q,\eta) & \leq \EE_{(Y_1,Y_2)\sim \gamma} \LP \|Y_1-Y_2\| \RP \\
    & =\EE_{Z\sim \Ncal(0,\Irm_d)} \EE_{U\sim \chi^2_p} \left\| \mu + \sigma Z \sqrt{\frac{p}{U}} - \mu -\sigma Z \right\|\\
    & = \sigma \EE_{Z\sim \Ncal(0,\Irm_d)} \EE_{U\sim \chi^2_p} \| Z\| \LN \sqrt{\frac{p}{U}} -1 \RN \\
    & \leq \sigma\sqrt{d}  \EE_{U\sim \chi^2_p} \| Z\| \LN \sqrt{\frac{p}{U}} -1 \RN,
\end{align*}
the last line holding thanks to the independence of $Z$ and $U$ and because $\EE_{Z\sim \Ncal(0,\Irm_d)} [\|Z\|] \leq \sqrt{d}$.
Plugging this in the bound above concludes the proof.
\end{proof}

\section{Postponed Proofs}
\label{sec:proofs}

\subsection{Proof of \Cref{theorem:pb-f-gamma}}
\theorempbfgamma*
\begin{proof}
From \Cref{def:f-gamma-divergence}, we can deduce that we have for all $\Q\in\Pcal(\H)$
\begin{align*}
\EE_{h\sim\Q}\phi_{\Scal}(h) \le D^{\Gamma}_{f}(\Q\|\P) + \Lambda^{\P}_{f}(\phi_\Scal).
\end{align*}
Moreover, we have 
\begin{align*}
\Lambda^{\P}_{f}(\phi_\Scal) = \ln\LB \exp(\Lambda^{\P}_{f}(\phi_\Scal))\RB
\end{align*}
and since $\exp(\Lambda^{\P}_{f}(\phi_\Scal))>0$, we can apply Markov's inequality to have with probability at least $1-\delta$ over $\Scal\sim\D^m$
\begin{align*}
\exp(\Lambda^{\P}_{f}(\phi_\Scal)) &\le \frac{1}{\delta}\EE_{\Scal\sim\D^m}\exp(\Lambda^{\P}_{f}(\phi_\Scal))\\
\iff \ln\LB\exp(\Lambda^{\P}_{f}(\phi_\Scal))\RB &\le \ln\!\frac{1}{\delta}{+}\ln\!\LB\EE_{\Scal\sim\D^m}\exp\LP\Lambda^{\P}_{f}(\phi_\Scal)\RP\RB.
\end{align*}
\end{proof}

\subsection{Proof of \Cref{theorem:pb-f-gamma-mcdiarmid}}

\theorempbfgammamcdiarmid*
\begin{proof}
From \Cref{def:f-gamma-divergence} and the upper bound on $\Lambda_f^\P$, we can deduce that we have for all $\Q\in\Pcal(\H)$
\begin{align*}
\EE_{h\sim\Q}\phi_{\Scal}(h) \le D^{\Gamma}_{f}(\Q\|\P) + B^{\P}(\phi_\Scal).
\end{align*}
The bounded-difference property alongside McDiarmid's inequality gives, with probability at least $1-\delta$,
\begin{align*}
\Lambda^{\P}_{f}(\phi_\Scal) \le \EE_{\Scal\sim\D^m}B^{\P}(\phi_\Scal) + {\textstyle\sqrt{\frac{\ln(\frac{1}{\delta})}{2}\sum_{i=1}^{m}c_i^2}}.
\end{align*}
Combining those two equations yields the desired result.
\end{proof}
\subsection{Proof of \Cref{theorem:KL-wass}}
\label{sec:proof-KL-wass}
\theoremKLgamma*
\begin{proof}
The proof is split into three steps.

\textbf{Step 1: finding the $f$-divergence.}
It is known that the $f$-divergence with $f: x\mapsto x\ln x$ is
\begin{align*}
D_{f}(\Q\|\P) \defeq \EE_{g\sim\P}f\LP\frac{d\Q}{d\P}(g)\RP = \EE_{g\sim\P}\frac{d\Q}{d\P}(g)\ln\LP\frac{d\Q}{d\P}(g)\RP = \EE_{g\sim\Q}\ln\LP\frac{d\Q}{d\P}(g)\RP \defeq \KL(\Q\|\P).
\end{align*}

\textbf{Step 2: finding the closed-form solution of $\Lambda^{\P}_{f}(\varphi)$.}
It is known for the KL divergence.
However, for the sake of completeness, we provide a proof.
Recall that with $f: x\mapsto x\ln x$, we have
\begin{align*}
f^{*}(y) \defeq \sup_{x\in\R}\LC yx-f(x)\RC = \sup_{x\in\R}\LC g(x,y) \RC, \quad\text{where}\quad g(x,y)\defeq yx-x\ln x.
\end{align*}
In order to find the supremum (which is attained), we can find the derivative of the function $g$, and we have
\begin{align*}
\frac{\partial g}{\partial x}(x, y) = y -\ln x -1.
\end{align*}
Since the Legendre transform is convex, we can set the derivative to $0$ to obtain
\begin{align*}
\frac{\partial g}{\partial x}(x, y) = 0 \iff y -\ln x -1 = 0 \iff x=e^{y-1}.
\end{align*}
Hence, we can deduce that we have
\begin{align*}
f^{*}(y)=g(e^{y-1}, y) = ye^{y-1}-e^{y-1}\ln e^{y-1} = ye^{y-1}-ye^{y-1}+e^{y-1} = e^{y-1}.
\end{align*}
Now, the goal is to find the value $c$ associated with $\Lambda^{\P}_{f}(\varphi) \defeq \inf_{c\in\R}\LC c + \EE_{g\sim\P}f^{*}(\varphi(g)-c)\RC = \inf_{c\in\R}\LC c + \EE_{g\sim\P}e^{\varphi(g)-c-1}\RC$. To do so, we set $c=\ln\EE_{g\sim\P}e^{\varphi(g)}-c'$ with $c'\in\R$, and we show that $\Lambda^{\P}_{f}(\varphi)$ is optimal with $c'=1$.
We have
\begin{align*}
c + \EE_{g\sim\P}f^{*}(\varphi(g)-c) = \ln\EE_{g\sim\P}e^{\varphi(g)}-c' + \EE_{g\sim\P}e^{\varphi(g)-\ln\EE_{g\sim\P}e^{\varphi(g)}+c'-1} = \ln\EE_{g\sim\P}e^{\varphi(g)} - c' + e^{c'-1} \defeq g'(c')
\end{align*}
Moreover, the derivative $g'(c')$ with respect to $c'$ is given by 
\begin{align*}
\frac{\partial g'}{\partial c'}(c') = e^{c'-1}-1.
\end{align*}
Then, notice that this derivative on $(-\infty, 1[$ and non-negative on $[1,+\infty)$, thus the minimum of $g'$ is reached for $c'=1$. We then deduce that 
\begin{align*}
\Lambda^{\P}_{f}(\varphi) = \ln\EE_{g\sim\P}e^{\phi(g)}.
\end{align*}

\textbf{Step 3: deriving the PAC-Bayesian bound.} 
We use \Cref{theorem:pb-f-gamma} with $f=f_{\KL}$ and $\Gamma= Lip_b^{\alpha(m,\delta)}$ with $\alpha(m,\delta)= 2mL(m,\delta/2)$. Then, for any $\phi_\S\in Lip_b^{\alpha(m,\delta)}$, we have, with probability at least $1-\delta/2$,  
\begin{align}
\EE_{h\sim\Q}\phi_{\Scal}(h) &\le D^{\Gamma}_{f}(\Q\|\P){+}\ln\!\frac{2}{\delta}{+}\ln\!\LB\EE_{\Scal\sim\D^m}\exp\LP\Lambda^{\P}_{f}(\phi_\Scal)\RP\RB\nonumber\\
&=  D^{\Gamma}_{f}(\Q\|\P){+}\ln\!\frac{2}{\delta}{+}\ln\!\LB\EE_{\Scal\sim\D^m}\EE_{h\sim\P}e^{\phi_{\Scal}(h)}\RB.\label{eq:proof-theorem-KL-1}
\end{align}
Let $\phi_{\Scal}(h) = 2m\Delta_\S^2$. Note that, thanks to our assumption, we know that with probability at least $1-\nicefrac{\delta}{2}$, $\Delta_\S^2$ is $L(m,\nicefrac{\delta}{2})$-Lipschitz, then $\phi_\S$ is $\alpha(m,\delta)$-Lipschitz. We then take a union bound, apply \Cref{eq:inf-conv-kl-wass}, and rearrange the terms to obtain, with probability at least $1-\delta$:
\begin{align*}
\EE_{h \sim \Q}\LB\Delta_\S(h)^2\RB \le \frac{1}{2m}\!\LB \inf_{\eta\in\Mcal_1(\H)}\LC \alpha(m,\delta)\W_1(\Q, \eta) + \KL(\eta\|\P) \RC{+}\ln\!\frac{2}{\delta}{+}\ln\!\LB\EE_{\Scal\sim\D^m}\EE_{h\sim\P}e^{2m\Delta_\S(h)^2}\RB \RB.
\end{align*}
From Jensen's inequality, we have $\LP \EE_{h \sim \Q}\vert\Risk_{\D}(h){-}\Risk_{\Scal}(h)\vert \RP^2 \le \EE_{h \sim \Q}\LB\Delta_\S(h)^2\RB$ and we can deduce that for any $\eta \in\Mcal_1(\Hcal)$,
\begin{align*}
\vert\EE_{h \sim \Q}\LB\Delta_\S(h)\RB \vert \le \sqrt{\frac{1}{2m}\!\LB \alpha(m,\delta)\W_1(\Q, \eta) + \KL(\eta\|\P){+}\ln\!\frac{2}{\delta}{+}\ln\!\LB\EE_{\Scal\sim\D^m}\EE_{h\sim\P}e^{2m\Delta_\S(h)^2}\RB \RB}.
\end{align*}
From Pinsker's inequality and \citet{maurer2004note}, we have
\begin{align*}
\ln\!\LB\EE_{\Scal\sim\D^m}\EE_{h\sim\P}e^{2m\Delta_\S(h)^2}\RB \le \ln\!\LB\EE_{\Scal\sim\D^m}\EE_{h\sim\P}e^{m\kl\LP\Risk_{\D}(h)\|\Risk_{\Scal}(h)\RP}\RB \le \ln\LP2\sqrt{m}\RP,
\end{align*}
where $\kl(a\|b)=a\ln\frac{a}{b}+(1{-}a)\ln\frac{1-a}{1-b}$.

Re-organising the terms alongisde with $\alpha(m,\delta)= 2mL(m,\nicefrac{\delta}{2})$ concludes the proof.
\end{proof}

\subsection{Proof of \Cref{theorem:KL-reverse-hellinger-gamma}}
\label{sec:proof-rev-KL-Hell}
We split the proof in half, one for each equation. 
We restate below \Cref{theorem:KL-reverse-hellinger-gamma} for completeness.
\theoremKLreverseHellgamma*

\subsubsection{Proof of \Cref{eq:reverse-KL}}
\begin{proof}
The proof is split into three steps.

\textbf{Step 1: finding the $f$-divergence.}
It is known that the $f$-divergence with $f: x\mapsto -\ln x$ is
\begin{align*}
D_{f}(\Q\|\P) \defeq \EE_{g\sim\P}f\LP\frac{d\Q}{d\P}(g)\RP = -\EE_{g\sim\P}\ln\LP\frac{d\Q}{d\P}(g)\RP \defeq \KLr(\Q\|\P).
\end{align*}

\textbf{Step 2: finding an upper bound $B^{\P}(\varphi)$ of $\Lambda^{\P}_{f}(\varphi)$.}
Recall that with $f: x\mapsto -\ln x$, we have
\begin{align*}
f^{*}(y) \defeq \sup_{x\in\R}\LC yx-f(x)\RC = \sup_{x\in\R}\LC g(x,y) \RC, \quad\text{where}\quad g(x,y)\defeq yx+\ln x.
\end{align*}
In order to find the supremum (which is attained), we can find the derivative of the function $g$, and we have
\begin{align*}
\frac{\partial g}{\partial x}(x, y) = y+\frac{1}{x}.
\end{align*}
Since the Legendre transform is convex, we can set the derivative to $0$ to obtain
\begin{align*}
\frac{\partial g}{\partial x}(x, y) = 0 \iff y +\frac{1}{x} = 0 \iff x=-\frac{1}{y}.
\end{align*}
Hence, we can deduce that we have
\begin{align*}
f^{*}(y)=g\LP-\frac{1}{y}, y\RP = -y\frac{1}{y}+\ln\LP-\frac{1}{y}\RP = \ln\LP-\frac{1}{y}\RP-1.
\end{align*}
Now, we upper-bound 
\begin{align*}
\Lambda^{\P}_{f}(\varphi) &\defeq \inf_{c\in\R}\LC c + \EE_{g\sim\P}f^{*}(\varphi(g)-c)\RC\\
&= \inf_{c\in\R}\LC c + \EE_{g\sim\P}\ln\LP-\frac{1}{\varphi(g)-c}\RP{-}1\RC\\
&\le \EE_{g\sim\P}\LB 1 + \ln\LP-\frac{1}{\varphi(g){-}1}\RP-1\RB\\
&=  \EE_{g\sim\P}\LB-\ln\LP1-\varphi(g)\RP\RB \defeq B^\P(\varphi).
\end{align*}

\textbf{Step 3: deriving the PAC-Bayesian bound.} 
We use \Cref{theorem:pb-f-gamma-mcdiarmid} with $B^\P(\varphi)=\EE_{g\sim\P}\LB-\ln\LP1-\varphi(g)\RP\RB$ and $\phi_{\Scal}(h) = \frac{1}{2}\vert\Risk_{\D}(h){-}\Risk_{\Scal}(h)\vert$ to obtain the following inequality holding with probability at least $1-\delta$ over $\Scal\sim\D^m$

\begin{align}
\EE_{h \sim \Q}\vert\Risk_{\D}(h){-}\Risk_{\Scal}(h)\vert \le 2\W^{\Gamma}(\Q, \eta) + 2\KLr(\eta\|\P) + 2\EE_{\Scal\sim\D^m}\EE_{h\sim\P}\LB{-}\ln\LP1{-}\frac{1}{2}\vert\Risk_{\D}(h){-}\Risk_{\Scal}(h)\vert\RP\RB + 2\sqrt{\frac{\ln\frac{1}{\delta}}{2}\sum_{i=1}^{m}c_i^2}.\label{eq:proof-theorem-KL-reverse-1}
\end{align}

The final step is to find the upper bound $c_i$ for each $i\in\{1,\dots, m\}$.
To do so, we first upper-bound the difference $|B^{\P}(\phi_\Scal) - B^{\P}(\phi_{\S_i'})|$ for all $i\in\{1,\dots, m\}$.
We first have
\begin{align*}
\left\vert B^{\P}(\phi_\Scal) - B^{\P}(\phi_{\S_i'})\right\vert &= \left\vert\EE_{h\sim\P}\LB{-}\ln\LP1{-}\phi_{\Scal}(h)\RP\RB-\EE_{h\sim\P}\LB{-}\ln\LP1{-}\phi_{\Scal_i'}(h)\RP\RB\right\vert\\
&= \left\vert \EE_{h\sim\P}\LB \ln\LP1{-}\phi_{\Scal_i'}(h)\RP -\ln\LP1{-}\phi_{\Scal}(h)\RP\RB \right\vert\\
&= \left\vert \EE_{h\sim\P} \ln\LP\frac{1{-}\phi_{\Scal_i'}(h)}{1{-}\phi_{\Scal}(h)}\RP \right\vert.
\end{align*}

Now, in order to upper-bound $\vert \EE_{h\sim\P} \ln(1{-}\phi_{\Scal_i'}(h))-\ln(1{-}\phi_{\Scal}(h))\vert$, the goal is to upper-bound $\EE_{h\sim\P} \ln(1{-}\phi_{\Scal_i'}(h))-\ln(1{-}\phi_{\Scal}(h))$ and  $\EE_{h\sim\P} \ln(1{-}\phi_{\Scal}(h))-\ln(1{-}\phi_{\Scal_i'}(h))$. 

Then, since we have $\max(\phi_{\Scal}(h){-}\frac{1}{2m}, 0) \le \phi_{\Scal_i'}(h) \le \min(\phi_{\Scal}(h){+}\frac{1}{2m}, \frac{1}{2})$, we have 
\begin{align*}
\EE_{h\sim\P} \ln\LP\frac{1{-}\phi_{\Scal_i'}(h)}{1{-}\phi_{\Scal}(h)}\RP \le \EE_{h\sim\P} \ln\LP\frac{1{-}\max(\phi_{\Scal}(h){-}\frac{1}{2m}, 0)}{1{-}\phi_{\Scal}(h)}\RP \le \max_{x\in[0,1]} \ln\LP\frac{1{-}\max(\frac{1}{2}x{-}\frac{1}{2m}, 0)}{1{-}\frac{1}{2}x}\RP.
\end{align*}

For $x\in[0,\frac{1}{m}]$, we have
\begin{align*}
\max_{x\in[0,\frac{1}{m}]} \ln\LP\frac{1}{1{-}\frac{1}{2}x}\RP = \ln\LP\frac{1}{1{-}\frac{1}{2m}}\RP 
\end{align*}
For $x\in[\frac{1}{m}, 1]$, we have
\begin{align*}
\max_{x\in[\frac{1}{m}, 1]} \ln\LP\frac{1{-}\frac{1}{2}x+\frac{1}{2m}}{1{-}\frac{1}{2}x}\RP \le \max_{x\in[\frac{1}{m}, 1]} \ln\LP 1 + \frac{1}{2m\LB 1{-}\frac{1}{2}x\RB}\RP = \ln\LP1+\frac{1}{m}\RP
\end{align*}
Since, for $m\ge1$, we have $\ln\LP\frac{1}{1{-}\frac{1}{2m}}\RP \le \ln\LP1+\frac{1}{m}\RP$, we can deduce that
\begin{align}
\EE_{h\sim\P} \ln\LP\frac{1{-}\phi_{\Scal_i'}(h)}{1{-}\phi_{\Scal}(h)}\RP \le \ln\LP1+\frac{1}{m}\RP.\label{eq:proof-theorem-KL-reverse-2}
\end{align}

Moreover, we have 
\begin{align*}
\EE_{h\sim\P} \ln\LP\frac{1{-}\phi_{\Scal}(h)}{1{-}\phi_{\Scal_i'}(h)}\RP \le \EE_{h\sim\P} \ln\LP\frac{1{-}\phi_{\Scal}(h)}{1{-}\min(\phi_{\Scal}(h){+}\frac{1}{2m}, \frac{1}{2})}\RP \le \max_{x\in[0,1]} \ln\LP\frac{1{-}\frac{1}{2}x}{1{-}\min(\frac{1}{2}x{+}\frac{1}{2m}, \frac{1}{2})}\RP
\end{align*}
For $x\in[0, 1-\frac{1}{m}]$, we have
\begin{align*}
\max_{x\in[0,1-\frac{1}{m}]} \ln\LP\frac{1{-}\frac{1}{2}x}{1{-}\frac{1}{2}x{-}\frac{1}{2m}}\RP = \ln\LP\frac{1{-}\frac{1}{2}+\frac{1}{2m}}{1{-}\frac{1}{2}+\frac{1}{2m}{-}\frac{1}{2m}}\RP = \ln\LP\frac{\frac{1}{2}+\frac{1}{2m}}{\frac{1}{2}}\RP = \ln\LP 1+\frac{1}{m}\RP
\end{align*}
For $x\in[1-\frac{1}{m}, 1]$, we have
\begin{align*}
\max_{x\in[1-\frac{1}{m}, 1]} \ln\LP\frac{1{-}\frac{1}{2}x}{1{-}\frac{1}{2}}\RP = \max_{x\in[1-\frac{1}{m},1]} \ln\LP2-x\RP = \ln\LP 1+\frac{1}{m}\RP.
\end{align*}
Hence, we can deduce that 
\begin{align}
\EE_{h\sim\P} \ln\LP\frac{1{-}\phi_{\Scal}(h)}{1{-}\phi_{\Scal_i'}(h)}\RP \le \ln\LP 1+\frac{1}{m}\RP. \label{eq:proof-theorem-KL-reverse-3}
\end{align}
By combining \Cref{eq:proof-theorem-KL-reverse-2,eq:proof-theorem-KL-reverse-3} we have for all $i\in\{1,\dots,m\}$
\begin{align*}
\left\vert \EE_{h\sim\P} \ln\LP\frac{1{-}\phi_{\Scal_i'}(h)}{1{-}\phi_{\Scal}(h)}\RP \right\vert \le \ln\LP 1+\frac{1}{m}\RP \defeq c_i.
\end{align*}
Then, by substituting in \Cref{eq:proof-theorem-KL-reverse-1}, we have 
\begin{align*}
\EE_{h\sim\Q}\vert\Risk_{\D}(h){-}\Risk_{\Scal}(h)\vert &\le 2\W^{\Gamma}(\Q, \eta) + 2\KLr(\eta\|\P)\\
&+ 2\EE_{\Scal\sim\D^m}\EE_{h\sim\P}\LB{-}\ln\LP1{-}\frac{1}{2}\vert\Risk_{\D}(h){-}\Risk_{\Scal}(h)\vert\RP\RB + 2\ln\LP 1{+}\frac{1}{m}\RP\sqrt{\frac{m}{2}\ln\frac{1}{\delta}}.
\end{align*}

Using the fact that for all $x\in[0,1]$, we have $-\ln(1-\frac{1}{2}x) \le x$.
Hence, we can deduce that 
\begin{align*}
\EE_{h\sim\Q}\vert\Risk_{\D}(h){-}\Risk_{\Scal}(h)\vert &\le 2\W^{\Gamma}(\Q, \eta) + 2\KLr(\eta\|\P)\\
&+ 2\EE_{\Scal\sim\D^m}\EE_{h\sim\P}\vert\Risk_{\D}(h){-}\Risk_{\Scal}(h)\vert + 2\ln\LP 1{+}\frac{1}{m}\RP\sqrt{\frac{m}{2}\ln\frac{1}{\delta}}.
\end{align*}
From Fubini's theorem and Hölder's inequality, we have
\begin{align*}
\EE_{\Scal\sim\D^m}\EE_{h\sim\P}\vert\Risk_{\D}(h){-}\Risk_{\Scal}(h)\vert &= \EE_{h\sim\P}\EE_{\Scal\sim\D^m}\vert\Risk_{\D}(h){-}\Risk_{\Scal}(h)\vert \le \sqrt{\EE_{h\sim\P}\EE_{\Scal\sim\D^m}(\Risk_{\D}(h){-}\Risk_{\Scal}(h))^2}.
\end{align*}
Thanks to \citet{begin2016pac}, we have
\begin{align*}
\sqrt{\EE_{h\sim\P}\EE_{\Scal\sim\D^m}(\Risk_{\D}(h){-}\Risk_{\Scal}(h))^2} \le \sqrt{\frac{1}{4m}},
\end{align*}
which allows us to obtain the desired result.
\end{proof}

\subsubsection{Proof of \Cref{eq:hellinger}}
\begin{proof}
The proof is split into three steps.

\textbf{Step 1: finding the $f$-divergence.}
It is known that the $f$-divergence with $f: x\mapsto (\sqrt{x}-1)^2$ is
\begin{align*}
D_{f}(\Q\|\P) \defeq \EE_{g\sim\P}f\LP\frac{d\Q}{d\P}(g)\RP = \EE_{g\sim\P}\LP\sqrt{\frac{d\Q}{d\P}(g)}-1\RP^2 \defeq \Hell(\Q\|\P).
\end{align*}

\textbf{Step 2: finding an upper bound $B^{\P}(\varphi)$ of $\Lambda^{\P}_{f}(\varphi)$.}
Recall that with $f: x\mapsto (\sqrt{x}-1)^2$, we have
\begin{align*}
f^{*}(y) \defeq \sup_{x\in\R}\LC yx-f(x)\RC = \sup_{x\in\R}\LC g(x,y) \RC, \quad\text{where}\quad g(x,y)\defeq yx+(\sqrt{x}-1)^2.
\end{align*}
In order to find the supremum (which is attained), we can find the derivative of the function $g$, and we have
\begin{align*}
\frac{\partial g}{\partial x}(x, y) = y-1+\frac{1}{\sqrt{x}}.
\end{align*}
Since the Legendre transform is convex, we can set the derivative to $0$ to obtain
\begin{align*}
\frac{\partial g}{\partial x}(x, y) = 0 \iff y-1+\frac{1}{\sqrt{x}} = 0 \iff x=\LP\frac{1}{1-y}\RP^2.
\end{align*}
Hence, we can deduce that we have
\begin{align*}
f^{*}(y)=g\LP\LP\frac{1}{1-y}\RP^2, y\RP = y\LP\frac{1}{1-y}\RP^2-\LP\frac{1}{1-y}-1\RP^2 = \frac{y}{1-y}.
\end{align*}
Now, we upper-bound 
\begin{align*}
\Lambda^{\P}_{f}(\varphi) &\defeq \inf_{c\in\R}\LC c + \EE_{g\sim\P}f^{*}(\varphi(g)-c)\RC\\
&= \inf_{c\in\R}\LC c + \EE_{g\sim\P}\frac{\varphi(g)-c}{1-\varphi(g)+c}\RC\\
&\le \EE_{g\sim\P}\frac{\varphi(g)}{1-\varphi(g)}\defeq B^\P(\varphi).
\end{align*}

\textbf{Step 3: deriving the PAC-Bayesian bound.} 
We use \Cref{theorem:pb-f-gamma-mcdiarmid} with the bound $B^\P(\varphi)=\EE_{g\sim\P}\frac{\varphi(g)}{1-\varphi(g)}$ and 
 the function $\phi_{\Scal}(h) = \frac{1}{2}\vert\Risk_{\D}(h){-}\Risk_{\Scal}(h)\vert$ to obtain the following inequality holding with probability at least $1-\delta$ over $\Scal\sim\D^m$
\begin{align}
\EE_{h \sim \Q}\vert\Risk_{\D}(h){-}\Risk_{\Scal}(h)\vert &\le 2\W^{\Gamma}(\Q, \eta) + 2\Hell(\eta\|\P)\nonumber\\
&+ 2\EE_{\Scal\sim\D^m}\EE_{h\sim\P}\frac{\frac{1}{2}\vert\Risk_{\D}(h){-}\Risk_{\Scal}(h)\vert}{1-\frac{1}{2}\vert\Risk_{\D}(h){-}\Risk_{\Scal}(h)\vert} + 2\sqrt{\frac{\ln\frac{1}{\delta}}{2}\sum_{i=1}^{m}c_i^2}.\label{eq:proof-theorem-hell-1}
\end{align}
The final step is to find the upper bound $c_i$ for each $i\in\{1,\dots, m\}$.
To do so, we first upper-bound the difference $|B^{\P}(\phi_\Scal) - B^{\P}(\phi_{\S_i'})|$ for all $i\in\{1,\dots, m\}$.
We first have
\begin{align*}
\left\vert B^{\P}(\phi_\Scal) - B^{\P}(\phi_{\S_i'})\right\vert &= \left\vert\EE_{h\sim\P}\frac{\phi_\Scal(h)}{1-\phi_\Scal(h)}-\EE_{h\sim\P}\frac{\phi_{\S_i'}(h)}{1-\phi_{\S_i'}(h)}\right\vert.
\end{align*}

Then, since we have $\max(\phi_{\Scal}(h){-}\frac{1}{2m}, 0) \le \phi_{\Scal_i'}(h) \le \min(\phi_{\Scal}(h){+}\frac{1}{2m}, \frac{1}{2})$, we have 
\begin{align*}
\EE_{h\sim\P}\LB\frac{\phi_\Scal(h)}{1-\phi_\Scal(h)}-\frac{\phi_{\S_i'}(h)}{1-\phi_{\S_i'}(h)}\RB &\le \EE_{h\sim\P}\LB\frac{\phi_\Scal(h)}{1-\phi_\Scal(h)}-\frac{\max(\phi_{\Scal}(h){-}\frac{1}{2m}, 0)}{1-\max(\phi_{\Scal}(h){-}\frac{1}{2m}, 0)}\RB\\
&\le \max_{x\in[0,1]} \LB\frac{\frac{1}{2}x}{1-\frac{1}{2}x}-\frac{\max(\frac{1}{2}x{-}\frac{1}{2m}, 0)}{1-\max(\frac{1}{2}x{-}\frac{1}{2m}, 0)}\RB.
\end{align*}

For $x\in[0,\frac{1}{m}]$, we have
\begin{align*}
\max_{x\in[0,\frac{1}{m}]} \LB\frac{\frac{1}{2}x}{1-\frac{1}{2}x}-\frac{\max(\frac{1}{2}x{-}\frac{1}{2m}, 0)}{1-\max(\frac{1}{2}x{-}\frac{1}{2m}, 0)}\RB &= \max_{x\in[0,\frac{1}{m}]} \LB\frac{\frac{1}{2}x}{1-\frac{1}{2}x}\RB\\
&= \frac{\frac{1}{2m}}{1-\frac{1}{2m}}\\
&= \frac{1}{2m-1}.
\end{align*}
For $x\in[\frac{1}{m}, 1]$, we have
\begin{align*}
\max_{x\in[\frac{1}{m}, 1]} \LB\frac{\frac{1}{2}x}{1-\frac{1}{2}x}-\frac{\max(\frac{1}{2}x{-}\frac{1}{2m}, 0)}{1-\max(\frac{1}{2}x{-}\frac{1}{2m}, 0)}\RB &= \max_{x\in[\frac{1}{m}, 1]} \LB\frac{\frac{1}{2}x}{1-\frac{1}{2}x}-\frac{\frac{1}{2}x{-}\frac{1}{2m}}{1-\frac{1}{2}x{+}\frac{1}{2m}}\RB\\ &= \LB \frac{1}{2}-\frac{\frac{1}{2}{-}\frac{1}{2m}}{\frac{1}{2}{+}\frac{1}{2m}}\RB\\
&= \frac{3-m}{2m+2}\\
&\le \frac{1}{2m-1}.
\end{align*}
Hence, we can deduce that
\begin{align}
\EE_{h\sim\P}\LB\frac{\phi_\Scal(h)}{1-\phi_\Scal(h)}-\frac{\phi_{\S_i'}(h)}{1-\phi_{\S_i'}(h)}\RB \le \frac{1}{2m-1}.\label{eq:proof-theorem-hell-2}
\end{align}

Moreover, we have 
\begin{align*}
\EE_{h\sim\P}\LB\frac{\phi_{\S_i'}(h)}{1-\phi_{\S_i'}(h)}-\frac{\phi_\Scal(h)}{1-\phi_\Scal(h)}\RB &\le \EE_{h\sim\P}\LB\frac{\min(\phi_{\Scal}(h){+}\frac{1}{2m}, \frac{1}{2})}{1-\min(\phi_{\Scal}(h){+}\frac{1}{2m}, \frac{1}{2})}-\frac{\phi_\Scal(h)}{1-\phi_\Scal(h)}\RB\\
&\le \max_{x\in[0,1]} \LB\frac{\min(\frac{1}{2}x{+}\frac{1}{2m}, \frac{1}{2})}{1-\min(\frac{1}{2}x{+}\frac{1}{2m}, \frac{1}{2})}-\frac{\frac{1}{2}x}{1-\frac{1}{2}x}\RB.
\end{align*}
For $x\in[0, 1-\frac{1}{m}]$, we have
\begin{align*}
\max_{x\in[0,1-\frac{1}{m}]} \LB\frac{\frac{1}{2}x{+}\frac{1}{2m}}{1-\frac{1}{2}x{-}\frac{1}{2m}}-\frac{\frac{1}{2}x}{1-\frac{1}{2}x}\RB &= \LB \frac{1}{2}-\frac{\frac{1}{2}{-}\frac{1}{2m}}{\frac{1}{2}{+}\frac{1}{2m}}\RB\\
&= \frac{3-m}{2m+2}\\
&\le \frac{1}{2m-1}.
\end{align*}
For $x\in[1-\frac{1}{m}, 1]$, we have
\begin{align*}
\max_{x\in[1-\frac{1}{m}, 1]} \LB\frac{\frac{1}{2}}{1-\frac{1}{2}}-\frac{\frac{1}{2}x}{1-\frac{1}{2}x}\RB &= \LB\frac{\frac{1}{2}}{1-\frac{1}{2}}-\frac{\frac{1}{2}-\frac{1}{2m}}{1-\frac{1}{2}+\frac{1}{2m}}\RB\\
&= \LB1-\frac{\frac{1}{2}-\frac{1}{2m}}{\frac{1}{2}+\frac{1}{2m}}\RB\\
&\le \frac{2}{m+1}.
\end{align*}
Hence, we can deduce that 
\begin{align}
\EE_{h\sim\P}\LB\frac{\phi_{\S_i'}(h)}{1-\phi_{\S_i'}(h)}-\frac{\phi_\Scal(h)}{1-\phi_\Scal(h)}\RB \le \frac{2}{m+1}. \label{eq:proof-theorem-hell-3}
\end{align}
By combining \Cref{eq:proof-theorem-hell-2,eq:proof-theorem-hell-3} we have for all $i\in\{1,\dots,m\}$
\begin{align*}
\left\vert B^{\P}(\phi_\Scal) - B^{\P}(\phi_{\S_i'})\right\vert &= \left\vert\EE_{h\sim\P}\frac{\phi_\Scal(h)}{1-\phi_\Scal(h)}-\EE_{h\sim\P}\frac{\phi_{\S_i'}(h)}{1-\phi_{\S_i'}(h)}\right\vert \le \frac{2}{m+1} \defeq c_i.
\end{align*}

Then, by substituting in \Cref{eq:proof-theorem-hell-1}, we have 
\begin{align*}
\EE_{h\sim\Q}\vert\Risk_{\D}(h){-}\Risk_{\Scal}(h)\vert &\le 2\W^{\Gamma}(\Q, \eta) + 2\Hell(\eta\|\P)\\
&+ 2\EE_{\Scal\sim\D^m}\EE_{h\sim\P}\frac{\frac{1}{2}\vert\Risk_{\D}(h){-}\Risk_{\Scal}(h)\vert}{1-\frac{1}{2}\vert\Risk_{\D}(h){-}\Risk_{\Scal}(h)\vert} + 2\frac{2}{m+1}\sqrt{\frac{m}{2}\ln\frac{1}{\delta}}.
\end{align*}

Using the fact that for all $x\in[0,\frac{1}{2}]$, we have $\frac{x}{1-x} \le 2x$.
Hence, we can deduce that 
\begin{align*}
\EE_{h\sim\Q}\vert\Risk_{\D}(h){-}\Risk_{\Scal}(h)\vert &\le 2\W^{\Gamma}(\Q, \eta) + 2\Hell(\eta\|\P)\\
&+ 4\EE_{\Scal\sim\D^m}\EE_{h\sim\P}\vert\Risk_{\D}(h){-}\Risk_{\Scal}(h)\vert + 2\frac{2}{m+1}\sqrt{\frac{m}{2}\ln\frac{1}{\delta}}.
\end{align*}
From Fubini's theorem and Hölder's inequality, we have
\begin{align*}
\EE_{\Scal\sim\D^m}\EE_{h\sim\P}\vert\Risk_{\D}(h){-}\Risk_{\Scal}(h)\vert &= \EE_{h\sim\P}\EE_{\Scal\sim\D^m}\vert\Risk_{\D}(h){-}\Risk_{\Scal}(h)\vert \le \sqrt{\EE_{h\sim\P}\EE_{\Scal\sim\D^m}(\Risk_{\D}(h){-}\Risk_{\Scal}(h))^2}.
\end{align*}
Thanks to \citet{begin2016pac}, we have
\begin{align*}
\sqrt{\EE_{h\sim\P}\EE_{\Scal\sim\D^m}(\Risk_{\D}(h){-}\Risk_{\Scal}(h))^2} \le \sqrt{\frac{1}{4m}},
\end{align*}
which allows us to obtain the desired result.
\end{proof}

\subsubsection{Proof of \Cref{eq:tv}}

\begin{proof}
The proof is split into three steps.

\textbf{Step 1: finding the $f$-divergence.}
It is known that the $f$-divergence with $f: x\mapsto \frac{1}{2}\vert x-1\vert$ is
\begin{align*}
D_{f}(\Q\|\P) \defeq \EE_{g\sim\P}f\LP\frac{d\Q}{d\P}(g)\RP = \EE_{g\sim\P}\frac{1}{2}\left\vert\frac{d\Q}{d\P}(g)-1\right\vert \defeq \TV(\Q\|\P).
\end{align*}

\textbf{Step 2: finding an upper bound $B^{\P}(\varphi)$ of $\Lambda^{\P}_{f}(\varphi)$.}
Recall that with $f: x\mapsto \frac{1}{2}\vert x-1\vert$, we have
\begin{align*}
f^{*}(y) \defeq \sup_{x\in[0,1]}\LC yx-f(x)\RC = \sup_{x\in[0,1]}\LC g(x,y) \RC, \quad\text{where}\quad g(x,y)\defeq yx-\tfrac{1}{2}\vert x-1\vert.
\end{align*}
First of all, remark that for any $x\in[0,1]$, we have
\begin{align*}
g(x,y) &= yx - \tfrac{1}{2}\vert x-1\vert\\
&= yx - \tfrac{1}{2}(x-1)\\
&= x(y+\tfrac{1}{2})-\tfrac{1}{2}.
\end{align*}
Hence, we can deduce that
\begin{align*}
f^{*}(y) = \sup_{x\in[0,1]}\LC g(x,y) \RC = \sup_{x\in[0,1]}\LC x(y+\tfrac{1}{2})-\tfrac{1}{2} \RC = \left\{ \begin{array}{cc} y & \text{if } y\ge0\\ -\frac{1}{2} & \text{otherwise}.
\end{array} \right.
\end{align*}
Now, we upper-bound 
\begin{align*}
\Lambda^{\P}_{f}(\varphi) &\defeq \inf_{c\in\R}\LC c + \EE_{g\sim\P}f^{*}(\varphi(g)-c)\RC\\
&\le \EE_{g\sim\P}\varphi(g) \defeq B^\P(\varphi).
\end{align*}

\textbf{Step 3: deriving the PAC-Bayesian bound.} 
We use \Cref{theorem:pb-f-gamma-mcdiarmid} with the bound $B^\P(\varphi)=\EE_{g\sim\P}\varphi(g)$ and 
 the function $\phi_{\Scal}(h) = \vert\Risk_{\D}(h){-}\Risk_{\Scal}(h)\vert$ to obtain the following inequality holding with probability at least $1-\delta$ over $\Scal\sim\D^m$
\begin{align}
\EE_{h \sim \Q}\vert\Risk_{\D}(h){-}\Risk_{\Scal}(h)\vert \le \W^{\Gamma}(\Q, \eta) +\TV(\eta, \P) + \EE_{\Scal\sim\D^m}\EE_{h\sim\P}\vert\Risk_{\D}(h){-}\Risk_{\Scal}(h)\vert + \sqrt{\frac{\ln\frac{1}{\delta}}{2}\sum_{i=1}^{m}c_i^2}.\label{eq:proof-theorem-TV-reverse-1}
\end{align}
The final step is to find the upper bound $c_i$ for each $i\in\{1,\dots, m\}$.
To do so, we first upper-bound the difference $|B^{\P}(\phi_\Scal) - B^{\P}(\phi_{\S_i'})|$ for all $i\in\{1,\dots, m\}$.
We first have
\begin{align*}
\left\vert B^{\P}(\phi_\Scal) - B^{\P}(\phi_{\S_i'})\right\vert &= \left\vert\EE_{h\sim\P}\phi_\Scal(h)-\EE_{h\sim\P}\phi_{\S_i'}(h)\right\vert.
\end{align*}

Then, for $\phi_{\Scal}(h) = \vert\Risk_{\D}(h){-}\Risk_{\Scal}(h)\vert$,  we have $\max(\phi_{\Scal}(h){-}\frac{1}{m}, 0) \le \phi_{\Scal_i'}(h) \le \min(\phi_{\Scal}(h){+}\frac{1}{m}, 1)$ and
\begin{align*}
\EE_{h\sim\P}\LB\phi_\Scal(h)-\phi_{\S_i'}(h)\RB &\le \EE_{h\sim\P}\LB\phi_\Scal(h)-\max(\phi_{\Scal}(h){-}\tfrac{1}{m}, 0)\RB\\
&\le \max_{x\in[0,1]} \LB x-\max(x{-}\tfrac{1}{m}, 0)\RB,\\
\text{and}\quad  
\EE_{h\sim\P}\LB\phi_{\S_i'}(h)-\phi_\Scal(h)\RB &\le \EE_{h\sim\P}\LB\min(\phi_{\Scal}(h){+}\tfrac{1}{m}, 1)-\phi_\Scal(h)\RB\\
&\le \max_{x\in[0,1]} \LB \min(x{+}\tfrac{1}{m}, 1)-x\RB.\\
\end{align*}
Moreover, we have 
\begin{align*}
&\max_{x\in[0,\frac{1}{m}]} \LB x-\max(x{-}\tfrac{1}{m}, 0)\RB = \frac{1}{m}, \quad \max_{x\in[\frac{1}{m},1]} \LB x-\max(x{-}\tfrac{1}{m}, 0)\RB = x-x+\tfrac{1}{m} = \frac{1}{m},\\
&\max_{x\in[0, 1-\frac{1}{m}]} \LB \min(x{+}\tfrac{1}{m}, 1)-x\RB = \frac{1}{m}, \quad\text{and}\quad \max_{x\in[1-\frac{1}{m},1]} \LB \min(x{+}\tfrac{1}{m}, 1)-x\RB = 1-x = \frac{1}{m}.
\end{align*}
Hence, we can deduce that 
\begin{align*}
c_i\defeq\left\vert B^{\P}(\phi_\Scal) - B^{\P}(\phi_{\S_i'})\right\vert &= \left\vert\EE_{h\sim\P}\phi_\Scal(h)-\EE_{h\sim\P}\phi_{\S_i'}(h)\right\vert \le \frac{1}{m}
\end{align*}
and we have 
\begin{align*}
\EE_{h\sim\Q}\vert\Risk_{\D}(h){-}\Risk_{\Scal}(h)\vert &\le \W^{\Gamma}(\Q, \eta) +\TV(\eta, \P) + \EE_{\Scal\sim\D^m}\EE_{h\sim\P}\vert\Risk_{\D}(h){-}\Risk_{\Scal}(h)\vert + \sqrt{\frac{\ln\frac{1}{\delta}}{2m}}.
\end{align*}
From Fubini's theorem and Hölder's inequality, we have
\begin{align*}
\EE_{\Scal\sim\D^m}\EE_{h\sim\P}\vert\Risk_{\D}(h){-}\Risk_{\Scal}(h)\vert &= \EE_{h\sim\P}\EE_{\Scal\sim\D^m}\vert\Risk_{\D}(h){-}\Risk_{\Scal}(h)\vert \le \sqrt{\EE_{h\sim\P}\EE_{\Scal\sim\D^m}(\Risk_{\D}(h){-}\Risk_{\Scal}(h))^2}.
\end{align*}
Thanks to \citet{begin2016pac}, we have
\begin{align*}
\sqrt{\EE_{h\sim\P}\EE_{\Scal\sim\D^m}(\Risk_{\D}(h){-}\Risk_{\Scal}(h))^2} \le \sqrt{\frac{1}{4m}},
\end{align*}
which allows us to obtain the desired result.
\end{proof}

\subsection{Proof of \Cref{theorem:lipschitz,theorem:lipschitz-computable}}
We define for conciseness, we define $\Hb = \{ (h,h')\in\Hb \;|\; h\ne h' \}$.
We first prove \Cref{lemma:supremum-Lipschitz,lemma:lipschitz} to further prove \Cref{theorem:lipschitz,theorem:lipschitz-computable}.

\begin{lemma}\label{lemma:supremum-Lipschitz}
For any hypothesis set $\H$ (with the metric $d_{\H}$), for any learning sample $\Scal\in\Zcal^m$, for any function $\phi_{\Scal}:\Hcal\to\R$, for any metric $d:\H\times\H\to\R$, we have
\begin{align*}
\sup_{(h, h') \in \Hb} \frac{\vert\phi_{\Scal}(h)-\phi_{\Scal}(h')\vert}{d_{\H}(h,h')}=\sup_{(h, h') \in \Hb} \frac{\phi_{\Scal}(h)-\phi_{\Scal}(h')}{d_{\H}(h,h')}.
\end{align*}
\end{lemma}
\begin{proof}
For the sake of readability, we denote by $\Hb_{\ge 0} \defeq \{(h, h') \in \Hb \;|\;  \phi_{\mathcal{S}}(h) - \phi_{\mathcal{S}}(h') \ge 0 \}$ and $\Hb_{\le 0} \defeq \{(h, h') \in \Hb \;|\;  \phi_{\mathcal{S}}(h) - \phi_{\mathcal{S}}(h') \le 0 \}$.
Then, we have
\begin{align*}
\sup_{(h, h') \in \Hb} \frac{\phi_{\Scal}(h)-\phi_{\Scal}(h')}{d_{\H}(h,h')} &= \max\LP \sup_{(h, h') \in \Hb_{\le 0}} \frac{\phi_{\Scal}(h)-\phi_{\Scal}(h')}{d_{\H}(h,h')} , \sup_{(h, h') \in \Hb_{\ge 0}} \frac{\phi_{\Scal}(h)-\phi_{\Scal}(h')}{d_{\H}(h,h')}  \RP\\
&= \sup_{(h, h') \in \Hb_{\ge 0}} \frac{\phi_{\Scal}(h)-\phi_{\Scal}(h')}{d_{\H}(h,h')}.
\end{align*}

Let $\Acal = \{ \frac{\vert\phi_{\Scal}(h)-\phi_{\Scal}(h')\vert}{d_{\H}(h,h')} \;|\; (h, h') \in \Hb \}$ and $\Acal_{\ge 0} = \{ \frac{\phi_{\Scal}(h)-\phi_{\Scal}(h')}{d_{\H}(h,h')} \;|\; (h, h') \in \Hb_{\ge 0} \}$,
then, note that since $\Acal_{\ge 0} \subseteq \Acal$ we have for all $a\in\Acal_{\ge 0} \Rightarrow a\in\Acal$.
Moreover, for all $a=\frac{\vert\phi_{\Scal}(h)-\phi_{\Scal}(h')\vert}{d_{\H}(h,h')} \in\Acal$, we have either $a=\frac{\phi_{\Scal}(h)-\phi_{\Scal}(h')}{d_{\H}(h,h')} \in\Acal_{\ge 0}$ (if $\phi_{\Scal}(h)-\phi_{\Scal}(h')\ge0$) or $a=\frac{\phi_{\Scal}(h')-\phi_{\Scal}(h)}{d_{\H}(h,h')} \in\Acal_{\ge 0}$ (if $\phi_{\Scal}(h)-\phi_{\Scal}(h')\le0$) by definition of $\Acal_{\ge 0}$.
Hence, we have $a\in\Acal\Rightarrow a\in\Acal_{\ge 0}$, and so $\Acal=\Acal_{\ge 0}$.

Since we have $\Acal_{\ge 0} = \Acal$, we can deduce that 
\begin{align*}
\sup_{(h, h') \in \Hb_{\ge 0}} \frac{\phi_{\Scal}(h)-\phi_{\Scal}(h')}{d_{\H}(h,h')} = \sup\Acal_{\ge 0} = \sup\Acal = \sup_{(h, h') \in \Hb} \frac{\vert\phi_{\Scal}(h)-\phi_{\Scal}(h')\vert}{d_{\H}(h,h')}.
\end{align*}
\end{proof}

\begin{lemma}\label{lemma:lipschitz}
For any hypothesis set $\H$ (with any metric $d_{\H}$), for any $L$-Lipschitz loss $\loss:\Hcal\times\Zcal\to\R$, with probability at least $1-\delta$ over $\S\sim\D^m$
\begin{align*}
h\mapsto \Delta_{\Scal}(h) \ \ \text{is}\ \ L(m,\delta){=}\!\LP 2\mathdbcal{R}(\H){+}L\sqrt{\tfrac{2\ln\frac{2}{\delta}}{m}}\RP\!\text{-Lipschitz},
\end{align*}
where $\mathdbcal{R}(\H)\defeq \EE_{\varepsilonbf\sim\rad^m}\EE_{\Scal\sim\D^m}\sup_{(h, h') \in \Hb}\frac{1}{m}\sum_{i=1}^{m}\varepsilon_i\frac{\LB\loss(h',\zbf_i)-\loss(h,\zbf_i)\RB}{d_{\H}(h,h')}$
\end{lemma}
\begin{proof}
From \Cref{lemma:supremum-Lipschitz}, we have  
\begin{align*}
\sup_{(h, h') \in \Hb} \frac{\vert\phi_{\Scal}(h)-\phi_{\Scal}(h')\vert}{d_{\H}(h,h')}=\sup_{(h, h') \in \Hb} \frac{\phi_{\Scal}(h)-\phi_{\Scal}(h')}{d_{\H}(h,h')}.
\end{align*}
Hence, the goal is to upper-bound the following quantity with high-probability over $\Scal\sim\D^m$
\begin{align}\label{eq:proof-lip-1}
\sup_{(h, h') \in \Hb} \frac{\phi_{\Scal}(h)-\phi_{\Scal}(h')}{d_{\H}(h,h')},
\end{align}
where $\phi_{\Scal}(h)=\Risk_{\D}(h)-\Risk_{\Scal}(h)$.
First of all, remark that we have
\begin{align*}
\sup_{(h, h') \in \Hb} \frac{\phi_{\Scal}(h)-\phi_{\Scal}(h')}{d_{\H}(h,h')} - \sup_{(h, h') \in \Hb} \frac{\phi_{\Scal'_i}(h)-\phi_{\Scal'_i}(h')}{d_{\H}(h,h')} &\le \sup_{(h, h') \in \Hb} \frac{\phi_{\Scal}(h)-\phi_{\Scal}(h')-\phi_{\Scal'_i}(h)+\phi_{\Scal'_i}(h')}{d_{\H}(h,h')}\\
&= \frac{1}{m} \sup_{(h, h') \in \Hb} \frac{\loss(h,\zbf_i)-\loss(h',\zbf_i)-\loss(h,\zbf'_i)+\loss(h',\zbf'_i)}{d_{\H}(h,h')}\\
&\le \frac{1}{m} \sup_{(h, h') \in \Hb} \frac{\vert \loss(h,\zbf_i)-\loss(h',\zbf_i)\vert +\vert\loss(h,\zbf'_i)-\loss(h',\zbf'_i)\vert}{d_{\H}(h,h')}\\
&\le \frac{2L}{m}.
\end{align*}
Note that by upper-bounding in the same way, we have
\begin{align*}
&\sup_{(h, h') \in \Hb} \frac{\phi_{\Scal'_i}(h)-\phi_{\Scal'_i}(h')}{d_{\H}(h,h')}-\sup_{(h, h') \in \Hb} \frac{\phi_{\Scal}(h)-\phi_{\Scal}(h')}{d_{\H}(h,h')} \le \frac{2L}{m}.
\end{align*}
We can thus deduce that \eqref{eq:proof-lip-1} has the bounded-difference property.
Hence, we can apply McDiarmid's inequality to have
\begin{align*}
\EE_{\Scal\sim\D^m}\sup_{(h, h') \in \Hb} \frac{\phi_{\Scal}(h)-\phi_{\Scal}(h')}{d_{\H}(h,h')} &= \EE_{\Scal\sim\D^m}\sup_{(h, h') \in \Hb} \frac{\Risk_{\D}(h)-\Risk_{\Scal}(h)-\Risk_{\D}(h')+\Risk_{\Scal}(h')}{d_{\H}(h,h')}\\
&\le \EE_{\Scal\sim\D^m}\EE_{\Scal'\sim\D} \sup_{(h, h') \in \Hb} \frac{\Risk_{\Scal'}(h)-\Risk_{\Scal}(h)-\Risk_{\Scal'}(h')+\Risk_{\Scal}(h')}{d_{\H}(h,h')}.
\end{align*}
Hence, from the symmetrization lemma, we have
\begin{align*}
&\EE_{\Scal\sim\D^m}\EE_{\Scal'\sim\D} \sup_{(h, h') \in \Hb} \frac{\Risk_{\Scal'}(h)-\Risk_{\Scal}(h)-\Risk_{\Scal'}(h')+\Risk_{\Scal}(h')}{d_{\H}(h,h')}\\
= &\EE_{\Scal\sim\D^m}\EE_{\Scal'\sim\D} \sup_{(h, h') \in \Hb} \frac{1}{d_{\H}(h,h')}\frac{1}{m}\sum_{i=1}^{m}\LB\loss(h,\zbf_i')-\loss(h,\zbf_i)-\loss(h',\zbf_i')+\loss(h',\zbf_i)\RB\\
= &\EE_{\varepsilonbf\sim\rad^m}\EE_{\Scal\sim\D^m}\EE_{\Scal'\sim\D} \sup_{(h, h') \in \Hb} \frac{1}{d_{\H}(h,h')}\frac{1}{m}\sum_{i=1}^{m}\varepsilon_i\LB\loss(h,\zbf_i')-\loss(h,\zbf_i)-\loss(h',\zbf_i')+\loss(h',\zbf_i)\RB\\
\le &\EE_{\varepsilonbf\sim\rad^m}\EE_{\Scal\sim\D^m}\sup_{(h, h') \in \Hb}\frac{1}{m}\sum_{i=1}^{m}\varepsilon_i\frac{\LB\loss(h',\zbf_i)-\loss(h,\zbf_i)\RB}{d_{\H}(h,h')}+
\EE_{\varepsilonbf\sim\rad^m}\EE_{\Scal'\sim\D}\sup_{(h, h') \in \Hb}\frac{1}{m}\sum_{i=1}^{m}\varepsilon_i\frac{\LB\loss(h,\zbf_i')-\loss(h',\zbf_i')\RB}{d_{\H}(h,h')}\\
= &\EE_{\varepsilonbf\sim\rad^m}\EE_{\Scal\sim\D^m}\sup_{(h, h') \in \Hb}\frac{2}{m}\sum_{i=1}^{m}\varepsilon_i\frac{\LB\loss(h',\zbf_i)-\loss(h,\zbf_i)\RB}{d_{\H}(h,h')}
\end{align*}

We have with probability at least $1-\delta$ over $\Scal\sim\D^m$
\begin{align}
\sup_{(h, h') \in \Hb} \frac{\phi_{\Scal}(h)-\phi_{\Scal}(h')}{d_{\H}(h,h')} &\le \EE_{\Scal\sim\D^m}\sup_{(h, h') \in \Hb} \frac{\phi_{\Scal}(h)-\phi_{\Scal}(h')}{d_{\H}(h,h')} + \sqrt{\frac{\ln\frac{1}{\delta}}{2}m\frac{4L^2}{m^2}}\nonumber\\
&= \EE_{\Scal\sim\D^m}\sup_{(h, h') \in \Hb} \frac{\phi_{\Scal}(h)-\phi_{\Scal}(h')}{d_{\H}(h,h')} + L\sqrt{\frac{2\ln\frac{1}{\delta}}{m}}\nonumber\\
&\le \EE_{\varepsilonbf\sim\rad^m}\EE_{\Scal\sim\D^m}\sup_{(h, h') \in \Hb}\frac{2}{m}\sum_{i=1}^{m}\varepsilon_i\frac{\LB\loss(h',\zbf_i)-\loss(h,\zbf_i)\RB}{d_{\H}(h,h')} + L\sqrt{\frac{2\ln\frac{1}{\delta}}{m}}.\label{eq:proof-lemma-lipschitz-1}
\end{align}

Finally, we apply the same proof with $\phi_{\Scal}(h)=\Risk_{\Scal}(h)-\Risk_{\D}(h)$ to obtain
\begin{align}
\sup_{(h, h') \in \Hb} \frac{\phi_{\Scal}(h)-\phi_{\Scal}(h')}{d_{\H}(h,h')} \le  \EE_{\varepsilonbf\sim\rad^m}\EE_{\Scal\sim\D^m}\sup_{(h, h') \in \Hb}\frac{2}{m}\sum_{i=1}^{m}\varepsilon_i\frac{\LB\loss(h',\zbf_i)-\loss(h,\zbf_i)\RB}{d_{\H}(h,h')} + L\sqrt{\frac{2\ln\frac{1}{\delta}}{m}}.\label{eq:proof-lemma-lipschitz-2}
\end{align}
Hence, by merging \Cref{eq:proof-lemma-lipschitz-1,eq:proof-lemma-lipschitz-2} with a union bound, we obtain the desired result.  
\end{proof}

We are now able to prove \Cref{theorem:lipschitz}.

\theoremLipschitz*
\begin{proof}
The proof boils down to upper-bound the term $\mathdbcal{R}(\H)$.
Indeed, we have
\begin{align*}
\mathdbcal{R}(\H)&= \EE_{\varepsilonbf\sim\rad^m}\EE_{\Scal\sim\D^m}\sup_{(h, h') \in \Hb}\frac{1}{m}\sum_{i=1}^{m}\varepsilon_i\LB\loss(h',\zbf_i)-\loss(h,\zbf_i)\RB\\
&\le \EE_{\varepsilonbf\sim\rad^m}\EE_{\Scal\sim\D^m}\sup_{(h, h') \in \H^2}\frac{1}{m}\sum_{i=1}^{m}\varepsilon_i\LB\loss(h',\zbf_i)-\loss(h,\zbf_i)\RB\\
&= \EE_{\varepsilonbf\sim\rad^m}\EE_{\Scal\sim\D^m}\sup_{h' \in \H^2}\frac{1}{m}\sum_{i=1}^{m}\varepsilon_i\loss(h',\zbf_i) + \EE_{\varepsilonbf\sim\rad^m}\EE_{\Scal\sim\D^m}\sup_{h \in \H}\frac{1}{m}\sum_{i=1}^{m}(-\varepsilon_i)\loss(h,\zbf_i)\\
&= 2\EE_{\varepsilonbf\sim\rad^m}\EE_{\Scal\sim\D^m}\sup_{h' \in \H^2}\frac{1}{m}\sum_{i=1}^{m}\varepsilon_i\loss(h',\zbf_i)\\
&\defeq 2\Rfrak(\Hcal).
\end{align*}
\end{proof}

\theoremLipschitzcomputable*
\begin{proof}
First of all, let $\varepsilonbf'_i$ be the vector $\varepsilonbf$ that differs only from its $i$-th element.
Now, remark that we have
\begin{align*}
\mathdbcal{R}_{\Scal}^{\varepsilonbf}(\H)-\mathdbcal{R}_{\Scal_i'}^{\varepsilonbf'_i}(\H) \le \sup_{(h, h') \in \Hb}\frac{1}{m}\LB\varepsilon_i\frac{\LB\loss(h',\zbf_i)-\loss(h,\zbf_i)\RB}{d_{\H}(h,h')} - \varepsilon'_i\frac{\LB\loss(h',\zbf_i')-\loss(h,\zbf_i')\RB}{d_{\H}(h,h')}\RB \le \frac{2L}{m}
\end{align*}
and similarly, we have
\begin{align*}
\mathdbcal{R}_{\Scal_i'}^{\varepsilonbf'_i}(\H)-\mathdbcal{R}_{\Scal}^{\varepsilonbf}(\H) \le \sup_{(h, h') \in \Hb}\frac{1}{m}\LB\varepsilon_i\frac{\LB\loss(h',\zbf_i')-\loss(h,\zbf_i')\RB}{d_{\H}(h,h')} - \varepsilon'_i\frac{\LB\loss(h',\zbf_i)-\loss(h,\zbf_i)\RB}{d_{\H}(h,h')}\RB \le \frac{2L}{m}.
\end{align*}
Hence, we can deduce that $\mathdbcal{R}_{\Scal}^{\varepsilonbf}(\H)$ has the bounded-difference property.
Hence, we can apply McDiarmid's inequality (with $\delta/2$ instead of $\delta$) to obtain with probability  at least $1-\delta/2$ over $\Scal\sim\D^m$ and $\varepsilonbf\sim\rad^m$
\begin{align}
2\mathdbcal{R}(\H) = 2\EE_{\varepsilonbf\sim\rad^m}\EE_{\Scal\sim\D^m}\mathdbcal{R}_{\Scal}^{\varepsilonbf}(\H) \le 2\mathdbcal{R}_{\Scal}^{\varepsilonbf}(\H) + 2L\sqrt{\frac{\ln\frac{2}{\delta}}{m}}.\label{eq:proof-theorem-Lipschitz-computable-1}
\end{align}
From \Cref{lemma:lipschitz} (with $\delta/2$ instead of $\delta$), we have
\begin{align}
\sup_{(h, h') \in \Hb} \frac{\vert\phi_{\Scal}(h)-\phi_{\Scal}(h')\vert}{d_{\H}(h,h')} \le 2\mathdbcal{R}(\H) + L\sqrt{\frac{2\ln\frac{4}{\delta}}{m}}.\label{eq:proof-theorem-Lipschitz-computable-2}
\end{align}
By combining \eqref{eq:proof-theorem-Lipschitz-computable-1} and \eqref{eq:proof-theorem-Lipschitz-computable-2}, we have
\begin{align*}
\sup_{(h, h') \in \Hb} \frac{\vert\phi_{\Scal}(h)-\phi_{\Scal}(h')\vert}{d_{\H}(h,h')} &\le 2\mathdbcal{R}_{\Scal}^{\varepsilonbf}(\H) + L\sqrt{\frac{2\ln\frac{4}{\delta}}{m}} + 2L\sqrt{\frac{2\ln\frac{2}{\delta}}{m}}\\
&\le 2\mathdbcal{R}_{\Scal}^{\varepsilonbf}(\H) + L\sqrt{\frac{2\ln\frac{4}{\delta}}{m}} + 2L\sqrt{\frac{2\ln\frac{4}{\delta}}{m}},
\end{align*}
which leads to the desired result.
\end{proof}

\subsection{Proof of \Cref{corollary:rademacher}}

To prove \Cref{corollary:rademacher}, we first prove the following corollary.

\begin{restatable}{corollary}{corollarytvlipschitz}\label{corollary:tv}
Assume that $\ell \in [0,1]$. Assume that, for any $\delta'\in (0,1)$, with probability $1-\delta'$, $h\rightarrow\Delta_\S(h)$ is $L(m,\delta)$-Lipschitz \wrt the Kronecker distance. Thus, for any data-free prior $\P$, with probability at least $1-\delta$ over $\Scal\sim\D^m$, we have for all $\Q\in\Pcal(\H)$, any $\eta\in\Pcal(\H)$,
\begin{align*}
    \EE_{h\sim\Q}\Delta_{\S}(h) \le L(m,\delta)\TV(\Q, \eta)  +\TV(\eta\|\P)+ \sqrt{\frac{1}{4m}} + \sqrt{\frac{\ln\frac{1}{\delta}}{2m}}.
\end{align*}
\end{restatable}
\begin{proof}
We apply \Cref{theorem:KL-reverse-hellinger-gamma}, in particular, \Cref{eq:tv} and we use the fact that $\W^{\Gamma}(\Q,\eta)=L(m,\delta)\TV(\eta, \P)$.
\end{proof}

We are now able to prove \Cref{corollary:rademacher}.
\corollaryrademacher*
\begin{proof}
We combine \Cref{corollary:tv} and \Cref{theorem:lipschitz} to obtain the first inequality.
The second inequality is obtained by setting $\eta=\P$ and upper-bounding $\TV(\Q, \eta)\le 1$.
\end{proof}

\subsection{Proofs for heavy-tailed SGD}
\label{sec:proof-ht-sgd}

We first state properly the assumption (\textbf{H1}), (\textbf{H2}) extracted from \citet{raj2023algo}. 
\begin{itemize}
    \item \textbf{(H1) (Quasi-smoothness)} There exist constants $\theta_0>0$ and $K \geq 0$ such that $\langle x-y, \nabla \hat{F}_\S (x)- \nabla \hat{F}_\S(y)\rangle \geq \theta_0\|x-y\|^2-K \quad$ for all $x, y \in \H ;$
    \item \textbf{(H2) (Regularity)} Rename $b= \nabla \hat{F}_\S$. Assume $b$ to be $\Ccal^3(\Rbb^d,\Rbb)$. There exist constants $\theta_1, \theta_2, \theta_3 \geq 0$ such that
$$
\begin{aligned}
\left\|\nabla_v b(x)\right\| & \leq \theta_1\|v\|, \quad v, x \in \mathbb{R}^d, \\
\left\|\nabla_{v_1} \nabla_{v_2} b(x)\right\| & \leq \theta_2\left\|v_1\right\|\left\|v_2\right\|, \quad v_1, v_2, x \in \mathbb{R}^d, \\
\left\|\nabla_{v_1} \nabla_{v_2} \nabla_{v_3} b(x)\right\| & \leq \theta_3\left\|v_1\right\|\left\|v_2\right\|\left\|v_3\right\|, \quad v_1, v_2, v_3, x \in \mathbb{R}^d,
\end{aligned}
$$
where the directional derivatives are defined as 
\begin{align*}
\nabla_{v_1} b(x)\defeq\lim _{\epsilon \rightarrow 0} \frac{b\left(x+\epsilon v_1\right)-b(x)}{\epsilon}
\end{align*}
\begin{align*}
\nabla_{v_2} \nabla_{v_1} f(x)\defeq\lim _{\epsilon \rightarrow 0} \frac{\nabla_{v_1} b\left(x+\epsilon v_2\right)-\nabla_{v_1} b(x)}{\epsilon}
\end{align*}
and
\begin{align*}
\nabla_{v_3} \nabla_{v_2} \nabla_{v_1} b(x)\defeq\lim _{\epsilon \rightarrow 0} \frac{\nabla_{v_2} \nabla_{v_1} b\left(x+\epsilon v_3\right)-\nabla_{v_2} \nabla_{v_1} b(x)}{\epsilon}.
\end{align*}
\end{itemize}

We are now ready to prove \Cref{theorem:heavy-tailed-sgd}.
\HTSGD*
\begin{proof}
    We start from \Cref{theorem:KL-wass} with the Gaussian prior $\P$. We have, with probability at least $1-\delta$, for any distributions $\eta,\Q$, 

    \begin{align*}
        \EE_{h\sim\Q}\Delta_{\S}(h)\le 
        \sqrt{ L(m,\nicefrac{\delta}{2})\W_1(\Q, \eta) + \frac{\KL(\eta\|\P)+ \ln\frac{4\sqrt{m}}{\delta}}{2m}} .
    \end{align*}

    We take $\eta= \Q_2, \Q= \Q_\alpha$. We then know that under $(H1), (H2)$, \citet[Theorem 1]{deng2023wasserstein} gives, for any $\alpha>\alpha_0$, 

    \[ \W_1 (\Q_\alpha,\Q_2)\leq C_{\alpha_0} f(\alpha,d).\]

    Plugging this into the previous bound concludes the proof.
\end{proof}

\section{About The Experiments}
\label{sec:supp-expe}

\subsection{Lipschitzness for neural networks}
\label{sec:lip-nn}

\begin{lemma}
\label{lemma:neural-lipschitz}
Consider the neural networks and the setting of \Cref{sec:setting-result} (with $\Xcal=\{ \xbf \in \R^n | \|\xbf\|_2 \le 1 \}$), then, the loss is $\alpha\sqrt{2(K{+}2)}$-Lipschitz \wrt the parameters on the subspace where all the weight matrices have their Frobenius norm bounded by 1 (no constraint on biases).
\end{lemma}
\begin{proof}
First of all, since the loss is $\alpha$-Lipschitz \wrt the outputs, we have
\begin{align*}
    |\loss(h_{\wbf}, \zbf) - \loss(h_{\wbf'}, \zbf)| &\le \alpha\|h_{\wbf}(\xbf)-h_{\wbf'}(\xbf)\|\\
    &= \alpha\|(Wh_{\wbf}^{K}(\xbf){+}b) - (W'h_{\wbf'}^{K}(\xbf){+}b')\|\\
    &= \alpha\|Wh_{\wbf}^{K}(\xbf) + W'h_{\wbf'}^{K}(\xbf) + W'h_{\wbf}^{K}(\xbf) - W'h_{\wbf}^{K}(\xbf) + b-b'\|\\
    &\le \alpha\Big(\|(W{-}W')h_{\wbf}^{K}(\xbf)\| + \|W'(h_{\wbf}^{K}(\xbf){-}h_{\wbf'}^{K}(\xbf))\| + \|b-b'\|\Big).
\end{align*}
Moreover, note that we have 
\begin{align*}
&\|(W{-}W')h_{\wbf}^{K}(\xbf)\| \le \|W{-}W'\|_{F}\|h_{\wbf}^{K}(\xbf)\| \le \|W{-}W'\|_{F},\\
\text{and}\quad &\|W'(h_{\wbf}^{K}(\xbf){-}h_{\wbf'}^{K}(\xbf))\| \le \|W'\|_F\|h_{\wbf}^{K}(\xbf){-}h_{\wbf'}^{K}(\xbf)\| \le \|h_{\wbf}^{K}(\xbf){-}h_{\wbf'}^{K}(\xbf)\|
\end{align*}
since we have $\|h_{\wbf}^{L}(\xbf)\| \le 1$ and $\|W'\|_F\le 1$.
Hence, we can deduce that
\begin{align*}
|\loss(h_{\wbf}, \zbf) - \loss(h_{\wbf'}, \zbf)| \le \alpha\Big(\|W{-}W'\|_F + \|h_{\wbf}^{K}(\xbf){-}h_{\wbf'}^{K}(\xbf)\| + \|b-b'\|\Big).
\end{align*}
Moreover, for any $i\in\{1,\dots,L\}$ (with $h^{0}(\xbf)=\xbf$), we have
\begin{align}
    \|h^{i}_{\wbf}(\xbf)-h^{i}_{\wbf'}(\xbf)\| &= \|\proj(\leaky(W_ih^{i-1}_{\wbf}(\xbf){+}b_i)) - \proj(\leaky(W_i'h_{\wbf'}^{i-1}(\xbf){+}b_i'))\|\nonumber\\
    &\le \|\leaky(W_ih^{i-1}_{\wbf}(\xbf){+}b_i) - \leaky(W_i'h_{\wbf'}^{i-1}(\xbf){+}b_i')\|\nonumber\\
    &\le \|(W_ih^{i-1}_{\wbf}(\xbf){+}b_i) - (W_i'h_{\wbf'}^{i-1}(\xbf){+}b_i')\|\nonumber\\
    &\le \Big(\|(W_i{-}W_i')h_{\wbf}^{i-1}(\xbf)\| + \|W_i'(h_{\wbf}^{i-1}(\xbf){-}h_{\wbf'}^{i-1}(\xbf))\| + \|b_i-b_i'\|\Big)\nonumber\\
    &\le \Big(\|W_i{-}W_i'\|_F + \|h_{\wbf}^{i-1}(\xbf){-}h_{\wbf'}^{i-1}(\xbf)\| + \|b_i-b_i'\|\Big).\label{eq:proof-neural-lipschitz-1}
\end{align}
Hence, by applying \Cref{eq:proof-neural-lipschitz-1} for any $i\in\{1,\dots,K\}$ we can deduce that
\begin{align}
    |\loss(h_{\wbf}, \zbf) - \loss(h_{\wbf'}, \zbf)| \le \alpha\LP \|W-W'\| + \|b-b'\| + \sum_{i=1}^{K}\LP\|W_i-W_i'\|{+}\|b_i-b_i'\|\RP\RP.\label{eq:proof-neural-lipschitz-2}
\end{align}
Now, the goal is to upper-bound $\|W-W'\| + \|b-b'\| + \sum_{i=1}^{K}\LP\|W_i-W_i'\|{+}\|b_i-b_i'\|\RP$ by $\|\wbf-\wbf'\|$ multiplied by a constant.
To do so, we can use the fact that for any real numbers $a,b$, we have $(a+b)^2\le 2(a^2 + b^2)$ in order to gather the norms.
If $K$ is odd, we can apply the property in a divide-and-conquer manner to obtain
\begin{align}
\|W-W'\| + \|b-b'\| + \sum_{i=1}^{L}\LP\|W_i-W_i'\|{+}\|b_i-b_i'\|\RP &\le \sqrt{2^{\log_2(2(K{+}1))}}\|\wbf-\wbf'\|\nonumber\\
&= \sqrt{2(K{+}1)}\|\wbf-\wbf'\|\nonumber\\
&\le \sqrt{2(K{+}2)}\|\wbf-\wbf'\|\label{eq:proof-neural-lipschitz-3}
\end{align}
and when $L$ is even, we also have
\begin{align}
\|W-W'\| + \|b-b'\| + \sum_{i=1}^{L}\LP\|W_i-W_i'\|{+}\|b_i-b_i'\|\RP &\le \sqrt{2^{\log_2(2(K{+}2))}}\|\wbf-\wbf'\|\nonumber\\
&= \sqrt{2(K{+}2)}\|\wbf-\wbf'\|.\label{eq:proof-neural-lipschitz-4}
\end{align}
Hence, we can deduce from \Cref{eq:proof-neural-lipschitz-2,eq:proof-neural-lipschitz-3,eq:proof-neural-lipschitz-4} that
\begin{align*}
    |\loss(h_{\wbf}, \zbf) - \loss(h_{\wbf'}, \zbf)| \le \alpha\sqrt{2(K{+}2)}\|\wbf-\wbf'\|,
\end{align*}
which concludes the proof.
\end{proof}

\subsection{Additional Insights on the Experiments}

We present, in this section, additional information concerning the experimental setting.
Moreover, we present complementary experiments with \Cref{eq:hellinger,eq:reverse-KL,eq:tv}; the experiments are presented from \Cref{table:expe-supp-1,table:expe-supp-2,table:expe-supp-3,table:expe-supp-4,table:expe-supp-5,table:expe-supp-6,table:expe-supp-7,table:expe-supp-8,table:expe-supp-9}.

\textbf{About additional experiments with data-dependent priors.} In the PAC-Bayesian framework, using data-dependent priors to tighten the bound has been a popular strategy \citep[see \eg,][]{ambroladze2006tighter,parradohernandez2012pac,dziugaite2021role,perez2021tighter,viallard2023general}.
Hence, we provide additional experiments where the prior vector $\wbf_{\P}$ is learned with a portion of the original training set. 
More precisely, the original training set is split into two sets $\Scal$ and $\Scal'$; the set $\Scal$ is used when we consider data-free priors while $\Scal'$ aims to produce a good vector $\wbf_{\P}$.
To do so, the weight vector $\wbf_{\P}$ by performing an empirical risk minimisation (while keeping the same parameters for the optimiser).
The vector is selected by early stopping with the set $\Scal$ (with the empirical risk as metric). 
Since we select the weights by early stopping, we have to perform a union bound to obtain a bound holding for all the intermediate vectors.
Hence, imagine that we have $T$ weight vectors (corresponding to $T$ epochs), we have to replace $\ln\frac{1}{\delta}$ by $\ln\frac{T}{\delta}$ to obtain a bound holding for the $T$ weight vectors with probability at least $1-\delta$.
Moreover, note that when we estimate the Lipschitz constant, we still use the original training set (to obtain a better Lipschitz constant).

\textbf{Learning the prior variance $\sigma^2_{\P}$.}
In order to learn the prior variance, we perform a union bound similar to the one of \citet{dziugaite2017computing}.
To do so, for each bound that uses a Gaussian prior $\Ncal(\wbf_{\P}, \sigma^2_{\P}\Irm_{d})$, we consider a bound that holds with probability at least $1-\frac{6}{\pi^2j^2}$ where the variance is defined by $\sigma^2_{\P}=c\exp(-j/b)$.
The value $c$ corresponds to an upper bound of the variance, and $b$ corresponds to a level of precision; we set $c=1.1$ and $b=100$.
By performing the union bound for all $j\in\N$, we obtain a bound holding for all discretised variance with probability at least $1-\delta$.
By doing so, the confidence term $\ln(\nicefrac{1}{\delta})$ is replaced by $2\ln(b\ln(c/\sigma^2_{\P})) + \ln(\nicefrac{\pi^2}{6\delta})$.
Note that during the optimization, we do not constrain the variance, but we discretise it only during the evaluation of the bound.

\textbf{Estimating the empirical risk for Gaussian posteriors.}
In order to evaluate the (expected) empirical risk $\EE_{h\sim\Q}\Risk_{\Scal}(h)$, we perform a Monte Carlo sampling.
However, to have a bound that remains valid, we use Hoeffding's inequality to obtain with probability at least $1-\delta$ over $h_1,\dots,h_T\sim \Q^T$
\begin{align*}
\EE_{h\sim\Q}\Riskhat_{\Scal}(h) \le \frac{1}{T}\sum_{t=1}^{T}\Riskhat_{\Scal}(h_t) + \sqrt{\frac{2\ln\frac{1}{\delta}}{T}}.
\end{align*}
In the experiments, we set $T=1000$.

\textbf{About the union bounds in the bounds.}
Depending on the bound we consider, we have to perform a union bound to take into account the Lipschitz constant and the sampling for the estimation of the empirical risk.
For instance, in the Dirac case, we only have to consider the Lipschitz constant.
Hence, to perform the union bound, the bound holds with probability at least $1-\delta/2$ as well as the Lipschitz constant $L(m, \delta/2)$. 
For the Gaussian case, when we have $\Q=\eta$ (\ie, with $\W_1(\Q,\eta)=0$), we perform a union bound with Hoeffding's bound holding with probability at least $1-\delta/2$ as well as the bound.
However, when we consider the Wasserstein distance, the original bound, the Lipschitz constant, and Hoeffding's bound hold individually with probability at least $1-\delta/3$ before applying the union bound.

\textbf{Value of the Wasserstein distance.} The value of the Wasserstein distance is given by the following formula.
\begin{itemize}
    \item Wasserstein distance (Gaussians posterior and prior):
    \begin{align*}
    \W_1(\Q,\eta) \le\W_2(\Q,\eta) = \sqrt{\|\wbf-\wbf_\eta\|^2 + \LB d(\sigma_{\eta}-\sigma_{\Q})\RB^2}
    \end{align*}
    \item Wasserstein distance (Dirac posterior, Gaussian prior):
    \begin{align*}
    \W_1(\Q,\eta) \le \|\wbf{-}\wbf_{\eta}\|_2 + \sigma_{\eta}\sqrt{2}\frac{\Gamma((d+1)/2)}{\Gamma(d/2)}
    \end{align*}
\end{itemize}

\textbf{Value of the $f$-divergences.} The different $f$-divergences between two Gaussian distributions are defined as follows.
\begin{itemize}
    \item KL divergence:
    \begin{align*}
    \KL(\eta\|\P)\! =\! \frac{1}{2}\!\left[\frac{\sigma^2_{\eta}}{\sigma^2_{\P}}d - d +\frac{1}{\sigma^2_{\P}}\|\wbf_{\eta}{-}\wbf_{\P}\|_2^2 +d\ln\!\LP\frac{\sigma^2_{\P}}{\sigma^2_{\eta}}\RP\right]
    \end{align*}
    \item (Squared) Hellinger distance:
    \begin{align*}
    \Hell(\eta\|\P) = 1-\exp\LP \frac{d}{4}\LP\ln(\sigma^2){+}\ln(\sigma_{\P}^2)\RP -\frac{d}{2}\ln(\tfrac{1}{2}(\sigma^2{+}\sigma^2_{\P})) -\frac{\|\wbf-\wbf_{\P}\|^2}{4(\sigma^2{+}\sigma^2_{\P})} \RP
    \end{align*}
    \item Reverse KL divergence:
    \begin{align*}
    \KLr(\eta\|\P) = \KL(\P\|\eta) = \frac{1}{2}\!\left[\frac{\sigma^2_{\P}}{\sigma^2_{\eta}}d - d +\frac{1}{\sigma^2_{\eta}}\|\wbf_{\P}{-}\wbf_{\eta}\|_2^2 +d\ln\!\LP\frac{\sigma^2_{\eta}}{\sigma^2_{\P}}\RP\right]
    \end{align*}
\end{itemize}

\begin{table*}
\caption{
Results of the bound minimisation of \Cref{theorem:KL-wass} (see \Cref{eq:alg-KL-wass}) and the ones associated with \citet{amit2022integral} and \citet{maurer2004note} when the prior is learned with 25\% of the original training set.
``Test'' is the test risk $\Riskhat_{\Tcal}(h_{\wbf})$, ``Bnd'' represents the bound value, ``Wass'' represents the upper bound of the Wasserstein distance multiplied by the Lipschitz constant, and ``KL'' is the KL divergence divided by $2m$.
}
\centering
\begin{minipage}{.55\linewidth}
  \centering
  \scalebox{\scalesize}{
  \begin{tabular}{l|c@{\hspace{0.1cm}}c@{\hspace{0.1cm}}@{\hspace{0.1cm}}c@{\hspace{0.1cm}}@{\hspace{0.1cm}}c|c@{\hspace{0.1cm}}@{\hspace{0.1cm}}c@{\hspace{0.1cm}}@{\hspace{0.1cm}}c|}
\toprule
 & \multicolumn{4}{c}{\Cref{theorem:KL-wass}} & \multicolumn{3}{c}{\citet{amit2022integral}} \\
 & Test & Bnd & Wass. & KL & Test & Bnd & Wass. \\
\midrule
\fashion & 0.063 & 0.088 & 0.000 & 0.000 & 0.064 & 0.289 & 0.052 \\
\mnist & 0.038 & 0.068 & 0.000 & 0.000 & 0.038 & 0.199 & 0.025 \\
\mushrooms & 0.004 & 0.077 & 0.000 & 0.000 & 0.004 & 0.073 & 0.002 \\
\phishing & 0.070 & 0.130 & 0.000 & 0.000 & 0.070 & 0.129 & 0.002 \\
\yeast & 0.188 & 0.363 & 0.000 & 0.000 & 0.188 & 0.393 & 0.025 \\
\bottomrule
\end{tabular}
}
  \caption*{(a) Linear models $h_\wbf$ with $\Q=\delta_{\wbf}$}
\end{minipage}%
\begin{minipage}{.4\linewidth}
  \centering
  \scalebox{\scalesize}{
  \begin{tabular}{|c@{\hspace{0.15cm} }c@{\hspace{0.15cm}}c@{\hspace{0.15cm}}c|c@{\hspace{0.15cm}}c@{\hspace{0.15cm}}c|}
\toprule
\multicolumn{4}{c}{\Cref{theorem:KL-wass}} & \multicolumn{3}{c}{\citet{amit2022integral}} \\
Test & Bnd & Wass. & KL & Test & Bnd & Wass. \\
\midrule
0.111 & 0.248 & 0.014 & 0.004 & 0.884 & 3.537 & 7.041 \\
0.088 & 0.238 & 0.016 & 0.004 & 0.776 & 2.414 & 2.680 \\
0.002 & 0.177 & 0.019 & 0.005 & 0.487 & 3.832 & 11.247 \\
0.068 & 0.211 & 0.014 & 0.002 & 0.438 & 1.903 & 2.095 \\
0.329 & 0.614 & 0.034 & 0.009 & 0.383 & 2.469 & 4.318 \\
\bottomrule
\end{tabular}
}
  \caption*{(b) Neural networks $h_\wbf$ with $\Q=\delta_{\wbf}$}
\end{minipage}
\begin{minipage}{.55\linewidth}
  \centering
  \scalebox{\scalesize}{
  \begin{tabular}{l|c@{\hspace{0.1cm}}c@{\hspace{0.1cm}}@{\hspace{0.1cm}}c@{\hspace{0.1cm}}@{\hspace{0.1cm}}c|c@{\hspace{0.1cm}}@{\hspace{0.1cm}}c@{\hspace{0.1cm}}@{\hspace{0.1cm}}c|}
\toprule
 & \multicolumn{4}{c}{\Cref{theorem:KL-wass}} & \multicolumn{3}{c}{\citet{maurer2004note}} \\
 & Test & Bnd & Wass. & KL & Test & Bnd & KL \\
\midrule
\fashion & 0.119 & 0.603 & 0.000 & 0.122 & 0.068 & 0.171 & 0.000 \\
\mnist & 0.079 & 0.531 & 0.000 & 0.098 & 0.042 & 0.153 & 0.000 \\
\mushrooms & 0.005 & 0.296 & 0.000 & 0.028 & 0.005 & 0.292 & 0.028 \\
\phishing & 0.076 & 0.226 & 0.000 & 0.000 & 0.076 & 0.396 & 0.042 \\
\yeast & 0.190 & 0.454 & 0.000 & 0.000 & 0.220 & 0.883 & 0.235 \\
\bottomrule
\end{tabular}
}
  \caption*{(c) Linear models $h_{\wbf}$ with $\Q=\Ncal(\wbf, \sigma^2\Irm_{d})$}
\end{minipage}%
\begin{minipage}{.4\linewidth}
  \centering
  \scalebox{\scalesize}{
  \begin{tabular}{|c@{\hspace{0.15cm} }c@{\hspace{0.15cm}}c@{\hspace{0.15cm}}c|c@{\hspace{0.15cm}}c@{\hspace{0.15cm}}c|}
\toprule
\multicolumn{4}{c}{\Cref{theorem:KL-wass}} & \multicolumn{3}{c}{\citet{maurer2004note}} \\
Test & Bnd & Wass. & KL & Test & Bnd & KL \\
\midrule
0.770 & 5.288 & 0.000 & 14.706 & 0.479 & 0.737 & 0.023 \\
0.825 & 5.057 & 0.000 & 12.867 & 0.551 & 0.937 & 0.065 \\
0.456 & 6.313 & 0.000 & 24.986 & 0.340 & 0.711 & 0.058 \\
0.386 & 4.546 & 0.000 & 12.394 & 0.239 & 0.593 & 0.049 \\
0.586 & 7.151 & 0.000 & 31.416 & 0.412 & 0.758 & 0.026 \\
\bottomrule
\end{tabular}
}
  \caption*{(d) Neural networks $h_{\wbf}$ with $\Q=\Ncal(\wbf, \sigma^2\Irm_{d})$}
\end{minipage}
\label{table:expe-supp-1}
\end{table*}

\begin{table*}
\caption{
Results of the bound minimisation for linear models of the different bounds when the prior is initialized with the vector of zeros and for the Dirac posterior distribution $\Q=\delta_{\wbf}$.
``Test'' is the test risk $\Riskhat_{\Tcal}(h_{\wbf})$, ``Bnd'' represents the bound value, ``Wass'' represents the upper bound of the Wasserstein distance multiplied by the Lipschitz constant, and ``KL'' is the KL divergence divided by $2m$.
}
\centering
\begin{minipage}{.4\linewidth}
  \centering
  \scalebox{\scalesize}{
  \begin{tabular}{l|c@{\hspace{0.15cm}}c@{\hspace{0.15cm}}c@{\hspace{0.15cm}}c@{\hspace{0.15cm}}|}
\toprule
 & Test & Bnd & Wass. \\
\midrule
\fashion & 0.361 & 1.040 & 0.465 \\
\mnist & 0.304 & 1.078 & 0.583 \\
\mushrooms & 0.498 & 0.614 & 0.012 \\
\phishing & 0.497 & 0.569 & 0.004 \\
\yeast & 0.504 & 1.203 & 0.482 \\
\bottomrule
\end{tabular}
}
  \caption*{\Cref{theorem:KL-wass} with $\P=\eta$\\\citep{amit2022integral}}
\end{minipage}%
\begin{minipage}{.4\linewidth}
  \centering
  \scalebox{\scalesize}{
  \begin{tabular}{l|c@{\hspace{0.15cm}}c@{\hspace{0.15cm}}c@{\hspace{0.15cm}}c@{\hspace{0.15cm}}|}
\toprule
 & Test & Bnd & Wass. \\
\midrule
\fashion & 0.442 & 0.819 & 0.371 \\
\mnist & 0.480 & 0.872 & 0.375 \\
\mushrooms & 0.494 & 0.533 & 0.009 \\
\phishing & 0.497 & 0.527 & 0.005 \\
\yeast & 0.892 & 0.972 & 0.012 \\
\bottomrule
\end{tabular}
}
  \caption*{{\hspace{1.7cm}\Cref{eq:tv} with $\P=\eta$}\\\citep[Similar to Theorem 6 of][]{viallard2023learning}}
\end{minipage}
\begin{minipage}{.33\linewidth}
  \centering
  \scalebox{\scalesize}{
  \begin{tabular}{l|c@{\hspace{0.15cm}}c@{\hspace{0.15cm}}c@{\hspace{0.15cm}}c@{\hspace{0.15cm}}c@{\hspace{0.15cm}}|}
\toprule
 & Test & Bnd & Wass. & KL \\
\midrule
\fashion & 0.059 & 2.077 & 0.001 & 0.000 \\
\mnist & 0.033 & 2.056 & 0.001 & 0.000 \\
\mushrooms & 0.000 & 2.098 & 0.000 & 0.000 \\
\phishing & 0.068 & 2.145 & 0.000 & 0.000 \\
\yeast & 0.346 & 2.566 & 0.001 & 0.001 \\
\bottomrule
\end{tabular}
}
  \caption*{\Cref{eq:hellinger}}
\end{minipage}
\begin{minipage}{.33\linewidth}
  \centering
  \scalebox{\scalesize}{
  \begin{tabular}{l|c@{\hspace{0.15cm}}c@{\hspace{0.15cm}}c@{\hspace{0.15cm}}c@{\hspace{0.15cm}}c@{\hspace{0.15cm}}|}
\toprule
 & Test & Bnd & Wass. & KL \\
\midrule
\fashion & 0.894 & 1.037 & 0.038 & 0.000 \\
\mnist & 0.896 & 1.017 & 0.034 & 0.000 \\
\mushrooms & 0.498 & 0.728 & 0.012 & 0.000 \\
\phishing & 0.499 & 0.688 & 0.007 & 0.000 \\
\yeast & 0.896 & 1.398 & 0.012 & 0.000 \\
\bottomrule
\end{tabular}
}
  \caption*{\Cref{eq:reverse-KL}}
\end{minipage}
\begin{minipage}{.33\linewidth}
  \centering
  \scalebox{\scalesize}{
  \begin{tabular}{l|c@{\hspace{0.15cm}}c@{\hspace{0.15cm}}c@{\hspace{0.15cm}}c@{\hspace{0.15cm}}c@{\hspace{0.15cm}}|}
\toprule
 & Test & Bnd & Wass. & KL \\
\midrule
\fashion & 0.115 & 0.317 & 0.017 & 0.025 \\
\mnist & 0.077 & 0.294 & 0.018 & 0.027 \\
\mushrooms & 0.026 & 0.190 & 0.009 & 0.015 \\
\phishing & 0.085 & 0.225 & 0.005 & 0.013 \\
\yeast & 0.353 & 0.566 & 0.014 & 0.017 \\
\bottomrule
\end{tabular}
}
  \caption*{\Cref{theorem:KL-wass}}
\end{minipage}
\label{table:expe-supp-2}
\end{table*}

\begin{table*}
\caption{
Results of the bound minimisation for linear models of the different bounds when the prior weight vector is learned with 25\% of the original training set and for the Dirac posterior distribution $\Q=\delta_{\wbf}$.
``Test'' is the test risk $\Riskhat_{\Tcal}(h_{\wbf})$, ``Bnd'' represents the bound value, ``Wass'' represents the upper bound of the Wasserstein distance multiplied by the Lipschitz constant, and ``KL'' is the KL divergence divided by $2m$.
}
\centering
\begin{minipage}{.4\linewidth}
  \centering
  \scalebox{\scalesize}{
  \begin{tabular}{l|c@{\hspace{0.15cm}}c@{\hspace{0.15cm}}c@{\hspace{0.15cm}}c@{\hspace{0.15cm}}|}
\toprule
 & Test & Bnd & Wass. \\
\midrule
\fashion & 0.064 & 0.289 & 0.052 \\
\mnist & 0.038 & 0.199 & 0.025 \\
\mushrooms & 0.004 & 0.073 & 0.002 \\
\phishing & 0.070 & 0.129 & 0.002 \\
\yeast & 0.188 & 0.393 & 0.025 \\
\bottomrule
\end{tabular}
}
  \caption*{\Cref{theorem:KL-wass} with $\P=\eta$\\\citep{amit2022integral}}
\end{minipage}%
\begin{minipage}{.4\linewidth}
  \centering
  \scalebox{\scalesize}{
  \begin{tabular}{l|c@{\hspace{0.15cm}}c@{\hspace{0.15cm}}c@{\hspace{0.15cm}}c@{\hspace{0.15cm}}|}
\toprule
 & Test & Bnd & Wass. \\
\midrule
\fashion & 0.065 & 0.080 & 0.008 \\
\mnist & 0.038 & 0.063 & 0.010 \\
\mushrooms & 0.004 & 0.061 & 0.004 \\
\phishing & 0.070 & 0.115 & 0.003 \\
\yeast & 0.188 & 0.337 & 0.016 \\
\bottomrule
\end{tabular}
}
  \caption*{{\hspace{1.7cm}\Cref{eq:tv} with $\P=\eta$}\\\citep[Similar to Theorem 6 of][]{viallard2023learning}}
\end{minipage}
\begin{minipage}{.33\linewidth}
  \centering
  \scalebox{\scalesize}{
  \begin{tabular}{l|c@{\hspace{0.15cm}}c@{\hspace{0.15cm}}c@{\hspace{0.15cm}}c@{\hspace{0.15cm}}c@{\hspace{0.15cm}}|}
\toprule
 & Test & Bnd & Wass. & KL \\
\midrule
\fashion & 0.058 & 2.083 & 0.001 & 0.000 \\
\mnist & 0.033 & 2.062 & 0.001 & 0.000 \\
\mushrooms & 0.000 & 2.137 & 0.000 & 0.000 \\
\phishing & 0.068 & 2.178 & 0.000 & 0.000 \\
\yeast & 0.185 & 2.514 & 0.001 & 0.001 \\
\bottomrule
\end{tabular}
}
  \caption*{\Cref{eq:hellinger}}
\end{minipage}
\begin{minipage}{.33\linewidth}
  \centering
  \scalebox{\scalesize}{
  \begin{tabular}{l|c@{\hspace{0.15cm}}c@{\hspace{0.15cm}}c@{\hspace{0.15cm}}c@{\hspace{0.15cm}}c@{\hspace{0.15cm}}|}
\toprule
 & Test & Bnd & Wass. & KL \\
\midrule
\fashion & 0.065 & 0.192 & 0.028 & 0.000 \\
\mnist & 0.038 & 0.246 & 0.065 & 0.000 \\
\mushrooms & 0.004 & 0.313 & 0.014 & 0.000 \\
\phishing & 0.070 & 0.320 & 0.007 & 0.000 \\
\yeast & 0.188 & 0.896 & 0.018 & 0.000 \\
\bottomrule
\end{tabular}
}
  \caption*{\Cref{eq:reverse-KL}}
\end{minipage}
\begin{minipage}{.33\linewidth}
  \centering
  \scalebox{\scalesize}{
  \begin{tabular}{l|c@{\hspace{0.15cm}}c@{\hspace{0.15cm}}c@{\hspace{0.15cm}}c@{\hspace{0.15cm}}c@{\hspace{0.15cm}}|}
\toprule
 & Test & Bnd & Wass. & KL \\
\midrule
\fashion & 0.063 & 0.088 & 0.000 & 0.000 \\
\mnist & 0.038 & 0.068 & 0.000 & 0.000 \\
\mushrooms & 0.004 & 0.077 & 0.000 & 0.000 \\
\phishing & 0.070 & 0.130 & 0.000 & 0.000 \\
\yeast & 0.188 & 0.363 & 0.000 & 0.000 \\
\bottomrule
\end{tabular}
}
  \caption*{\Cref{theorem:KL-wass}}
\end{minipage}
\label{table:expe-supp-3}
\end{table*}

\begin{table*}
\caption{
Results of the bound minimisation for neural network models of the different bounds when the prior is initialized with the vector of zeros and for the Dirac posterior distribution $\Q=\delta_{\wbf}$.
``Test'' is the test risk $\Riskhat_{\Tcal}(h_{\wbf})$, ``Bnd'' represents the bound value, ``Wass'' represents the upper bound of the Wasserstein distance multiplied by the Lipschitz constant, and ``KL'' is the KL divergence divided by $2m$.
}
\centering
\begin{minipage}{.4\linewidth}
  \centering
  \scalebox{\scalesize}{
  \begin{tabular}{l|c@{\hspace{0.15cm}}c@{\hspace{0.15cm}}c@{\hspace{0.15cm}}c@{\hspace{0.15cm}}|}
\toprule
 & Test & Bnd & Wass. \\
\midrule
\fashion & 0.884 & 3.537 & 7.041 \\
\mnist & 0.776 & 2.414 & 2.680 \\
\mushrooms & 0.487 & 3.832 & 11.247 \\
\phishing & 0.438 & 1.903 & 2.095 \\
\yeast & 0.383 & 2.469 & 4.318 \\
\bottomrule
\end{tabular}
}
  \caption*{\Cref{theorem:KL-wass} with $\P=\eta$\\\citep{amit2022integral}}
\end{minipage}%
\begin{minipage}{.4\linewidth}
  \centering
  \scalebox{\scalesize}{
  \begin{tabular}{l|c@{\hspace{0.15cm}}c@{\hspace{0.15cm}}c@{\hspace{0.15cm}}c@{\hspace{0.15cm}}|}
\toprule
 & Test & Bnd & Wass. \\
\midrule
\fashion & 0.893 & 1.637 & 0.736 \\
\mnist & 0.806 & 1.583 & 0.769 \\
\mushrooms & 0.490 & 1.276 & 0.763 \\
\phishing & 0.436 & 1.458 & 0.979 \\
\yeast & 0.431 & 3.089 & 2.586 \\
\bottomrule
\end{tabular}
}
  \caption*{{\hspace{1.7cm}\Cref{eq:tv} with $\P=\eta$}\\\citep[Similar to Theorem 6 of][]{viallard2023learning}}
\end{minipage}
\begin{minipage}{.33\linewidth}
  \centering
  \scalebox{\scalesize}{
  \begin{tabular}{l|c@{\hspace{0.15cm}}c@{\hspace{0.15cm}}c@{\hspace{0.15cm}}c@{\hspace{0.15cm}}c@{\hspace{0.15cm}}|}
\toprule
 & Test & Bnd & Wass. & KL \\
\midrule
\fashion & 0.095 & 2.237 & 0.060 & 0.000 \\
\mnist & 0.057 & 2.204 & 0.060 & 0.000 \\
\mushrooms & 0.000 & 2.278 & 0.089 & 0.000 \\
\phishing & 0.047 & 2.232 & 0.060 & 0.000 \\
\yeast & 0.346 & 2.751 & 0.088 & 0.001 \\
\bottomrule
\end{tabular}
}
  \caption*{\Cref{eq:hellinger}}
\end{minipage}
\begin{minipage}{.33\linewidth}
  \centering
  \scalebox{\scalesize}{
  \begin{tabular}{l|c@{\hspace{0.15cm}}c@{\hspace{0.15cm}}c@{\hspace{0.15cm}}c@{\hspace{0.15cm}}c@{\hspace{0.15cm}}|}
\toprule
 & Test & Bnd & Wass. & KL \\
\midrule
\fashion & 0.891 & 3.785 & 1.421 & 0.000 \\
\mnist & 0.836 & 3.323 & 1.217 & 0.000 \\
\mushrooms & 0.493 & 6.007 & 2.501 & 0.000 \\
\phishing & 0.444 & 8.790 & 3.390 & 0.000 \\
\yeast & 0.435 & 9.357 & 3.312 & 0.001 \\
\bottomrule
\end{tabular}
}
  \caption*{\Cref{eq:reverse-KL}}
\end{minipage}
\begin{minipage}{.33\linewidth}
  \centering
  \scalebox{\scalesize}{
  \begin{tabular}{l|c@{\hspace{0.15cm}}c@{\hspace{0.15cm}}c@{\hspace{0.15cm}}c@{\hspace{0.15cm}}c@{\hspace{0.15cm}}|}
\toprule
 & Test & Bnd & Wass. & KL \\
\midrule
\fashion & 0.162 & 0.683 & 0.135 & 0.139 \\
\mnist & 0.111 & 0.673 & 0.152 & 0.158 \\
\mushrooms & 0.082 & 0.645 & 0.167 & 0.148 \\
\phishing & 0.123 & 0.642 & 0.132 & 0.136 \\
\yeast & 0.335 & 0.707 & 0.061 & 0.059 \\
\bottomrule
\end{tabular}
}
  \caption*{\Cref{theorem:KL-wass}}
\end{minipage}
\label{table:expe-supp-4}
\end{table*}

\begin{table*}
\caption{
Results of the bound minimisation for neural network models of the different bounds when the prior weight vector is learned with 25\% of the original training set and for the Dirac posterior distribution $\Q=\delta_{\wbf}$.
``Test'' is the test risk $\Riskhat_{\Tcal}(h_{\wbf})$, ``Bnd'' represents the bound value, ``Wass'' represents the upper bound of the Wasserstein distance multiplied by the Lipschitz constant, and ``KL'' is the KL divergence divided by $2m$.
}
\centering
\begin{minipage}{.4\linewidth}
  \centering
  \scalebox{\scalesize}{
  \begin{tabular}{l|c@{\hspace{0.15cm}}c@{\hspace{0.15cm}}c@{\hspace{0.15cm}}c@{\hspace{0.15cm}}|}
\toprule
 & Test & Bnd & Wass. \\
\midrule
\fashion & 0.125 & 0.640 & 0.267 \\
\mnist & 0.089 & 0.286 & 0.037 \\
\mushrooms & 0.002 & 0.277 & 0.073 \\
\phishing & 0.068 & 0.965 & 0.805 \\
\yeast & 0.328 & 2.138 & 3.217 \\
\bottomrule
\end{tabular}
}
  \caption*{\Cref{theorem:KL-wass} with $\P=\eta$\\\citep{amit2022integral}}
\end{minipage}%
\begin{minipage}{.4\linewidth}
  \centering
  \scalebox{\scalesize}{
  \begin{tabular}{l|c@{\hspace{0.15cm}}c@{\hspace{0.15cm}}c@{\hspace{0.15cm}}c@{\hspace{0.15cm}}|}
\toprule
 & Test & Bnd & Wass. \\
\midrule
\fashion & 0.113 & 0.213 & 0.090 \\
\mnist & 0.091 & 0.190 & 0.082 \\
\mushrooms & 0.003 & 0.437 & 0.380 \\
\phishing & 0.068 & 0.345 & 0.234 \\
\yeast & 0.325 & 1.291 & 0.824 \\
\bottomrule
\end{tabular}
}
  \caption*{{\hspace{1.7cm}\Cref{eq:tv} with $\P=\eta$}\\\citep[Similar to Theorem 6 of][]{viallard2023learning}}
\end{minipage}
\begin{minipage}{.33\linewidth}
  \centering
  \scalebox{\scalesize}{
  \begin{tabular}{l|c@{\hspace{0.15cm}}c@{\hspace{0.15cm}}c@{\hspace{0.15cm}}c@{\hspace{0.15cm}}c@{\hspace{0.15cm}}|}
\toprule
 & Test & Bnd & Wass. & KL \\
\midrule
\fashion & 0.427 & 2.580 & 0.060 & 0.000 \\
\mnist & 0.210 & 2.372 & 0.060 & 0.000 \\
\mushrooms & 0.000 & 2.315 & 0.088 & 0.000 \\
\phishing & 0.055 & 2.276 & 0.060 & 0.000 \\
\yeast & 0.345 & 2.856 & 0.087 & 0.001 \\
\bottomrule
\end{tabular}
}
  \caption*{\Cref{eq:hellinger}}
\end{minipage}
\begin{minipage}{.33\linewidth}
  \centering
  \scalebox{\scalesize}{
  \begin{tabular}{l|c@{\hspace{0.15cm}}c@{\hspace{0.15cm}}c@{\hspace{0.15cm}}c@{\hspace{0.15cm}}c@{\hspace{0.15cm}}|}
\toprule
 & Test & Bnd & Wass. & KL \\
\midrule
\fashion & 0.112 & 6.877 & 3.350 & 0.000 \\
\mnist & 0.090 & 5.453 & 2.770 & -0.000 \\
\mushrooms & 0.002 & 2.269 & 0.962 & 0.000 \\
\phishing & 0.069 & 1.816 & 0.725 & 0.000 \\
\yeast & 0.325 & 3.433 & 1.116 & 0.000 \\
\bottomrule
\end{tabular}
}
  \caption*{\Cref{eq:reverse-KL}}
\end{minipage}
\begin{minipage}{.33\linewidth}
  \centering
  \scalebox{\scalesize}{
  \begin{tabular}{l|c@{\hspace{0.15cm}}c@{\hspace{0.15cm}}c@{\hspace{0.15cm}}c@{\hspace{0.15cm}}c@{\hspace{0.15cm}}|}
\toprule
 & Test & Bnd & Wass. & KL \\
\midrule
\fashion & 0.111 & 0.248 & 0.014 & 0.006 \\
\mnist & 0.088 & 0.238 & 0.016 & 0.005 \\
\mushrooms & 0.002 & 0.177 & 0.019 & 0.007 \\
\phishing & 0.068 & 0.211 & 0.014 & 0.003 \\
\yeast & 0.329 & 0.614 & 0.034 & 0.012 \\
\bottomrule
\end{tabular}
}
  \caption*{\Cref{theorem:KL-wass}}
\end{minipage}
\label{table:expe-supp-5}
\end{table*}

\begin{table*}
\caption{
Results of the bound minimisation for linear models of the different bounds when the prior is initialized with the vector of zeros and for the Gaussian posterior distribution $\Q=\Ncal(\wbf, \sigma^2\Irm_{d})$.
``Test'' is the test risk $\Riskhat_{\Tcal}(h_{\wbf})$, ``Bnd'' represents the bound value, ``Wass'' represents the upper bound of the Wasserstein distance multiplied by the Lipschitz constant, and ``KL'' is the KL divergence divided by $2m$.
}
\centering
\begin{minipage}{.4\linewidth}
  \centering
  \scalebox{\scalesize}{
  \begin{tabular}{l|c@{\hspace{0.15cm}}c@{\hspace{0.15cm}}c@{\hspace{0.15cm}}c@{\hspace{0.15cm}}|}
\toprule
 & Test & Bnd & Wass. \\
\midrule
\fashion & 0.256 & 6.153 & 33.787 \\
\mnist & 0.190 & 8.651 & 69.962 \\
\mushrooms & 0.368 & 0.937 & 0.228 \\
\phishing & 0.498 & 0.709 & 0.012 \\
\yeast & 0.635 & 1.347 & 0.370 \\
\bottomrule
\end{tabular}
}
  \caption*{\Cref{theorem:KL-wass} with $\P=\eta$\\\citep{amit2022integral}}
\end{minipage}%
\begin{minipage}{.4\linewidth}
  \centering
  \scalebox{\scalesize}{
  \begin{tabular}{l|c@{\hspace{0.15cm}}c@{\hspace{0.15cm}}c@{\hspace{0.15cm}}c@{\hspace{0.15cm}}|}
\toprule
 & Test & Bnd & Wass. \\
\midrule
\fashion & 0.515 & 8.684 & 8.067 \\
\mnist & 0.510 & 8.761 & 8.142 \\
\mushrooms & 0.460 & 3.161 & 2.559 \\
\phishing & 0.485 & 1.895 & 1.275 \\
\yeast & 0.820 & 3.811 & 2.779 \\
\bottomrule
\end{tabular}
}
  \caption*{{\hspace{1.7cm}\Cref{eq:tv} with $\P=\eta$}\\\citep[Similar to Theorem 6 of][]{viallard2023learning}}
\end{minipage}
\begin{minipage}{.33\linewidth}
  \centering
  \scalebox{\scalesize}{
  \begin{tabular}{l|c@{\hspace{0.15cm}}c@{\hspace{0.15cm}}c@{\hspace{0.15cm}}c@{\hspace{0.15cm}}|}
\toprule
 & Test & Bnd & KL \\
\midrule
\fashion & 0.059 & 2.150 & 0.000 \\
\mnist & 0.033 & 2.129 & 0.000 \\
\mushrooms & 0.000 & 2.147 & 0.000 \\
\phishing & 0.069 & 2.201 & 0.000 \\
\yeast & 0.167 & 2.389 & 0.001 \\
\bottomrule
\end{tabular}
}
  \caption*{\Cref{eq:hellinger} with $\Q=\eta$}
\end{minipage}
\begin{minipage}{.33\linewidth}
  \centering
  \scalebox{\scalesize}{
  \begin{tabular}{l|c@{\hspace{0.15cm}}c@{\hspace{0.15cm}}c@{\hspace{0.15cm}}c@{\hspace{0.15cm}}|}
\toprule
 & Test & Bnd & KL \\
\midrule
\fashion & 0.843 & 1.006 & 0.000 \\
\mnist & 0.841 & 0.997 & 0.000 \\
\mushrooms & 0.481 & 0.846 & 0.000 \\
\phishing & 0.491 & 0.768 & 0.000 \\
\yeast & 0.726 & 1.309 & 0.000 \\
\bottomrule
\end{tabular}
}
  \caption*{\Cref{eq:reverse-KL} with $\Q=\eta$}
\end{minipage}
\begin{minipage}{.33\linewidth}
  \centering
  \scalebox{\scalesize}{
  \begin{tabular}{l|c@{\hspace{0.15cm}}c@{\hspace{0.15cm}}c@{\hspace{0.15cm}}c@{\hspace{0.15cm}}|}
\toprule
 & Test & Bnd & KL \\
\midrule
\fashion & 0.140 & 0.359 & 0.019 \\
\mnist & 0.096 & 0.333 & 0.021 \\
\mushrooms & 0.035 & 0.258 & 0.016 \\
\phishing & 0.094 & 0.297 & 0.013 \\
\yeast & 0.372 & 0.638 & 0.016 \\
\bottomrule
\end{tabular}
}
  \caption*{\Cref{theorem:KL-wass} with $\Q=\eta$}
\end{minipage}
\begin{minipage}{.33\linewidth}
  \centering
  \scalebox{\scalesize}{
  \begin{tabular}{l|c@{\hspace{0.15cm}}c@{\hspace{0.15cm}}c@{\hspace{0.15cm}}c@{\hspace{0.15cm}}c@{\hspace{0.15cm}}|}
\toprule
 & Test & Bnd & Wass. & KL \\
\midrule
\fashion & 0.217 & 2.415 & 0.045 & 0.000 \\
\mnist & 0.128 & 2.338 & 0.045 & 0.000 \\
\mushrooms & 0.476 & 0.713 & 0.001 & 0.000 \\
\phishing & 0.498 & 0.684 & 0.000 & 0.000 \\
\yeast & 0.688 & 1.094 & 0.000 & 0.000 \\
\bottomrule
\end{tabular}
}
  \caption*{\Cref{eq:hellinger}}
\end{minipage}
\begin{minipage}{.33\linewidth}
  \centering
  \scalebox{\scalesize}{
  \begin{tabular}{l|c@{\hspace{0.15cm}}c@{\hspace{0.15cm}}c@{\hspace{0.15cm}}c@{\hspace{0.15cm}}c@{\hspace{0.15cm}}|}
\toprule
 & Test & Bnd & Wass. & KL \\
\midrule
\fashion & 0.784 & 0.965 & 0.011 & 0.000 \\
\mnist & 0.783 & 0.960 & 0.011 & 0.000 \\
\mushrooms & 0.490 & 0.814 & 0.002 & 0.000 \\
\phishing & 0.499 & 0.765 & 0.000 & 0.000 \\
\yeast & 0.726 & 1.312 & 0.000 & 0.000 \\
\bottomrule
\end{tabular}
}
  \caption*{\Cref{eq:reverse-KL}}
\end{minipage}
\begin{minipage}{.33\linewidth}
  \centering
  \scalebox{\scalesize}{
  \begin{tabular}{l|c@{\hspace{0.15cm}}c@{\hspace{0.15cm}}c@{\hspace{0.15cm}}c@{\hspace{0.15cm}}c@{\hspace{0.15cm}}|}
\toprule
 & Test & Bnd & Wass. & KL \\
\midrule
\fashion & 0.140 & 0.364 & 0.000 & 0.019 \\
\mnist & 0.097 & 0.338 & 0.000 & 0.021 \\
\mushrooms & 0.017 & 0.265 & 0.000 & 0.022 \\
\phishing & 0.094 & 0.302 & 0.000 & 0.013 \\
\yeast & 0.372 & 0.644 & 0.000 & 0.016 \\
\bottomrule
\end{tabular}
}
  \caption*{\Cref{theorem:KL-wass}}
\end{minipage}
\label{table:expe-supp-6}
\end{table*}

\renewcommand{\scalesize}{0.71}

\begin{table*}
\caption{
Results of the bound minimisation for linear models of the different bounds when the prior weight vector is learned with 25\% of the original training set, \ie, and for the Gaussian posterior distribution $\Q=\Ncal(\wbf, \sigma^2\Irm_{d})$.
``Test'' is the test risk $\Riskhat_{\Tcal}(h_{\wbf})$, ``Bnd'' represents the bound value, ``Wass'' represents the upper bound of the Wasserstein distance multiplied by the Lipschitz constant, and ``KL'' is the KL divergence divided by $2m$.
}
\centering
\begin{minipage}{.4\linewidth}
  \centering
  \scalebox{\scalesize}{
  \begin{tabular}{l|c@{\hspace{0.15cm}}c@{\hspace{0.15cm}}c@{\hspace{0.15cm}}c@{\hspace{0.15cm}}|}
\toprule
 & Test & Bnd & Wass. \\
\midrule
\fashion & 0.540 & 6.338 & 32.621 \\
\mnist & 0.458 & 6.798 & 38.968 \\
\mushrooms & 0.060 & 3.570 & 11.686 \\
\phishing & 0.071 & 0.534 & 0.136 \\
\yeast & 0.199 & 1.460 & 1.330 \\
\bottomrule
\end{tabular}
}
  \caption*{\Cref{theorem:KL-wass} with $\P=\eta$\\\citep{amit2022integral}}
\end{minipage}%
\begin{minipage}{.4\linewidth}
  \centering
  \scalebox{\scalesize}{
  \begin{tabular}{l|c@{\hspace{0.15cm}}c@{\hspace{0.15cm}}c@{\hspace{0.15cm}}c@{\hspace{0.15cm}}|}
\toprule
 & Test & Bnd & Wass. \\
\midrule
\fashion & 0.835 & 1.588 & 0.647 \\
\mnist & 0.838 & 1.575 & 0.632 \\
\mushrooms & 0.401 & 0.633 & 0.080 \\
\phishing & 0.465 & 0.669 & 0.061 \\
\yeast & 0.858 & 1.193 & 0.094 \\
\bottomrule
\end{tabular}
}
  \caption*{{\hspace{1.7cm}\Cref{eq:tv} with $\P=\eta$}\\\citep[Similar to Theorem 6 of][]{viallard2023learning}}
\end{minipage}
\begin{minipage}{.33\linewidth}
  \centering
  \scalebox{\scalesize}{
  \begin{tabular}{l|c@{\hspace{0.15cm}}c@{\hspace{0.15cm}}c@{\hspace{0.15cm}}c@{\hspace{0.15cm}}|}
\toprule
 & Test & Bnd & KL \\
\midrule
\fashion & 0.058 & 2.157 & 0.000 \\
\mnist & 0.034 & 2.136 & 0.000 \\
\mushrooms & 0.000 & 2.193 & 0.000 \\
\phishing & 0.069 & 2.238 & 0.000 \\
\yeast & 0.185 & 2.535 & 0.001 \\
\bottomrule
\end{tabular}
}
  \caption*{\Cref{eq:hellinger} with $\Q=\eta$}
\end{minipage}
\begin{minipage}{.33\linewidth}
  \centering
  \scalebox{\scalesize}{
  \begin{tabular}{l|c@{\hspace{0.15cm}}c@{\hspace{0.15cm}}c@{\hspace{0.15cm}}c@{\hspace{0.15cm}}|}
\toprule
 & Test & Bnd & KL \\
\midrule
\fashion & 0.782 & 0.966 & 0.000 \\
\mnist & 0.785 & 0.963 & 0.000 \\
\mushrooms & 0.062 & 0.749 & 0.000 \\
\phishing & 0.151 & 0.479 & 0.000 \\
\yeast & 0.283 & 1.081 & 0.000 \\
\bottomrule
\end{tabular}
}
  \caption*{\Cref{eq:reverse-KL} with $\Q=\eta$}
\end{minipage}
\begin{minipage}{.33\linewidth}
  \centering
  \scalebox{\scalesize}{
  \begin{tabular}{l|c@{\hspace{0.15cm}}c@{\hspace{0.15cm}}c@{\hspace{0.15cm}}c@{\hspace{0.15cm}}|}
\toprule
 & Test & Bnd & KL \\
\midrule
\fashion & 0.068 & 0.171 & 0.000 \\
\mnist & 0.042 & 0.153 & 0.000 \\
\mushrooms & 0.005 & 0.292 & 0.038 \\
\phishing & 0.076 & 0.396 & 0.056 \\
\yeast & 0.220 & 0.883 & 0.312 \\
\bottomrule
\end{tabular}
}
  \caption*{\Cref{theorem:KL-wass} with $\Q=\eta$}
\end{minipage}
\begin{minipage}{.33\linewidth}
  \centering
  \scalebox{\scalesize}{
  \begin{tabular}{l|c@{\hspace{0.15cm}}c@{\hspace{0.15cm}}c@{\hspace{0.15cm}}c@{\hspace{0.15cm}}c@{\hspace{0.15cm}}|}
\toprule
 & Test & Bnd & Wass. & KL \\
\midrule
\fashion & 0.229 & 2.349 & 0.005 & 0.000 \\
\mnist & 0.131 & 2.262 & 0.005 & 0.000 \\
\mushrooms & 0.000 & 2.216 & 0.000 & 0.000 \\
\phishing & 0.077 & 2.269 & 0.000 & 0.000 \\
\yeast & 0.285 & 0.737 & 0.000 & 0.000 \\
\bottomrule
\end{tabular}
}
  \caption*{\Cref{eq:hellinger}}
\end{minipage}
\begin{minipage}{.33\linewidth}
  \centering
  \scalebox{\scalesize}{
  \begin{tabular}{l|c@{\hspace{0.15cm}}c@{\hspace{0.15cm}}c@{\hspace{0.15cm}}c@{\hspace{0.15cm}}c@{\hspace{0.15cm}}|}
\toprule
 & Test & Bnd & Wass. & KL \\
\midrule
\fashion & 0.832 & 1.106 & 0.003 & 0.000 \\
\mnist & 0.834 & 1.085 & 0.003 & 0.000 \\
\mushrooms & 0.055 & 0.419 & 0.000 & 0.000 \\
\phishing & 0.130 & 0.454 & 0.000 & 0.000 \\
\yeast & 0.278 & 1.042 & 0.000 & 0.000 \\
\bottomrule
\end{tabular}
}
  \caption*{\Cref{eq:reverse-KL}}
\end{minipage}
\begin{minipage}{.33\linewidth}
  \centering
  \scalebox{\scalesize}{
  \begin{tabular}{l|c@{\hspace{0.15cm}}c@{\hspace{0.15cm}}c@{\hspace{0.15cm}}c@{\hspace{0.15cm}}c@{\hspace{0.15cm}}|}
\toprule
 & Test & Bnd & Wass. & KL \\
\midrule
\fashion & 0.119 & 0.603 & 0.000 & 0.163 \\
\mnist & 0.079 & 0.531 & 0.000 & 0.131 \\
\mushrooms & 0.005 & 0.296 & 0.000 & 0.037 \\
\phishing & 0.076 & 0.226 & 0.000 & 0.000 \\
\yeast & 0.190 & 0.454 & 0.000 & 0.000 \\
\bottomrule
\end{tabular}
}
  \caption*{\Cref{theorem:KL-wass}}
\end{minipage}
\label{table:expe-supp-7}
\end{table*}

\begin{table*}
\caption{
Results of the bound minimisation for neural network models of the different bounds when the prior is initialized with the vector of zeros and for the Gaussian posterior distribution $\Q=\Ncal(\wbf, \sigma^2\Irm_{d})$.
``Test'' is the test risk $\Riskhat_{\Tcal}(h_{\wbf})$, ``Bnd'' represents the bound value, ``Wass'' represents the upper bound of the Wasserstein distance multiplied by the Lipschitz constant, and ``KL'' is the KL divergence divided by $2m$.
}
\centering
\begin{minipage}{.4\linewidth}
  \centering
  \scalebox{\scalesize}{
  \begin{tabular}{l|c@{\hspace{0.15cm}}c@{\hspace{0.15cm}}c@{\hspace{0.15cm}}c@{\hspace{0.15cm}}|}
\toprule
 & Test & Bnd & Wass. \\
\midrule
\fashion & 0.844 & 101.597 & 10133.044 \\
\mnist & 0.846 & 40.370 & 1554.983 \\
\mushrooms & 0.489 & 134.804 & 18016.426 \\
\phishing & 0.488 & 51.590 & 2602.070 \\
\yeast & 0.726 & 69.722 & 4747.634 \\
\bottomrule
\end{tabular}
}
  \caption*{\Cref{theorem:KL-wass} with $\P=\eta$\\\citep{amit2022integral}}
\end{minipage}%
\begin{minipage}{.4\linewidth}
  \centering
  \scalebox{\scalesize}{
  \begin{tabular}{l|c@{\hspace{0.15cm}}c@{\hspace{0.15cm}}c@{\hspace{0.15cm}}c@{\hspace{0.15cm}}|}
\toprule
 & Test & Bnd & Wass. \\
\midrule
\fashion & 0.855 & 390.291 & 389.331 \\
\mnist & 0.856 & 1153.956 & 1152.995 \\
\mushrooms & 0.509 & 425.913 & 425.253 \\
\phishing & 0.464 & 2371.850 & 2371.239 \\
\yeast & 0.736 & 968.250 & 967.294 \\
\bottomrule
\end{tabular}
}
  \caption*{{\hspace{1.7cm}\Cref{eq:tv} with $\P=\eta$}\\\citep[Similar to Theorem 6 of][]{viallard2023learning}}
\end{minipage}
\begin{minipage}{.33\linewidth}
  \centering
  \scalebox{\scalesize}{
  \begin{tabular}{l|c@{\hspace{0.15cm}}c@{\hspace{0.15cm}}c@{\hspace{0.15cm}}c@{\hspace{0.15cm}}|}
\toprule
 & Test & Bnd & KL \\
\midrule
\fashion & 0.147 & 2.246 & 0.000 \\
\mnist & 0.087 & 2.193 & 0.000 \\
\mushrooms & 0.000 & 2.147 & 0.000 \\
\phishing & 0.070 & 2.203 & 0.000 \\
\yeast & 0.335 & 2.566 & 0.001 \\
\bottomrule
\end{tabular}
}
  \caption*{\Cref{eq:hellinger} with $\Q=\eta$}
\end{minipage}
\begin{minipage}{.33\linewidth}
  \centering
  \scalebox{\scalesize}{
  \begin{tabular}{l|c@{\hspace{0.15cm}}c@{\hspace{0.15cm}}c@{\hspace{0.15cm}}c@{\hspace{0.15cm}}|}
\toprule
 & Test & Bnd & KL \\
\midrule
\fashion & 0.888 & 53.574 & 0.000 \\
\mnist & 0.887 & 53.435 & 0.000 \\
\mushrooms & 0.502 & 32.720 & 0.002 \\
\phishing & 0.493 & 28.921 & 0.001 \\
\yeast & 0.772 & 24.389 & 0.008 \\
\bottomrule
\end{tabular}
}
  \caption*{\Cref{eq:reverse-KL} with $\Q=\eta$}
\end{minipage}
\begin{minipage}{.33\linewidth}
  \centering
  \scalebox{\scalesize}{
  \begin{tabular}{l|c@{\hspace{0.15cm}}c@{\hspace{0.15cm}}c@{\hspace{0.15cm}}c@{\hspace{0.15cm}}|}
\toprule
 & Test & Bnd & KL \\
\midrule
\fashion & 0.647 & 1.115 & 0.147 \\
\mnist & 0.623 & 1.105 & 0.154 \\
\mushrooms & 0.376 & 0.934 & 0.220 \\
\phishing & 0.243 & 0.813 & 0.232 \\
\yeast & 0.652 & 2.335 & 2.472 \\
\bottomrule
\end{tabular}
}
  \caption*{\Cref{theorem:KL-wass} with $\Q=\eta$}
\end{minipage}
\begin{minipage}{.33\linewidth}
  \centering
  \scalebox{\scalesize}{
  \begin{tabular}{l|c@{\hspace{0.15cm}}c@{\hspace{0.15cm}}c@{\hspace{0.15cm}}c@{\hspace{0.15cm}}c@{\hspace{0.15cm}}|}
\toprule
 & Test & Bnd & Wass. & KL \\
\midrule
\fashion & 0.895 & 47.296 & 22.157 & 0.000 \\
\mnist & 0.895 & 16.096 & 6.591 & 0.000 \\
\mushrooms & 0.501 & 12.524 & 5.236 & 0.000 \\
\phishing & 0.500 & 9.806 & 3.891 & 0.000 \\
\yeast & 0.748 & 6.761 & 1.932 & 0.001 \\
\bottomrule
\end{tabular}
}
  \caption*{\Cref{eq:hellinger}}
\end{minipage}
\begin{minipage}{.33\linewidth}
  \centering
  \scalebox{\scalesize}{
  \begin{tabular}{l|c@{\hspace{0.15cm}}c@{\hspace{0.15cm}}c@{\hspace{0.15cm}}c@{\hspace{0.15cm}}c@{\hspace{0.15cm}}|}
\toprule
 & Test & Bnd & Wass. & KL \\
\midrule
\fashion & 0.891 & 59.620 & 14.966 & 0.000 \\
\mnist & 0.891 & 59.621 & 14.966 & 0.000 \\
\mushrooms & 0.501 & 14.762 & 2.610 & 0.001 \\
\phishing & 0.499 & 11.764 & 1.196 & 0.000 \\
\yeast & 0.768 & 16.625 & 1.772 & 0.004 \\
\bottomrule
\end{tabular}
}
  \caption*{\Cref{eq:reverse-KL}}
\end{minipage}
\begin{minipage}{.33\linewidth}
  \centering
  \scalebox{\scalesize}{
  \begin{tabular}{l|c@{\hspace{0.15cm}}c@{\hspace{0.15cm}}c@{\hspace{0.15cm}}c@{\hspace{0.15cm}}c@{\hspace{0.15cm}}|}
\toprule
 & Test & Bnd & Wass. & KL \\
\midrule
\fashion & 0.663 & 1.132 & 0.001 & 0.143 \\
\mnist & 0.610 & 1.105 & 0.001 & 0.160 \\
\mushrooms & 0.384 & 0.935 & 0.001 & 0.209 \\
\phishing & 0.287 & 0.972 & 0.001 & 0.348 \\
\yeast & 0.648 & 2.295 & 0.003 & 2.342 \\
\bottomrule
\end{tabular}
}
  \caption*{\Cref{theorem:KL-wass}}
\end{minipage}
\label{table:expe-supp-8}
\end{table*}

\begin{table*}
\caption{
Results of the bound minimisation for neural network models of the different bounds when the prior weight vector is learned with 25\% of the original training set, \ie, and for the Gaussian posterior distribution $\Q=\Ncal(\wbf, \sigma^2\Irm_{d})$.
``Test'' is the test risk $\Riskhat_{\Tcal}(h_{\wbf})$, ``Bnd'' represents the bound value, ``Wass'' represents the upper bound of the Wasserstein distance multiplied by the Lipschitz constant, and ``KL'' is the KL divergence divided by $2m$.
}
\centering
\begin{minipage}{.4\linewidth}
  \centering
  \scalebox{\scalesize}{
  \begin{tabular}{l|c@{\hspace{0.15cm}}c@{\hspace{0.15cm}}c@{\hspace{0.15cm}}c@{\hspace{0.15cm}}|}
\toprule
 & Test & Bnd & Wass. \\
\midrule
\fashion & 0.898 & 24.128 & 535.465 \\
\mnist & 0.898 & 39.281 & 1466.337 \\
\mushrooms & 0.500 & 19.780 & 368.224 \\
\phishing & 0.500 & 7.916 & 53.671 \\
\yeast & 0.760 & 20.853 & 399.991 \\
\bottomrule
\end{tabular}
}
  \caption*{\Cref{theorem:KL-wass} with $\P=\eta$\\\citep{amit2022integral}}
\end{minipage}%
\begin{minipage}{.4\linewidth}
  \centering
  \scalebox{\scalesize}{
  \begin{tabular}{l|c@{\hspace{0.15cm}}c@{\hspace{0.15cm}}c@{\hspace{0.15cm}}c@{\hspace{0.15cm}}|}
\toprule
 & Test & Bnd & Wass. \\
\midrule
\fashion & 0.898 & 411.972 & 410.968 \\
\mnist & 0.898 & 412.984 & 411.980 \\
\mushrooms & 0.500 & 228.537 & 227.884 \\
\phishing & 0.500 & 124.881 & 124.238 \\
\yeast & 0.890 & 101.518 & 100.387 \\
\bottomrule
\end{tabular}
}
  \caption*{{\hspace{1.7cm}\Cref{eq:tv} with $\P=\eta$}\\\citep[Similar to Theorem 6 of][]{viallard2023learning}}
\end{minipage}
\begin{minipage}{.33\linewidth}
  \centering
  \scalebox{\scalesize}{
  \begin{tabular}{l|c@{\hspace{0.15cm}}c@{\hspace{0.15cm}}c@{\hspace{0.15cm}}c@{\hspace{0.15cm}}|}
\toprule
 & Test & Bnd & KL \\
\midrule
\fashion & 0.138 & 2.246 & 0.000 \\
\mnist & 0.188 & 2.305 & 0.000 \\
\mushrooms & 0.004 & 2.197 & 0.000 \\
\phishing & 0.065 & 2.232 & 0.000 \\
\yeast & 0.332 & 2.693 & 0.001 \\
\bottomrule
\end{tabular}
}
  \caption*{\Cref{eq:hellinger} with $\Q=\eta$}
\end{minipage}
\begin{minipage}{.33\linewidth}
  \centering
  \scalebox{\scalesize}{
  \begin{tabular}{l|c@{\hspace{0.15cm}}c@{\hspace{0.15cm}}c@{\hspace{0.15cm}}c@{\hspace{0.15cm}}|}
\toprule
 & Test & Bnd & KL \\
\midrule
\fashion & 0.897 & 1.121 & 0.000 \\
\mnist & 0.897 & 1.136 & 0.000 \\
\mushrooms & 0.500 & 0.949 & 0.000 \\
\phishing & 0.500 & 1.191 & 0.000 \\
\yeast & 0.807 & 18.957 & 0.008 \\
\bottomrule
\end{tabular}
}
  \caption*{\Cref{eq:reverse-KL} with $\Q=\eta$}
\end{minipage}
\begin{minipage}{.33\linewidth}
  \centering
  \scalebox{\scalesize}{
  \begin{tabular}{l|c@{\hspace{0.15cm}}c@{\hspace{0.15cm}}c@{\hspace{0.15cm}}c@{\hspace{0.15cm}}|}
\toprule
 & Test & Bnd & KL \\
\midrule
\fashion & 0.479 & 0.737 & 0.030 \\
\mnist & 0.551 & 0.937 & 0.087 \\
\mushrooms & 0.340 & 0.711 & 0.077 \\
\phishing & 0.239 & 0.593 & 0.065 \\
\yeast & 0.412 & 0.758 & 0.035 \\
\bottomrule
\end{tabular}
}
  \caption*{\Cref{theorem:KL-wass} with $\Q=\eta$}
\end{minipage}
\begin{minipage}{.33\linewidth}
  \centering
  \scalebox{\scalesize}{
  \begin{tabular}{l|c@{\hspace{0.15cm}}c@{\hspace{0.15cm}}c@{\hspace{0.15cm}}c@{\hspace{0.15cm}}c@{\hspace{0.15cm}}|}
\toprule
 & Test & Bnd & Wass. & KL \\
\midrule
\fashion & 0.898 & 5.087 & 1.998 & 0.000 \\
\mnist & 0.898 & 5.085 & 1.998 & 0.000 \\
\mushrooms & 0.500 & 2.963 & 1.108 & 0.000 \\
\phishing & 0.500 & 1.647 & 0.458 & 0.000 \\
\yeast & 0.754 & 4.557 & 1.032 & 0.001 \\
\bottomrule
\end{tabular}
}
  \caption*{\Cref{eq:hellinger}}
\end{minipage}
\begin{minipage}{.33\linewidth}
  \centering
  \scalebox{\scalesize}{
  \begin{tabular}{l|c@{\hspace{0.15cm}}c@{\hspace{0.15cm}}c@{\hspace{0.15cm}}c@{\hspace{0.15cm}}c@{\hspace{0.15cm}}|}
\toprule
 & Test & Bnd & Wass. & KL \\
\midrule
\fashion & 0.898 & 5.113 & 1.997 & 0.000 \\
\mnist & 0.898 & 6.115 & 2.496 & 0.000 \\
\mushrooms & 0.500 & 2.527 & 0.830 & 0.000 \\
\phishing & 0.500 & 1.460 & 0.305 & 0.000 \\
\yeast & 0.767 & 8.521 & 0.967 & 0.002 \\
\bottomrule
\end{tabular}
}
  \caption*{\Cref{eq:reverse-KL}}
\end{minipage}
\begin{minipage}{.33\linewidth}
  \centering
  \scalebox{\scalesize}{
  \begin{tabular}{l|c@{\hspace{0.15cm}}c@{\hspace{0.15cm}}c@{\hspace{0.15cm}}c@{\hspace{0.15cm}}c@{\hspace{0.15cm}}|}
\toprule
 & Test & Bnd & Wass. & KL \\
\midrule
\fashion & 0.770 & 5.288 & 0.000 & 19.609 \\
\mnist & 0.825 & 5.057 & 0.000 & 17.155 \\
\mushrooms & 0.456 & 6.313 & 0.000 & 33.309 \\
\phishing & 0.386 & 4.546 & 0.000 & 16.523 \\
\yeast & 0.586 & 7.151 & 0.000 & 41.850 \\
\bottomrule
\end{tabular}
}
  \caption*{\Cref{theorem:KL-wass}}
\end{minipage}
\label{table:expe-supp-9}
\end{table*}

\end{document}